\documentclass{article}

\usepackage[preprint]{neurips_2026}

\usepackage[utf8]{inputenc} 
\usepackage[T1]{fontenc}    
\usepackage{hyperref}       
\usepackage{url}            
\usepackage{booktabs}       
\usepackage{multirow}       
\usepackage{amsfonts}       
\usepackage{amsmath,amssymb}
\usepackage{lmodern}        
\usepackage{nicefrac}       
\usepackage{microtype}      
\usepackage{xcolor}         
\usepackage{natbib}
\usepackage{graphicx}       
\usepackage{float}          
\usepackage{tikz}           
\usetikzlibrary{arrows.meta, positioning, calc, fit, shapes.geometric,
                decorations.pathreplacing, patterns, shadows.blur, backgrounds}
\usepackage[breakable, skins]{tcolorbox}  
\usepackage{algorithm}      
\usepackage{algpseudocode}  
\usepackage{hyperref}
\usepackage{caption}
\usepackage{subcaption}
\definecolor{hueA}{HTML}{2E4A6B}
\definecolor{hueAm}{HTML}{5D7293}
\definecolor{hueAl}{HTML}{8DA1B8}
\definecolor{hueAll}{HTML}{C1CCD9}

\definecolor{hueB}{HTML}{3E7A7E}
\definecolor{hueBm}{HTML}{6A9DA1}
\definecolor{hueBl}{HTML}{A8C6C8}

\definecolor{hueC}{HTML}{A47A3D}
\definecolor{hueCl}{HTML}{C9A677}

\definecolor{hueD}{HTML}{6B4E75}
\definecolor{hueDl}{HTML}{9B83A2}

\definecolor{hueE}{HTML}{B76A5B}
\definecolor{hueEl}{HTML}{D29185}

\definecolor{hueF}{HTML}{4E6B52}
\definecolor{hueFl}{HTML}{83A089}

\definecolor{emph}{HTML}{8B3A2E}

\definecolor{inkdark}{HTML}{1F2937}
\definecolor{inkmed}{HTML}{4B5563}
\definecolor{inklight}{HTML}{94A3B8}
\definecolor{bgwarm}{HTML}{F7F4EE}

\colorlet{primary}{hueA}
\colorlet{accentA}{hueB}
\colorlet{accentB}{hueC}
\colorlet{accentC}{hueD}
\colorlet{accentD}{hueE}
\colorlet{accentE}{hueF}

\tikzset{
  >=Stealth,
  font=\sffamily,
  every node/.style={inner sep=0pt, outer sep=0pt},
  arw/.style={->, line width=1.3pt, color=inkmed!75},
  thinarw/.style={->, line width=0.7pt, color=inkmed!60},
  modarw/.style={->, line width=1.3pt, color=hueD!80, dashed},
  lossarw/.style={->, dashed, line width=1pt, color=emph!75},
  imgframe/.style={draw=inkmed!60, line width=0.4pt, inner sep=1pt},
  layerblock/.style={draw, rounded corners=2pt, thick, line width=0.5pt,
                     fill=hueD!12, draw=hueD!70,
                     minimum width=0.9cm, minimum height=0.5cm, inner sep=2pt,
                     font=\tiny\sffamily\color{hueD!35!black}, align=center},
  opblock/.style={draw, rounded corners=2pt, thick, line width=0.5pt,
                  fill=hueD!22, draw=hueD!70,
                  minimum width=0.9cm, minimum height=0.55cm, inner sep=2pt,
                  font=\tiny\sffamily\bfseries\color{hueD!30!black}, align=center},
  pill/.style={draw, rounded corners=4pt, thick, fill=hueC!15, draw=hueC!70,
               minimum width=1.2cm, minimum height=0.45cm, inner sep=2pt,
               font=\scriptsize\sffamily\color{hueC!40!black}},
  twoline-pill/.style={draw, rounded corners=3pt, thick, line width=0.5pt,
                       minimum width=1.25cm, minimum height=0.75cm, inner sep=2pt,
                       align=center, font=\tiny\sffamily},
  headerstrip/.style={rounded corners=3pt, line width=0.7pt},
  sumnode/.style={draw, circle, thick, fill=hueD!15, draw=hueD!75,
                  minimum size=0.7cm, inner sep=0pt,
                  font=\normalsize\bfseries\color{hueD!55!black}},
  frozenborder/.style={hueA!55, line width=1pt, dashed, rounded corners=4pt},
}

\newcommand{\layerstack}[3]{%
  \foreach \i in {0,...,\numexpr#1-1\relax} {
    \pgfmathsetmacro{\yy}{\i*0.30}
    \pgfmathtruncatemacro{\op}{25+\i*7}
    \fill[#2!\op, rounded corners=2pt] (0, \yy) rectangle (#3, \yy+0.24);
    \draw[#2!70, very thin, rounded corners=2pt] (0, \yy) rectangle (#3, \yy+0.24);
  }
}

\title{SDR: Set-Distance Rewards for Radiology Report Generation}

\author{%
  H.Ibrahim Gulluk\thanks{Equal contribution.}\,\,$^{1}$ \\
  \texttt{gulluk@stanford.edu}
  \And
  Max Van Puyvelde$^{*}$\,$^{2, 3}$ \\
  \texttt{maxvpuyv@stanford.edu}
  \And
  Wim Van Criekinge$^{3}$ \\ \texttt{wim.vancriekinge@ugent.be}
  \And
  Olivier Gevaert$^{2}$ \\ \texttt{ogevaert@stanford.edu}
  \AND
  \normalfont
  $^{1}$Department of Electrical Engineering, Stanford University \\
  $^{2}$Department of Biomedical Data Science, Stanford University School of Medicine \\
  $^{3}$Department of Mathematical Modelling, Statistics \& Bioinformatics, Ghent University \\
}

\begin{document}

\maketitle

\begin{abstract}

Reinforcement learning with verifiable rewards has rapidly advanced reasoning in vision--language models. 
However, for chest X-ray report generation, the standard rewards (i.e. exact-match accuracy and step-level processes) are incompatible because the reports consist of unordered and orthogonal findings, rather than a causal reasoning chain. 
We address this gap with a set-based view: each report is split into sentences and embedded by a frozen sentence transformer, yielding unordered embedding sets. 
We propose the use of set-to-set distances between generated and reference embeddings as continuous, permutation-invariant rewards. 
Across two datasets and three vision--language models (Qwen3-VL-2B/4B, Gemma3-4B), post-training with set-to-set distance based rewards via GRPO consistently outperforms supervised fine-tuning and exact-match GRPO on all headline metrics  (BERTScore, RadGraph F1 and CheXbert F1 by average \%6.80, \%7.82 and \%4.45 relative improvements respectively). 
The same set distances also enable test-time best-of-$N$ selection: scoring candidates by their distance to training-report embeddings outperforms random selection on our trained models as well as three closed-source LLMs (Mistral-Small, Gemini-2.5 Flash-Lite, GPT-4o-mini) with on average \%16.4 relative improvement on BERTScore. 
Used as a streaming signal, they support a more efficient form of test-time scaling: pruning low-scoring candidates mid-generation reduces generated tokens by over 50\% while preserving the Findings quality of full best-of-$N$ selection. Together these results establish set-distance rewards as a unified signal for both post-training and
test-time scaling in chest X-ray report generation. Our code is publicly \href{https://anonymous.4open.science/r/Set-Distance-Rewards-CXR-BFDA}{available}.
\end{abstract}

\section{Introduction}

Medical image reports play a central role in clinical workflows, including diagnosis, treatment planning, and patient monitoring. Therefore, improving the efficiency and accuracy of medical image reporting using AI models has attracted increasing attention. Researchers have developed vision–language models for medical image report generation across various imaging modalities \citep{li2025towards, liu2021medical, hamamci2024ct2rep}

Similar to other medical imaging modalities, chest X-ray report generation constitutes a critical component of clinical workflows, as chest radiography is among the most commonly performed and widely accessible imaging techniques in medicine. Assisted report generation systems have the potential to reduce radiologist workload while improving reporting consistency and accuracy. Consequently, chest X-ray report generation using vision–language models has been extensively studied \cite{liu2019clinically, li2023dynamic, endo2021retrieval}.

In addition, recent advances in the reasoning capabilities of both language-only and vision–language models have demonstrated improved performance on complex tasks such as mathematical problem solving, coding, and medical visual question answering \cite{shao2024deepseekmath, luo2024improve, li2023chain, wu2025medreason}. Reinforcement learning–based fine-tuning has shown promising results in further enhancing the reasoning abilities of these models \cite{rafailov2023direct, schulman2017proximal, shao2024deepseekmath}.

Specifically, GRPO has been shown to achieve competitive performance without requiring explicit preferred and non-preferred pairs. However, this type of reward-based reinforcement learning raises an important question regarding the design of the reward function, i.e., how to appropriately reward or penalize a language model based on its generated outputs.
Binary reward functions based on the correctness of the outputs are commonly used in reward design. However, such discrete supervision can induce noise, motivating approaches assigning partial rewards to intermediate steps or the overall generation process even when the final answer is incorrect. This has led to the development of process reward models \cite{khalifa2025process, zhang2025lessons, lightman2024let}.

However, assigning rewards to each step in the reasoning process may not be feasible, as step-level annotations are often unavailable, and verifying each step using external sources can be computationally expensive.
Moreover, in chest X-ray reports, clinicians provide findings that do not necessarily form a causal or sequential structure that can be interpreted as a chain of thought, making step-by-step verification less meaningful. Instead, these findings are often independent of one another and may be presented in an arbitrary order.

To address these challenges, we propose a set distance–based reward formulation. Specifically, we obtain embeddings of sentences from both the ground truth report and the generated report, and compute distances between these two sets of vectors. These distances are then used as a reward signal during GRPO training. In this way, we provide a continuous reward signal that accounts for the unordered and independent nature of chest X-ray findings.

\paragraph{Contributions.}
Our main contributions are summarized as follows:
\begin{itemize}
    \item \textbf{Set-distance reward functions for GRPO post-training.}
        We address the infeasibility of process reward modelling for
        radiology reports by treating each report as an unordered set
        of sentence embeddings and using set-to-set distances (Chamfer
        and Hausdorff over cosine distance) as a continuous, permutation-%
        invariant reward signal during GRPO. Across three vision--language
        backbones and two report-generation
        benchmarks our set-distance reward
        consistently outperforms both the SFT baseline and discrete
        exact-match GRPO post-training on the evaluation metrics.
    \item \textbf{Set-distance--guided test-time response selection.}
        We further use the same family of set distances as a test-time
        scaling / best-of-$N$ selection signal: for each test image we
        sample $K$ candidate reports from the model and pick the one
        whose embedding set is closest to the embedding sets of the
        training reports. This inference-time procedure improves 
        closed-source generalist LLMs (GPT, Gemini, Mistral) 
        over a random-selection baseline,
        averaged over multiple candidates per sample.
    \item \textbf{Distance-based on-the-fly pruning of generations.}
        As an extension of the above, we compute the running set distance
        between the partially generated text and the training distribution
        during inference, and prune candidates whose distance crosses
        a threshold before they are fully generated. This early-stopping
        scheme attains comparable quality with substantially
        fewer generated tokens, demonstrating that the same set-distance
        signal that drives our reward can also be used to lower the
        compute cost of test-time scaling.
\end{itemize}

\section{Related Work}

Medical vision–language models have been proposed to expand the applications of AI in the medical domain. 
MedViLL, a BERT-based model, was introduced in \cite{moon2022multi} and is capable of performing tasks such as medical diagnosis, image–report retrieval, and medical visual question answering. 
Med-Flamingo adapts the Flamingo architecture to medical image–text data for tasks including medical VQA and rationale generation \cite{moor2023med}.

Enhancing reasoning in medical vision-language models has gained attention following the reasoning improvements in general models. 
The MedReason dataset was proposed to enhance this field \cite{wu2025medreason}. 
Med-R1 models are generalist vision-language models trained with reinforcement learning on diverse medical image modalities \cite{lai2026med}. 
Similarly, MedVLM-R1, which is trained with GRPO, increases medical image reasoning \cite{pan2025medvlm}

Beyond medical models, reward design in RL post-training, particularly for GRPO, remains an active area of research. 
While discrete correctness-based rewards have shown strong gains, especially in mathematical reasoning \cite{shao2024deepseekmath}, continuous rewards are being investigated to reduce the noise introduced by binary supervision, since partially correct intermediate reasoning steps may still be valuable even when the final answer is incorrect \cite{khalifa2025process}.

To address these challenges, the authors proposed reasoning-driven process reward modeling \cite{she2025r}. Entropy-Regularized Process Reward Modeling (ER-PRM) \cite{zhang2024entropy} was introduced to add a KL-regularized Markov decision process to ensure that the model remains close to its initial distribution during process reward modeling. EDU-PRM, on the other hand, applies entropy-driven sampling to generate reasoning steps \cite{cao2025more}.

\section{Method}
\label{sec:method}

We first fine-tune vision--language models using SFT on 
chest X-ray reports and then we post-train them via 
Group Relative Policy Optimization (GRPO)~\citep{shao2024deepseekmath}
with a format reward that constrains the output to a structured reasoning template,
and additional set-based semantic rewards that score the 
clinical content of the generated report against the reference report.
The key design choice behind
the semantic reward is to treat each section of a report as an
unordered set of sentence embeddings rather than as a single
sequence, reflecting the observation that individual chest X-ray findings
are permutation-invariant and generally orthogonal rather than forming a
causal chain.

\subsection{Sentence-level report representation}
\label{sec:report-rep}

\begin{figure}[t]
    \centering
    \resizebox{\linewidth}{!}{%
    \begin{tikzpicture}[
        sentbox/.style={draw=inklight!90, rounded corners=2pt, line width=0.4pt,
                        fill=inkmed!4, inner sep=3pt, text width=5.8cm,
                        align=left, font=\footnotesize\color{inkdark}},
        sectbar/.style={line width=2.2pt, line cap=round},
        sectlbl/.style={font=\footnotesize\bfseries\sffamily,
                        inner sep=0pt, anchor=base west},
        stbox/.style={draw=hueA!70, rounded corners=5pt, line width=0.8pt,
                      fill=hueA!8, align=center, minimum width=2.7cm,
                      inner xsep=6pt, inner ysep=8pt},
        sttitle/.style={font=\normalsize\bfseries\sffamily\color{hueA!25!black},
                        align=center, inner sep=0pt},
        stsub/.style={font=\scriptsize\ttfamily\color{inkmed}, inner sep=0pt},
        vecbox/.style={draw=hueC!75, rounded corners=1.5pt, line width=0.5pt,
                       fill=hueC!20, minimum width=0.42cm, minimum height=1.40cm,
                       inner sep=0},
        veclbl/.style={font=\footnotesize\sffamily\color{inkdark}, inner sep=1pt},
        xraycap/.style={font=\footnotesize\sffamily\color{inkmed}, inner sep=1pt},
        frozenpill/.style={draw=hueA!60, rounded corners=2pt, fill=hueA!12,
                           font=\tiny\bfseries\sffamily\color{hueA!40!black},
                           inner xsep=3pt, inner ysep=1.5pt}
    ]

    \coordinate (fOrigin) at (0, 0);
    \node[sectlbl, color=hueA!35!black, anchor=base west]
         at (fOrigin) (fhdr) {Findings};

    \node[sentbox, anchor=north west] at ([yshift=-5pt] fhdr.south west) (f1)
        {The left hemithorax is almost completely opacified, presumedly related to further enlargement of the pleural base carcinoma.};
    \node[sentbox, anchor=north west] at ([yshift=-3pt] f1.south west) (f2)
        {The visualized aerated portion of the left lung shows extensive interstitial and airspace density that could be caused by coexisting atelectasis or pneumonia.};
    \node[sentbox, anchor=north west] at ([yshift=-3pt] f2.south west) (f3)
        {The right lobe is grossly clear, although the lung volume is low.};
    \node[sentbox, anchor=north west] at ([yshift=-3pt] f3.south west) (f4)
        {The heart size may be enlarged.};
    \node[sentbox, anchor=north west] at ([yshift=-3pt] f4.south west) (f5)
        {The tip of the right Mediport catheter is located in the superior vena cava.};

    \draw[sectbar, color=hueA!60]
        ([xshift=-6pt, yshift=4pt] fhdr.base west) --
        ([xshift=-6pt] f5.south west);

    \node[sectlbl, color=hueC!40!black, anchor=base west]
         at ([yshift=-14pt] f5.south west) (ihdr) {Impression};

    \node[sentbox, anchor=north west] at ([yshift=-5pt] ihdr.south west) (i1)
        {Interval worsening of the left pleural base mass, with almost complete opacification of the left hemithorax.};
    \node[sentbox, anchor=north west] at ([yshift=-3pt] i1.south west) (i2)
        {Pneumonia and or atelectasis in the aerated portion of the left lung is not excluded.};
    \node[sentbox, anchor=north west] at ([yshift=-3pt] i2.south west) (i3)
        {Grossly clear right lung.};
    \node[sentbox, anchor=north west] at ([yshift=-3pt] i3.south west) (i4)
        {Possible cardiomegaly.};

    \draw[sectbar, color=hueC!65]
        ([xshift=-6pt, yshift=4pt] ihdr.base west) --
        ([xshift=-6pt] i4.south west);

    \node[imgframe, anchor=east] (xray)
        at ($(f1.north west)!0.5!(i4.south west) + (-0.80cm,0)$)
        {\includegraphics[width=3.8cm]{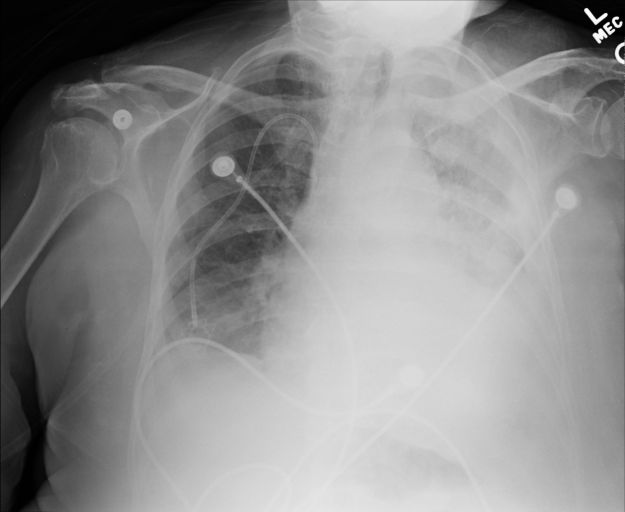}};
    \node[xraycap, anchor=north] at ([yshift=-5pt] xray.south) {Chest X-ray};

    \node[stbox, minimum height=8.9cm] (st)
        at ($(f1.east)!0.5!(i4.east) + (3.1cm,0)$) {};

    \node[sttitle, anchor=north] at ([yshift=-0.35cm] st.north)
        {Sentence\\[-1pt]Transformer};

    \node[frozenpill, anchor=north]
          at ([yshift=-1.15cm] st.north) {FROZEN};

    \begin{scope}[shift={($(st.center) + (-0.65cm,-0.30cm)$)}]
        \layerstack{5}{hueA}{1.30}
    \end{scope}

    \node[stsub, anchor=south] at ([yshift=0.30cm] st.south)
        {all-mpnet-base-v2};

    \foreach \s in {f1,f2,f3,f4,f5,i1,i2,i3,i4}{
        \draw[arw] (\s.east) -- (\s.east -| st.west);
    }

    \foreach \s/\idx/\sec in {%
        f1/1/F, f2/2/F, f3/3/F, f4/4/F, f5/5/F,
        i1/1/I, i2/2/I, i3/3/I, i4/4/I%
    }{
        \node[vecbox] (v\s) at ([xshift=2.0cm] st.east |- \s) {};
        \foreach \k in {1,2,3,4,5,6}{
            \draw[hueC!55, very thin]
                ([yshift=-0.2cm*\k] v\s.north west) --
                ([yshift=-0.2cm*\k] v\s.north east);
        }
        \draw[arw] (st.east |- \s) -- (v\s.west);
        \node[veclbl, right=3pt of v\s] {$\mathbf{e}^{\sec}_{\idx}$};
    }

    \coordinate (setBraceTop) at ($(vf1.north east) + (1.15cm,0)$);
    \coordinate (setBraceBot) at (setBraceTop |- vi4.south east);

    \draw[decorate, decoration={brace, amplitude=5pt}, line width=0.7pt,
          color=inkmed!85]
         (setBraceTop) -- (setBraceBot)
         node[midway, right=9pt, align=left,
              font=\footnotesize\sffamily\color{inkdark}]
         {$\mathcal{E}(r)$\\[3pt]
          $\subset \mathbb{R}^{d}$};

    \end{tikzpicture}%
    }
    \caption{\textbf{Sentence-level encoding of a chest X-ray report.}
    Each study pairs a radiograph with a free-text report composed of a
    Findings and an Impression section. We split both
    sections into individual sentences and embed each sentence
    independently with the frozen pre-trained \texttt{all-mpnet-base-v2}
    sentence transformer, producing one $d$-dimensional vector per
    sentence. The resulting unordered collection of sentence
    embeddings
    $\mathcal{E}(r)=\{\mathbf{e}^{F}_{1},\dots,\mathbf{e}^{F}_{5},\mathbf{e}^{I}_{1},\dots,\mathbf{e}^{I}_{4}\}\subset\mathbb{R}^{d}$
    serves as the ground-truth report representation in our set-reward
    pipeline.}
    \label{fig:report-encoding}
\end{figure}
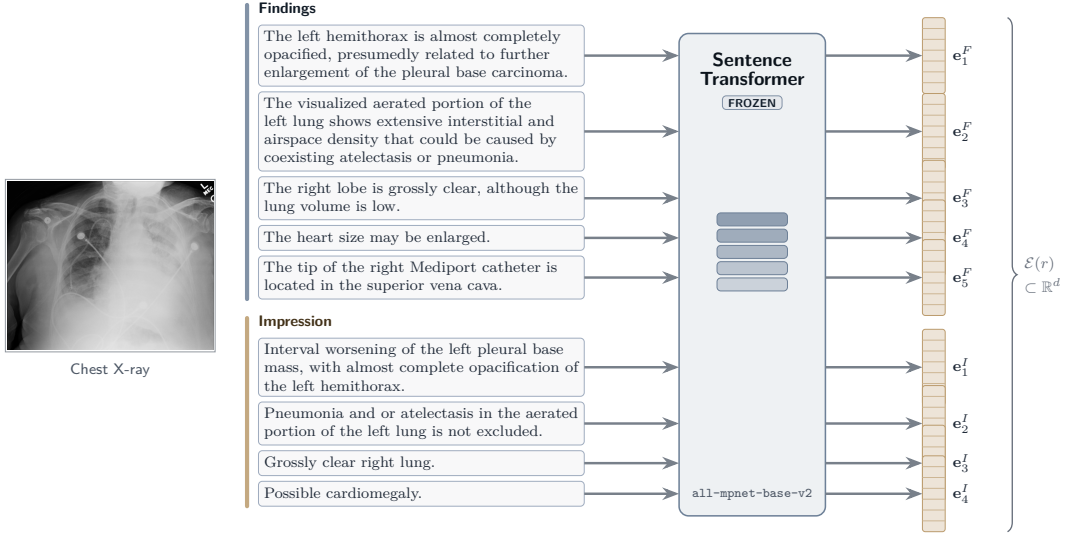

A chest X-ray report $y$ consists of two sections, a Findings
section $y^{F}$ and an Impression section $y^{I}$. We split each
section into individual sentences using a standard sentence segmenter,
yielding $y^{F} = (s^{F}_{1},\dots,s^{F}_{n_{F}})$ and
$y^{I} = (s^{I}_{1},\dots,s^{I}_{n_{I}})$, where the sentence counts
$n_{F}, n_{I} \in \mathbb{N}$ may vary across studies. Each sentence $s$ is
mapped to a fixed-dimensional semantic embedding
$\mathbf{e} = E_{\phi}(s) \in \mathbb{R}^{d}$ by a frozen pre-trained
sentence transformer $E_{\phi}$ (specifically \texttt{all-mpnet-base-v2} \cite{reimers-2019-sentence-bert}).
The report is then represented by the two sets of embeddings
\begin{equation}
    \mathcal{E}^{F}(y)
    \;=\; \bigl\{\, E_{\phi}(s^{F}_{i}) \,:\, 1 \le i \le n_{F} \,\bigr\},
    \qquad
    \mathcal{E}^{I}(y)
    \;=\; \bigl\{\, E_{\phi}(s^{I}_{j}) \,:\, 1 \le j \le n_{I} \,\bigr\},
    \label{eq:report-sets}
\end{equation}
both subsets of $\mathbb{R}^{d}$ (Figure~\ref{fig:report-encoding}).
Because $\mathcal{E}^{F}(y)$ and $\mathcal{E}^{I}(y)$ are sets, they are
invariant under permutations of the underlying sentences, which captures
the clinical intuition that the listing order of individual findings
carries no diagnostic meaning. Throughout this section the unhatted
symbol $y$ denotes the ground-truth reference report paired with the
input X-ray, and $\hat{y}$ denotes a generation produced by the model.

\paragraph{Format reward.}
To make the generated report easily parseable, we require the output
$\hat{y}$ to follow the template
\texttt{<think>}\,$\hat{y}^{F}$\,\texttt{</think>\,<answer>}\,$\hat{y}^{I}$\,\texttt{</answer>},
where $\hat{y}^{F}$ and $\hat{y}^{I}$ are the generated Findings
and Impression respectively, with each tag occurring exactly
once, in the specified order, and enclosing non-empty content. The
format reward is the binary indicator
$R_{\mathrm{fmt}}(\hat{y})=\mathrm{valid}(\hat{y})\in\{0,1\}$; when
$R_{\mathrm{fmt}}=1$ the two section strings can be extracted
unambiguously from $\hat{y}$ and fed into the semantic reward defined
below.

\subsection{Set-based semantic reward}
\label{sec:semantic-reward}

Given a generated report $\hat{y}$ whose sections $\hat{y}^{F}$ and
$\hat{y}^{I}$ have been extracted according to the template above, we
form their sets of sentence embeddings $\mathcal{E}^{F}(\hat{y})$ and
$\mathcal{E}^{I}(\hat{y})$ using exactly the same encoder $E_{\phi}$ as
for the reference report in Eq.~\eqref{eq:report-sets}. This yields, for
each training example, two pairs of embedding sets,
\begin{equation}
    \bigl(\mathcal{E}^{F}(\hat{y}),\,\mathcal{E}^{F}(y)\bigr),
    \qquad
    \bigl(\mathcal{E}^{I}(\hat{y}),\,\mathcal{E}^{I}(y)\bigr),
    \label{eq:section-pairs}
\end{equation}
each consisting of one generated and one reference set of vectors in
$\mathbb{R}^{d}$.

Let
\[
    \mathcal{D}\,:\,
    2^{\mathbb{R}^{d}} \times 2^{\mathbb{R}^{d}}
    \;\longrightarrow\; [0,1]
\]
be any symmetric set-to-set distance normalised to the unit interval, that
is, $\mathcal{D}(\mathcal{A},\mathcal{B}) =
\mathcal{D}(\mathcal{B},\mathcal{A})$ and
$\mathcal{D}(\mathcal{A},\mathcal{A}) = 0$ for all non-empty finite
$\mathcal{A},\mathcal{B}\subset\mathbb{R}^{d}$. Concrete choices for
$\mathcal{D}$ are described in Section~\ref{sec:distances}. For each
section $S \in \{F, I\}$ we define the section-level semantic reward as
the similarity induced by $\mathcal{D}$,
\begin{equation}
    R^{S}_{\mathrm{sem}}(\hat{y},\,y)
    \;=\;
    1 \;-\;
    \mathcal{D}\!\bigl(\mathcal{E}^{S}(\hat{y}),\,\mathcal{E}^{S}(y)\bigr)
    \;\in\; [0,1],
    \qquad
    S \in \{F, I\}.
    \label{eq:r-sem-section}
\end{equation}
The total set-based semantic reward is the sum of the two section-level
rewards,
\begin{equation}
    R_{\mathrm{sem}}(\hat{y},\,y)
    \;=\;
    R^{F}_{\mathrm{sem}}(\hat{y},y)
    \;+\;
    R^{I}_{\mathrm{sem}}(\hat{y},y)
    \;\in\; [0,2].
    \label{eq:r-sem}
\end{equation}

We instantiate $\mathcal{D}$ with set-to-set distances built on top of
the cosine distance between unit-norm sentence embeddings: for two
embeddings $\mathbf{u},\mathbf{v}\in\mathbb{R}^{d}$ we use
$d(\mathbf{u},\mathbf{v}) = \tfrac{1}{2}(1 - \mathbf{u}^{\top}\mathbf{v})
\in [0,1]$, which is well matched to the unit-normalised outputs of
$E_{\phi}$. Throughout the rest of this subsection we use
$\hat{\mathbf{e}}\in\mathcal{E}^{S}(\hat{y})$ to denote the embedding of
a sentence in the generated section $\hat{y}^{S}$ and
$\mathbf{e}\in\mathcal{E}^{S}(y)$ for that of a sentence in the matched
reference section $y^{S}$, with cardinalities
$n=|\mathcal{E}^{S}(\hat{y})|$ and $m=|\mathcal{E}^{S}(y)|$. With this
notation, for each section $S\in\{F,I\}$, the \emph{Chamfer} set
distance averages the nearest-neighbour cost in each direction,
\begin{equation}
    \mathcal{D}_{\mathrm{Chamfer}}\!\bigl(\mathcal{E}^{S}(\hat{y}),\,\mathcal{E}^{S}(y)\bigr)
    \;=\;
    \tfrac{1}{2}\!\left(
        \frac{1}{n}\!\sum_{\hat{\mathbf{e}}\,\in\,\mathcal{E}^{S}(\hat{y})}
            \min_{\mathbf{e}\,\in\,\mathcal{E}^{S}(y)} d(\hat{\mathbf{e}},\mathbf{e})
        \;+\;
        \frac{1}{m}\!\sum_{\mathbf{e}\,\in\,\mathcal{E}^{S}(y)}
            \min_{\hat{\mathbf{e}}\,\in\,\mathcal{E}^{S}(\hat{y})} d(\hat{\mathbf{e}},\mathbf{e})
    \right),
    \label{eq:chamfer-method}
\end{equation}
so that the first term rewards every generated sentence for matching
some reference sentence and the second term penalises reference
sentences not covered by any generated sentence. The \emph{Hausdorff}
set distance replaces the means with maxima, returning the worst-case
uncovered sentence in either direction,
\begin{equation}
    \mathcal{D}_{\mathrm{Hausdorff}}\!\bigl(\mathcal{E}^{S}(\hat{y}),\,\mathcal{E}^{S}(y)\bigr)
    \;=\;
    \max\!\left\{
        \max_{\hat{\mathbf{e}}\,\in\,\mathcal{E}^{S}(\hat{y})}
            \min_{\mathbf{e}\,\in\,\mathcal{E}^{S}(y)} d(\hat{\mathbf{e}},\mathbf{e}),\;
        \max_{\mathbf{e}\,\in\,\mathcal{E}^{S}(y)}
            \min_{\hat{\mathbf{e}}\,\in\,\mathcal{E}^{S}(\hat{y})} d(\hat{\mathbf{e}},\mathbf{e})
    \right\},
    \label{eq:hausdorff-method}
\end{equation}
which makes it strictly harsher than Chamfer so that a single uncovered
sentence on either side is enough to dominate the reward. 
Considering that we use these rewards in post-training, 
meaning the model is already trained with SFT, 
the Hausdorff-based reward can be seen as penalizing the model 
when it generates outliers or contradictory observations
Both metrics are symmetric, invariant to the orderings of $\mathcal{E}^{S}(\hat{y})$
and $\mathcal{E}^{S}(y)$ by construction, and lie in $[0,1]$ since the
base distance $d$ does. Concretely,
$1 - \mathcal{D}_{\mathrm{Chamfer}}$ behaves like a soft coverage
reward -- it grows whenever any additional generated sentence
finds a close reference match -- whereas
$1 - \mathcal{D}_{\mathrm{Hausdorff}}$ behaves like a worst-case
recall reward that only saturates once every clinically
relevant finding has been mentioned. 
Empirically we find that these two choices are the most useful set
rewards for GRPO post-training, and we report results
for them throughout Sec.~\ref{sec:grpo-experiments}.

\paragraph{Combined reward.}
The final scalar reward passed to GRPO is the weighted sum
$R(\hat{y},y) = \lambda_{\mathrm{fmt}}\, R_{\mathrm{fmt}}(\hat{y}) +
\lambda_{\mathrm{sem}}\, R_{\mathrm{sem}}(\hat{y},y)$ with non-negative
weights $\lambda_{\mathrm{fmt}},\lambda_{\mathrm{sem}} \ge 0$. When
$R_{\mathrm{fmt}}(\hat{y}) = 0$ the section strings cannot be reliably
extracted, and we set $R_{\mathrm{sem}}(\hat{y},y):=0$ so that a
malformed generation is penalised through both reward components
simultaneously. 
In the experiments, we simply set $\lambda_{\mathrm{fmt}}=\lambda_{\mathrm{sem}}=1$.

\subsection{Inference-time response selection}
\label{sec:response-selection}

\begin{figure}[t]
    \centering
    \begin{subfigure}[b]{0.42\linewidth}
        \centering
        \resizebox{\linewidth}{!}{
        \begin{tikzpicture}[
            pred_dot/.style={circle, draw=hueC!85, fill=hueC!55,
                              minimum size=7.5pt, inner sep=0,
                              line width=0.7pt},
            ref_dot/.style={circle, draw=hueA!85, fill=hueA!70,
                             minimum size=5.6pt, inner sep=0,
                             line width=0.45pt},
            pred_region/.style={draw=hueC!60, fill=hueC!10,
                                 line width=0.55pt},
            ref_region/.style={draw=hueA!55, fill=hueA!10,
                                line width=0.55pt},
            pdistline/.style={dashed, color=emph!55,
                               line width=0.55pt,
                               shorten <=2pt, shorten >=2pt},
            plabel/.style={font=\scriptsize\sffamily\color{hueC!30!black},
                            inner sep=1pt},
            rlabel/.style={font=\scriptsize\sffamily\color{hueA!40!black},
                            inner sep=1pt},
            setlabel/.style={font=\footnotesize\bfseries\sffamily,
                              inner sep=2pt},
            dlabel/.style={font=\tiny\sffamily\color{inkdark},
                            inner sep=1.2pt, fill=bgwarm,
                            rounded corners=1.2pt,
                            draw=inkmed!20, line width=0.25pt}
        ]
            \draw[pred_region] (-1.6, 0) ellipse (1.4 and 1.55);
            \node[setlabel, color=hueC!30!black, anchor=south]
                  at (-1.6, 1.6) {$\mathcal{E}^{F}(\hat{y})$};

            \draw[ref_region] (1.7, 0) ellipse (1.6 and 1.7);
            \node[setlabel, color=hueA!40!black, anchor=south]
                  at (1.7, 1.75) {$\mathcal{E}^{F}(y)$};

            \node[pred_dot] (e1) at (-2.2,  0.6) {};
            \node[plabel,  left=1.5pt of e1] {$\hat{\mathbf{e}}_1$};
            \node[pred_dot] (e2) at (-1.5, -0.6) {};
            \node[plabel,  below=1.5pt of e2] {$\hat{\mathbf{e}}_2$};
            \node[pred_dot] (e3) at (-0.7,  0.3) {};
            \node[plabel,  right=1.5pt of e3] {$\hat{\mathbf{e}}_3$};

            \node[ref_dot] (r1) at (1.0,  1.0) {};
            \node[rlabel, above=1.5pt of r1] {$\mathbf{e}_1$};
            \node[ref_dot] (r2) at (1.4, -0.9) {};
            \node[rlabel, below=1.5pt of r2] {$\mathbf{e}_2$};
            \node[ref_dot] (r3) at (2.6,  0.5) {};
            \node[rlabel, right=1.5pt of r3] {$\mathbf{e}_3$};
            \node[ref_dot] (r4) at (2.1, -0.2) {};
            \node[rlabel, below=1.5pt of r4] {$\mathbf{e}_4$};

            \draw[pdistline] (e1) -- (r1);
            \draw[pdistline] (e2) -- (r2);
            \draw[pdistline] (e3) -- (r1);
            \draw[pdistline] (e3) -- (r3);
            \draw[pdistline] (e3) -- (r4);

            \node[dlabel] at ($(e2)!0.55!(r2) + (0,-0.18)$)
                {$d(\hat{\mathbf{e}}_2,\mathbf{e}_2)$};
        \end{tikzpicture}
        }
        \caption{Set-distance computation between two embedding sets.
        Each dot is one sentence embedding; the two ellipses are the
        full embedding sets of a generated report (amber) and the
        matched reference report (slate). Dashed lines are the
        nearest-neighbour cosine distances that Chamfer / Hausdorff
        aggregate.(Eqs.~\eqref{eq:chamfer-method} and
    \eqref{eq:hausdorff-method})}
        \label{fig:set-distance-illustration}
    \end{subfigure}\hfill
    \begin{subfigure}[b]{0.56\linewidth}
        \centering
        \resizebox{\linewidth}{!}{
        \begin{tikzpicture}[
            train/.style={circle, draw=hueA!65, fill=hueA!45,
                          minimum size=3.4pt, inner sep=0,
                          line width=0.2pt},
            trainNN/.style={circle, draw=hueA!85, fill=hueA!70,
                            minimum size=5.2pt, inner sep=0,
                            line width=0.45pt},
            cand/.style={circle, draw=hueC!85, fill=hueC!55,
                         minimum size=7.5pt, inner sep=0,
                         line width=0.7pt},
            sel/.style={star, star points=5, star point ratio=2.3,
                        draw=emph!85, fill=emph!45,
                        minimum size=13pt, inner sep=0,
                        line width=0.9pt},
            distline/.style={dashed, color=inkmed!70, line width=0.55pt,
                             shorten <=2pt, shorten >=2pt},
            selline/.style={color=emph!85, line width=1.3pt,
                            shorten <=2pt, shorten >=2pt},
            glabel/.style={font=\scriptsize\sffamily\color{hueC!30!black},
                           inner sep=1pt},
            slabel/.style={font=\footnotesize\bfseries\sffamily\color{emph!80!black},
                           inner sep=1.5pt},
            dlabel/.style={font=\tiny\sffamily\color{inkdark},
                           inner sep=1.5pt, fill=bgwarm,
                           rounded corners=1.2pt,
                           draw=inkmed!20, line width=0.25pt},
            sdlabel/.style={font=\tiny\bfseries\sffamily\color{emph!80!black},
                            inner sep=1.5pt, fill=bgwarm,
                            rounded corners=1.2pt,
                            draw=emph!40, line width=0.35pt},
            panel/.style={draw=inkmed!25, fill=bgwarm!70,
                          rounded corners=5pt, line width=0.5pt},
            spacelbl/.style={font=\scriptsize\itshape\sffamily\color{inkmed},
                             inner sep=3pt},
            trainregion/.style={draw=hueA!45, line width=0.5pt, dashed,
                                fill=hueA!6, rounded corners=12pt},
            legend/.style={font=\footnotesize\sffamily\color{inkdark},
                           inner sep=2pt}
        ]

        \node[panel, minimum width=9.8cm, minimum height=4.5cm] (P)
            at (-1.85, 0) {};
        \node[spacelbl, anchor=north west]
             at ([xshift=4pt, yshift=-4pt] P.north west)
             {Embedding space $\mathbb{R}^{d}$ \,(2D schematic)};

        \draw[trainregion] (-0.6, 0.2) ellipse (2.3 and 1.55);

        \foreach \x/\y in {
            -0.7/-0.2,  -0.2/0.5,  -1.2/0.4,  0.0/0.9,  -0.4/0.0,
            -1.6/0.7,   0.3/0.6,  -0.8/1.1,  -0.3/0.2, -1.1/0.1,
            -0.9/0.8,   0.1/0.2,  -1.4/0.0, -0.6/0.6, -0.1/1.2,
            -1.0/0.3,  -0.4/0.8, -1.3/1.0, -0.2/-0.2,-1.7/0.3,
            -0.7/0.4,   0.2/0.7,  -0.9/-0.1,-0.5/1.0, -1.5/0.8,
             0.0/0.3,  -1.1/0.6, -0.3/1.1, -0.8/0.2,  0.3/1.0,
            -1.2/-0.2, -0.1/0.6, -0.6/0.0, -1.6/0.5, -0.2/0.4,
            -1.0/1.2,   0.1/0.8, -1.4/0.2, -0.5/0.1, -0.7/0.9,
            -1.8/0.6,   0.4/0.4, -0.3/-0.4,-1.3/-0.4
        }{
            \node[train] at (\x,\y) {};
        }

        \coordinate (nn1) at (-1.80,  0.60);
        \coordinate (nn2) at (-1.70,  0.30);
        \coordinate (nn3) at (-1.30, -0.40);
        \coordinate (nn4) at (-1.60,  0.70);
        \coordinate (nns) at (-1.40,  0.20);

        \node[cand] (g1) at (-4.9,  1.2) {};
        \node[glabel, above=1pt of g1] {$\hat{y}^{(1)}$};

        \node[cand] (g2) at (-4.9, -0.5) {};
        \node[glabel, left=1pt of g2] {$\hat{y}^{(2)}$};

        \node[cand] (g3) at (-4.3, -1.4) {};
        \node[glabel, below=1pt of g3] {$\hat{y}^{(3)}$};

        \node[cand] (g4) at (-3.9,  0.7) {};
        \node[glabel, above=1pt of g4] {$\hat{y}^{(4)}$};

        \node[sel] (gs) at (-2.95, 0.10) {};
        \node[slabel, below=3pt of gs] {$\hat{y}^{\star}$};

        \node[trainNN] at (nn1) {};
        \node[trainNN] at (nn2) {};
        \node[trainNN] at (nn3) {};
        \node[trainNN] at (nn4) {};
        \node[trainNN] at (nns) {};

        \draw[distline] (g1) -- (nn1);
        \draw[distline] (g2) -- (nn2);
        \draw[distline] (g3) -- (nn3);
        \draw[distline] (g4) -- (nn4);


        \draw[selline] (gs) -- (nns);

        \begin{scope}[shift={(-6.30, -2.55)}]
            \node[train] (lt) at (0,0) {};
            \node[legend, right=5pt of lt] (lttxt)
                  {training reports $\{r^{(t)}\}_{t=1}^{N}$};

            \node[cand, right=0.7cm of lttxt] (lc) {};
            \node[legend, right=5pt of lc] (lctxt)
                  {candidate $\hat{y}^{(k)}\!\sim\!\pi(\cdot\mid x)$};

            \node[sel] (ls) at (0,-0.65) {};
            \node[legend, right=5pt of ls] (lstxt)
                  {selected $\hat{y}^{\star}=\arg\min_k \mathfrak{D}(\hat{y}^{(k)})$};
        \end{scope}

        \end{tikzpicture}
        }
        \caption{Inference-time response selection in embedding space.
        Each dot is itself an embedding \emph{set} of the form drawn in
        (a): amber dots are candidates $\hat{y}^{(k)}$, small slate
        dots are training reports $r^{(t)}$, and the rust star is the
        selected response $\hat{y}^{\star}$.}
        \label{fig:response-selection}
    \end{subfigure}
    \caption{\textbf{Set distances and inference-time response
    selection.} \textbf{(a)} A single set-distance computation between
    a generated report (three sentence embeddings) and a
    matched reference report (four sentence embeddings) visualization. \textbf{(b)} The full inference-time
    selection pipeline of Eq.~\eqref{eq:selection-rule}.}
    \label{fig:method-overview}
\end{figure}
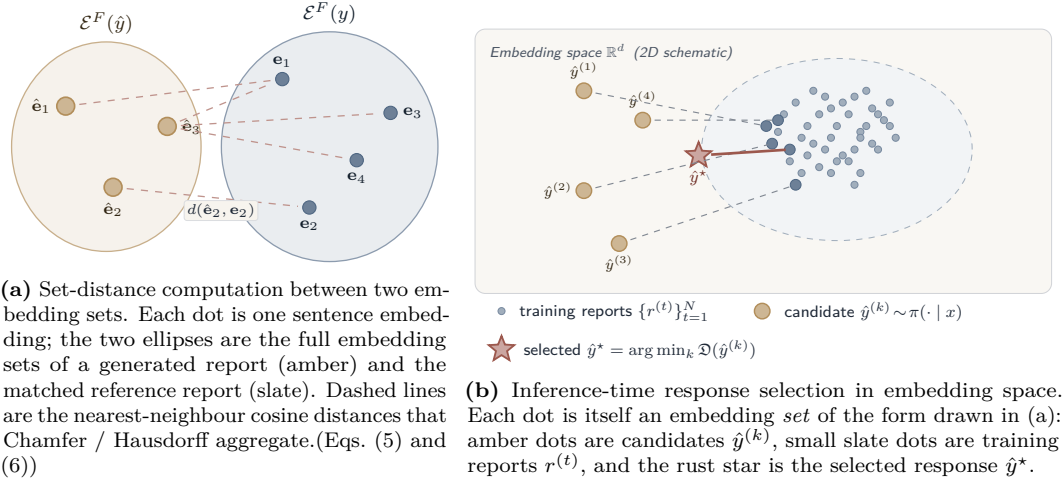

The set-based distance metrics introduced in
Section~\ref{sec:distances} are not restricted to reward design during
GRPO fine-tuning; they can equally well serve as a purely
\emph{inference-time} selection criterion that does not require any
gradient updates. In this section we describe a simple but effective
test-time pipeline in which a generative policy $\pi$ — either our
GRPO-fine-tuned VLM or a frozen closed-source generalist LLM such as
GPT-4o or Gemini — is queried $K$ times on the same chest X-ray, and the
candidate report that is most consistent with the training distribution of
real radiology reports is retained.

Concretely, let $\mathcal{T} = \{\,r^{(t)}\,\}_{t=1}^{N}$ denote the
training corpus, with each report $r^{(t)}$ represented by its two section
embedding sets $\mathcal{E}^{F}(r^{(t)})$ and $\mathcal{E}^{I}(r^{(t)})$ as
defined in Section~\ref{sec:report-rep}. For a chest X-ray $x$ at test
time, we draw $K$ independent generations
$\hat{y}^{(1)},\,\hat{y}^{(2)},\,\dots,\,\hat{y}^{(K)}
    \;\overset{\mathrm{i.i.d.}}{\sim}\;
    \pi(\,\cdot \mid x\,)$
extract the Findings and Impression sections from each
$\hat{y}^{(k)}$ (via the \texttt{<think>}/\texttt{<answer>} template of
Sec.~\ref{sec:method}) and compute their
sentence-embedding sets $\mathcal{E}^{S}(\hat{y}^{(k)})$ for
$S \in \{F, I\}$.

\paragraph{Distance from a generation to the training distribution.}
Since $\mathcal{T}$ is itself a set of sets of embeddings,
scoring a single candidate against $\mathcal{T}$ requires a second
aggregation on top of the set-to-set metric
$\mathcal{D}(\cdot,\cdot)$ of Section~\ref{sec:distances}. We first
compute, for each section $S \in \{F, I\}$ and each training report $t$,
the per-report distance
\begin{equation}
    \mathfrak{D}_{S,\,t}(\hat{y})
    \;=\;
    \mathcal{D}\!\bigl(\mathcal{E}^{S}(\hat{y}),\,\mathcal{E}^{S}(r^{(t)})\bigr),
    \qquad t = 1,\dots,N.
    \label{eq:per-train-dist}
\end{equation}
The distance from the generation $\hat{y}$ to the training distribution
$\mathcal{T}$ is then any of the following three aggregations,
\begin{equation}
    \mathfrak{D}^{\min}_{S}(\hat{y}) = \min_{1 \le t \le N} \mathfrak{D}_{S,\,t}(\hat{y}),
    \quad
    \mathfrak{D}^{\mathrm{avg}}_{S}(\hat{y}) = \frac{1}{N}\sum_{t=1}^{N} \mathfrak{D}_{S,\,t}(\hat{y}),
    \quad
    \mathfrak{D}^{\mathrm{kNN}}_{S}(\hat{y}) = \frac{1}{k}\!\sum_{t \in \mathcal{N}_{k}(\hat{y})}\! \mathfrak{D}_{S,\,t}(\hat{y}),
    \label{eq:aggregations}
\end{equation}
where $\mathcal{N}_{k}(\hat{y}) \subseteq \{1,\dots,N\}$ indexes the $k$
training reports with the smallest $\mathfrak{D}_{S,\,t}(\hat{y})$. Each aggregation
encodes a different notion of ``closeness to the training
distribution'': $\mathfrak{D}^{\min}$ asks whether $\hat{y}$ resembles \emph{any}
single real report, $\mathfrak{D}^{\mathrm{avg}}$ scores against the whole corpus
uniformly, and $\mathfrak{D}^{\mathrm{kNN}}$ is a noise-robust middle ground.

\paragraph{Selection rule.}
The total distance of a candidate to the training distribution is the sum
of its two section-level distances,
\begin{equation}
    \mathfrak{D}(\hat{y})
    \;=\;
    \mathfrak{D}_{F}(\hat{y})
    \;+\;
    \mathfrak{D}_{I}(\hat{y}),
    \label{eq:total-train-dist}
\end{equation}
matching the additive structure of the semantic reward in
Eq.~\eqref{eq:r-sem}. The selected response is the candidate with the
smallest such distance,
\begin{equation}
    \hat{y}^{\star}
    \;=\;
    \operatorname*{arg\,min}_{k \in \{1,\dots,K\}}\; \mathfrak{D}\bigl(\hat{y}^{(k)}\bigr).
    \label{eq:selection-rule}
\end{equation}
Any of the distance metrics $\mathcal{D}$ from Section~\ref{sec:distances}
can be plugged into Eq.~\eqref{eq:per-train-dist} and combined with any of
the three aggregations of Eq.~\eqref{eq:aggregations}, yielding a large
family of selection policies whose relative merits we study empirically
in Section~\ref{sec:response-selection}.

As the pipeline is purely inference-time, no parameter of the generation
policy $\pi$ is updated, and the only training-time artefact it relies on
is the pre-computed set
$\{\,\mathcal{E}^{S}(r^{(t)})\,\}_{t,S}$ of reference embeddings, which
depends only on the frozen sentence encoder $E_{\phi}$ and can be cached
on disk once per training corpus. This makes the approach applicable to 
closed-source generalist LLMs (GPT-4o, Gemini) for which
gradients are unavailable and domain-specific fine-tuning is not
possible at all. As generalist LLMs are known to
hallucinate clinically implausible findings, drift off-topic, or produce
phrasings that are absent from genuine radiology practice. By preferring,
among $K$ stochastic candidates, the one whose sentence embeddings fall
into regions of $\mathbb{R}^{d}$ that are closest to the real reports
in $\mathcal{T}$, the selection rule in Eq.~\eqref{eq:selection-rule}
exploits the training corpus as a gradient-free prior on clinically
plausible output. Although our method adds additional cost of 
sentence embedding and distance calculations, this can be done in
parallel without a need for GPU.
We report, in Section~\ref{sec:response-selection}, the
performance gain that this simple test-time procedure yields 
closed-source generalist LLMs, across every
combination of set-distance metric
$\mathcal{D} \in \{\mathrm{Chamfer},\,\mathrm{Hausdorff},\,\mathrm{OT},\,
\mathrm{POT}\}$ and
aggregation in $\{\textsc{min},\,\textsc{avg},\,\textsc{kNN}\}$. 

In addition to known set-distance metrics, 
We derive hungarian matching based set-distances with additional modifications 
on the samples that are not matched during hungarian matching, that are
denotes as $\mathrm{Hung\text{-}NN},\,\mathrm{Hung\text{-}Pen}$. 
The details are provided in Section \ref{sec:distances}.


\section{Experiments}
\label{sec:grpo-experiments}

Throughout the experiments we use MIMIC-CXR and RexGradient datasets 
\cite{PhysioNet-mimic-cxr-2.1.0, zhang2025rexgradient}.
We fine-tune (SFT) and then post-train three 
vision--language models -- Qwen3-VL-2B, Qwen3-VL-4B and
Gemma3-4B \cite{yang2025qwen3, team2024gemma}-- on the chest X-ray report generation task with GRPO 
via the reward configurations we described. 
Throughout this section we abbreviate them as:
\textsc{SFT} (the supervised fine-tuned checkpoint without GRPO);
$R_{\mathrm{fmt}}$ (the format reward alone, i.e.\ the binary template
indicator of Sec.~\ref{sec:method});
$R_{\mathrm{exact}}$ (format and a discrete exact-match accuracy reward);
and our two proposed set-based semantic rewards combined with the format
reward, $R_{\mathrm{Cham}}$ and $R_{\mathrm{Haus}}$ (Chamfer and Hausdorff
set distances respectively, defined in App.~\ref{sec:distances}).
Tabs.~\ref{tab:grpo-rexgradient-findings} and
\ref{tab:grpo-mimic-findings} report mean test-set scores over 5
random seeds on headline metrics covering embedding-based
(BERTScore F$_1$, COMET) and clinical (RadGraph averaged F$_1$,
CheXbert macro F$_1$) families on the Findings section. The
final Mean block in each table averages each (reward, metric)
cell across the models in that dataset. We observe that Chamfer based 
rewarding produce the highest performance across all trained models and all evaluation metrics 
for RexGradient dataset. For MIMIC-CXR on the other hand, both Chamfer and Hausdorff based rewardings 
reaching the highest performances. These results suggest that, in addition to Supervised Finetuning, 
Set-Distance Based Rewarding improves the performance of report generation drastically.
Additional results including impression section results and 
more evaluation metrics are provided in App.~\ref{sec:grpo-appendix}.
Experimental setup is provided in Appendix ~\ref{sec:experimental-setup}

\begin{table}[t]
\caption{\textbf{GRPO post-training results on ReXGradient (Findings).} Results for different reward functions are provided. Chamfer Distance based rewarding outperforms all the other methods on average for all the evaluation metrics.}
\label{tab:grpo-rexgradient-findings}
\centering
\footnotesize
\setlength{\tabcolsep}{3pt}
\renewcommand{\arraystretch}{0.95}
\begin{tabular}{llccccc}
\toprule
Model & Metric & SFT & $R_{\mathrm{fmt}}$ & $R_{\mathrm{exact}}$ & $R_{\mathrm{Cham}}$ & $R_{\mathrm{Haus}}$ \\
\midrule
\multirow{4}{*}{Qwen3-VL-2B} & BERTScore F1 & 0.339\,{\scriptsize $\pm$0.004} & 0.342\,{\scriptsize $\pm$0.002} & 0.332\,{\scriptsize $\pm$0.002} & \textbf{0.367}\,{\scriptsize $\pm$0.002} & 0.339\,{\scriptsize $\pm$0.003} \\
 & COMET & 0.619\,{\scriptsize $\pm$0.002} & 0.620\,{\scriptsize $\pm$0.001} & 0.619\,{\scriptsize $\pm$0.001} & \textbf{0.657}\,{\scriptsize $\pm$0.001} & 0.623\,{\scriptsize $\pm$0.001} \\
 & RadGraph F1 & 0.247\,{\scriptsize $\pm$0.006} & 0.251\,{\scriptsize $\pm$0.003} & 0.244\,{\scriptsize $\pm$0.002} & 0.224\,{\scriptsize $\pm$0.003} & \textbf{0.258}\,{\scriptsize $\pm$0.003} \\
 & CheXbert F1 & 0.705\,{\scriptsize $\pm$0.003} & 0.708\,{\scriptsize $\pm$0.007} & 0.663\,{\scriptsize $\pm$0.008} & \textbf{0.734}\,{\scriptsize $\pm$0.000} & 0.686\,{\scriptsize $\pm$0.009} \\
\midrule
\multirow{4}{*}{Qwen3-VL-4B} & BERTScore F1 & 0.332\,{\scriptsize $\pm$0.002} & 0.333\,{\scriptsize $\pm$0.003} & 0.331\,{\scriptsize $\pm$0.002} & \textbf{0.351}\,{\scriptsize $\pm$0.000} & 0.333\,{\scriptsize $\pm$0.002} \\
 & COMET & 0.620\,{\scriptsize $\pm$0.001} & 0.618\,{\scriptsize $\pm$0.001} & 0.617\,{\scriptsize $\pm$0.001} & \textbf{0.677}\,{\scriptsize $\pm$0.000} & 0.572\,{\scriptsize $\pm$0.001} \\
 & RadGraph F1 & 0.237\,{\scriptsize $\pm$0.004} & 0.239\,{\scriptsize $\pm$0.004} & 0.242\,{\scriptsize $\pm$0.001} & \textbf{0.282}\,{\scriptsize $\pm$0.000} & 0.242\,{\scriptsize $\pm$0.002} \\
 & CheXbert F1 & 0.689\,{\scriptsize $\pm$0.003} & 0.694\,{\scriptsize $\pm$0.004} & 0.665\,{\scriptsize $\pm$0.008} & \textbf{0.731}\,{\scriptsize $\pm$0.000} & 0.726\,{\scriptsize $\pm$0.001} \\
\midrule
\multirow{4}{*}{Gemma3-4B} & BERTScore F1 & 0.315\,{\scriptsize $\pm$0.004} & 0.320\,{\scriptsize $\pm$0.002} & 0.336\,{\scriptsize $\pm$0.003} & \textbf{0.341}\,{\scriptsize $\pm$0.001} & 0.333\,{\scriptsize $\pm$0.003} \\
 & COMET & 0.620\,{\scriptsize $\pm$0.002} & 0.623\,{\scriptsize $\pm$0.001} & 0.621\,{\scriptsize $\pm$0.002} & \textbf{0.625}\,{\scriptsize $\pm$0.002} & 0.623\,{\scriptsize $\pm$0.001} \\
 & RadGraph F1 & 0.220\,{\scriptsize $\pm$0.003} & 0.226\,{\scriptsize $\pm$0.003} & 0.255\,{\scriptsize $\pm$0.003} & \textbf{0.257}\,{\scriptsize $\pm$0.004} & 0.241\,{\scriptsize $\pm$0.005} \\
 & CheXbert F1 & 0.670\,{\scriptsize $\pm$0.010} & 0.672\,{\scriptsize $\pm$0.005} & 0.645\,{\scriptsize $\pm$0.009} & \textbf{0.690}\,{\scriptsize $\pm$0.005} & 0.687\,{\scriptsize $\pm$0.005} \\
\midrule
\multirow{4}{*}{\textbf{Mean}} & BERTScore F1 & 0.329\,{\scriptsize $\pm$0.013} & 0.332\,{\scriptsize $\pm$0.011} & 0.333\,{\scriptsize $\pm$0.003} & \textbf{0.353}\,{\scriptsize $\pm$0.013} & 0.335\,{\scriptsize $\pm$0.004} \\
 & COMET & 0.620\,{\scriptsize $\pm$0.000} & 0.621\,{\scriptsize $\pm$0.003} & 0.619\,{\scriptsize $\pm$0.002} & \textbf{0.653}\,{\scriptsize $\pm$0.026} & 0.606\,{\scriptsize $\pm$0.029} \\
 & RadGraph F1 & 0.234\,{\scriptsize $\pm$0.014} & 0.239\,{\scriptsize $\pm$0.012} & 0.247\,{\scriptsize $\pm$0.007} & \textbf{0.254}\,{\scriptsize $\pm$0.029} & 0.247\,{\scriptsize $\pm$0.010} \\
 & CheXbert F1 & 0.688\,{\scriptsize $\pm$0.017} & 0.692\,{\scriptsize $\pm$0.018} & 0.657\,{\scriptsize $\pm$0.011} & \textbf{0.718}\,{\scriptsize $\pm$0.025} & 0.699\,{\scriptsize $\pm$0.023} \\
\bottomrule
\end{tabular}
\end{table}
\begin{table}[t]
\caption{\textbf{GRPO post-training results on MIMIC-CXR (Findings).} Results for different reward functions are provided. Hausdorff Distance based rewarding has the highest performance on average.}
\label{tab:grpo-mimic-findings}
\centering
\footnotesize
\setlength{\tabcolsep}{3pt}
\renewcommand{\arraystretch}{0.95}
\begin{tabular}{llccccc}
\toprule
Model & Metric & SFT & $R_{\mathrm{fmt}}$ & $R_{\mathrm{exact}}$ & $R_{\mathrm{Cham}}$ & $R_{\mathrm{Haus}}$ \\
\midrule
\multirow{4}{*}{Qwen3-VL-2B} & BERTScore F1 & 0.313\,{\scriptsize $\pm$0.003} & 0.312\,{\scriptsize $\pm$0.003} & 0.276\,{\scriptsize $\pm$0.002} & 0.320\,{\scriptsize $\pm$0.001} & \textbf{0.335}\,{\scriptsize $\pm$0.002} \\
 & COMET & 0.621\,{\scriptsize $\pm$0.002} & 0.618\,{\scriptsize $\pm$0.002} & 0.602\,{\scriptsize $\pm$0.001} & 0.604\,{\scriptsize $\pm$0.000} & \textbf{0.635}\,{\scriptsize $\pm$0.001} \\
 & RadGraph F1 & 0.198\,{\scriptsize $\pm$0.001} & 0.197\,{\scriptsize $\pm$0.004} & 0.153\,{\scriptsize $\pm$0.002} & 0.191\,{\scriptsize $\pm$0.001} & \textbf{0.231}\,{\scriptsize $\pm$0.003} \\
 & CheXbert F1 & 0.499\,{\scriptsize $\pm$0.007} & 0.502\,{\scriptsize $\pm$0.006} & 0.469\,{\scriptsize $\pm$0.008} & 0.481\,{\scriptsize $\pm$0.003} & \textbf{0.508}\,{\scriptsize $\pm$0.005} \\
\midrule
\multirow{4}{*}{Qwen3-VL-4B} & BERTScore F1 & 0.308\,{\scriptsize $\pm$0.004} & 0.310\,{\scriptsize $\pm$0.004} & 0.263\,{\scriptsize $\pm$0.002} & 0.341\,{\scriptsize $\pm$0.001} & \textbf{0.343}\,{\scriptsize $\pm$0.001} \\
 & COMET & 0.619\,{\scriptsize $\pm$0.002} & 0.619\,{\scriptsize $\pm$0.001} & 0.574\,{\scriptsize $\pm$0.001} & 0.606\,{\scriptsize $\pm$0.001} & \textbf{0.635}\,{\scriptsize $\pm$0.000} \\
 & RadGraph F1 & 0.195\,{\scriptsize $\pm$0.004} & 0.197\,{\scriptsize $\pm$0.004} & 0.157\,{\scriptsize $\pm$0.002} & \textbf{0.240}\,{\scriptsize $\pm$0.001} & 0.212\,{\scriptsize $\pm$0.002} \\
 & CheXbert F1 & 0.488\,{\scriptsize $\pm$0.006} & 0.507\,{\scriptsize $\pm$0.007} & 0.469\,{\scriptsize $\pm$0.006} & 0.511\,{\scriptsize $\pm$0.002} & \textbf{0.522}\,{\scriptsize $\pm$0.004} \\
\midrule
\multirow{4}{*}{Gemma3-4B} & BERTScore F1 & 0.326\,{\scriptsize $\pm$0.003} & 0.327\,{\scriptsize $\pm$0.003} & 0.321\,{\scriptsize $\pm$0.003} & \textbf{0.340}\,{\scriptsize $\pm$0.003} & 0.329\,{\scriptsize $\pm$0.002} \\
 & COMET & 0.627\,{\scriptsize $\pm$0.001} & 0.627\,{\scriptsize $\pm$0.001} & 0.618\,{\scriptsize $\pm$0.001} & \textbf{0.629}\,{\scriptsize $\pm$0.001} & 0.629\,{\scriptsize $\pm$0.001} \\
 & RadGraph F1 & 0.212\,{\scriptsize $\pm$0.005} & 0.214\,{\scriptsize $\pm$0.004} & 0.206\,{\scriptsize $\pm$0.003} & \textbf{0.224}\,{\scriptsize $\pm$0.004} & 0.214\,{\scriptsize $\pm$0.002} \\
 & CheXbert F1 & 0.499\,{\scriptsize $\pm$0.007} & 0.492\,{\scriptsize $\pm$0.008} & 0.482\,{\scriptsize $\pm$0.005} & 0.501\,{\scriptsize $\pm$0.006} & \textbf{0.514}\,{\scriptsize $\pm$0.012} \\
\midrule
\multirow{4}{*}{\textbf{Mean}} & BERTScore F1 & 0.316\,{\scriptsize $\pm$0.009} & 0.316\,{\scriptsize $\pm$0.009} & 0.287\,{\scriptsize $\pm$0.030} & 0.333\,{\scriptsize $\pm$0.012} & \textbf{0.336}\,{\scriptsize $\pm$0.007} \\
 & COMET & 0.622\,{\scriptsize $\pm$0.004} & 0.621\,{\scriptsize $\pm$0.005} & 0.598\,{\scriptsize $\pm$0.023} & 0.613\,{\scriptsize $\pm$0.014} & \textbf{0.633}\,{\scriptsize $\pm$0.004} \\
 & RadGraph F1 & 0.202\,{\scriptsize $\pm$0.009} & 0.203\,{\scriptsize $\pm$0.010} & 0.172\,{\scriptsize $\pm$0.030} & 0.218\,{\scriptsize $\pm$0.025} & \textbf{0.219}\,{\scriptsize $\pm$0.011} \\
 & CheXbert F1 & 0.495\,{\scriptsize $\pm$0.007} & 0.500\,{\scriptsize $\pm$0.008} & 0.473\,{\scriptsize $\pm$0.007} & 0.498\,{\scriptsize $\pm$0.015} & \textbf{0.515}\,{\scriptsize $\pm$0.007} \\
\bottomrule
\end{tabular}
\end{table}

%
%

\section{Inference-time response selection}
\label{sec:response-selection-results}

Beyond the GRPO post-training results of Sec.~\ref{sec:grpo-experiments},
we also evaluate the inference-time response selection pipeline of
Sec.~\ref{sec:response-selection} on 13 models -- our GRPO-fine-tuned
Qwen3-VL-4B variants and the closed-source generalist LLMs Mistral-Small,
Gemini Flash-Lite, GPT-4o mini and GPT-5 mini. For each closed-source
model we evaluate two distinct prompt templates, denoted \texttt{[p1]}
and \texttt{[p2]} in the tables; the verbatim prompts are reproduced in
App.~\ref{sec:prompts}. For every test sample we draw multiple
generations from the model and select one with each (distance metric,
aggregation) combination of Sec.~\ref{sec:distances}. Selected responses
are scored on a suite of NLP and clinical metrics and compared against a
random-selection baseline averaged over multiple runs. 

\begin{table}[t]
\caption{\textbf{Headline results (Findings).} For every model and every of five clinically meaningful NLP metrics we report the best score obtained by any (distance metric, aggregation) selection policy. The matched random-selection baseline is shown in italics under each model row, and the percentage improvement over random is given in parentheses. Bold marks the best value per column.}
\label{tab:headline-results}
\centering
\resizebox{\linewidth}{!}{%
\begin{tabular}{lccccc}
\toprule
Model & RL-F & METEOR & BS-F1 & RG-F1 & CXB-14 \\
\midrule
Qwen3-VL-4B GRPO ($R_{\mathrm{exact}}$) & 0.329\textsubscript{(+6.5\%)} & 0.343\textsubscript{(+4.1\%)} & 0.355\textsubscript{(+7.3\%)} & 0.266\textsubscript{(+9.6\%)} & 0.723\textsubscript{(+9.7\%)} \\
\quad\textit{random} & \textit{0.309} & \textit{0.329} & \textit{0.331} & \textit{0.242} & \textit{0.659} \\[1pt]
Qwen3-VL-4B GRPO ($R_{\mathrm{fmt}}$) & 0.329\textsubscript{(+9.3\%)} & 0.338\textsubscript{(+5.5\%)} & \textbf{0.360}\textsubscript{(+8.6\%)} & 0.272\textsubscript{(+14.8\%)} & \textbf{0.730}\textsubscript{(+5.0\%)} \\
\quad\textit{random} & \textit{0.301} & \textit{0.320} & \textit{0.331} & \textit{0.237} & \textit{0.695} \\[1pt]
Qwen3-VL-4B GRPO ($R_{\mathrm{Cham}}$) & \textbf{0.343}\textsubscript{(+0.5\%)} & \textbf{0.383}\textsubscript{(+0.5\%)} & 0.351\textsubscript{(+0.0\%)} & \textbf{0.282}\textsubscript{(+0.2\%)} & 0.729\textsubscript{(-0.2\%)} \\
\quad\textit{random} & \textit{0.341} & \textit{0.381} & \textit{0.351} & \textit{0.282} & \textit{0.731} \\[1pt]
Qwen3-VL-4B GRPO ($R_{\mathrm{Haus}}$) & 0.321\textsubscript{(+9.5\%)} & 0.324\textsubscript{(+7.9\%)} & 0.353\textsubscript{(+6.0\%)} & 0.263\textsubscript{(+9.0\%)} & 0.729\textsubscript{(+0.5\%)} \\
\quad\textit{random} & \textit{0.293} & \textit{0.301} & \textit{0.333} & \textit{0.242} & \textit{0.725} \\[1pt]
Mistral-Small [p1] & 0.186\textsubscript{(+10.8\%)} & 0.259\textsubscript{(+15.5\%)} & 0.231\textsubscript{(+13.2\%)} & 0.073\textsubscript{(+36.0\%)} & 0.619\textsubscript{(+18.5\%)} \\
\quad\textit{random} & \textit{0.168} & \textit{0.224} & \textit{0.204} & \textit{0.053} & \textit{0.522} \\[1pt]
Mistral-Small [p2] & 0.291\textsubscript{(+21.4\%)} & 0.330\textsubscript{(+14.6\%)} & 0.327\textsubscript{(+16.4\%)} & 0.229\textsubscript{(+47.2\%)} & 0.726\textsubscript{(+2.1\%)} \\
\quad\textit{random} & \textit{0.240} & \textit{0.288} & \textit{0.281} & \textit{0.155} & \textit{0.711} \\[1pt]
Gemini 2.5 Flash-Lite [p1] & 0.224\textsubscript{(+15.3\%)} & 0.291\textsubscript{(+19.2\%)} & 0.235\textsubscript{(+20.6\%)} & 0.141\textsubscript{(+34.9\%)} & 0.676\textsubscript{(+5.0\%)} \\
\quad\textit{random} & \textit{0.194} & \textit{0.244} & \textit{0.195} & \textit{0.105} & \textit{0.644} \\[1pt]
Gemini 2.5 Flash-Lite [p2] & 0.261\textsubscript{(+14.4\%)} & 0.313\textsubscript{(+14.2\%)} & 0.283\textsubscript{(+17.7\%)} & 0.172\textsubscript{(+31.0\%)} & 0.701\textsubscript{(+8.2\%)} \\
\quad\textit{random} & \textit{0.228} & \textit{0.274} & \textit{0.241} & \textit{0.132} & \textit{0.648} \\[1pt]
Gemini 3.1 Flash-Lite [p1] & 0.237\textsubscript{(+4.1\%)} & 0.335\textsubscript{(+2.0\%)} & 0.257\textsubscript{(+5.4\%)} & 0.120\textsubscript{(+7.7\%)} & 0.713\textsubscript{(+1.6\%)} \\
\quad\textit{random} & \textit{0.228} & \textit{0.328} & \textit{0.244} & \textit{0.112} & \textit{0.702} \\[1pt]
Gemini 3.1 Flash-Lite [p2] & 0.287\textsubscript{(+8.2\%)} & 0.369\textsubscript{(+4.9\%)} & 0.312\textsubscript{(+7.8\%)} & 0.196\textsubscript{(+15.9\%)} & 0.713\textsubscript{(+3.3\%)} \\
\quad\textit{random} & \textit{0.266} & \textit{0.352} & \textit{0.289} & \textit{0.169} & \textit{0.690} \\[1pt]
GPT-4o mini [p1] & 0.200\textsubscript{(+13.7\%)} & 0.273\textsubscript{(+17.0\%)} & 0.245\textsubscript{(+18.7\%)} & 0.105\textsubscript{(+29.6\%)} & 0.716\textsubscript{(+3.0\%)} \\
\quad\textit{random} & \textit{0.176} & \textit{0.233} & \textit{0.206} & \textit{0.081} & \textit{0.695} \\[1pt]
GPT-4o mini [p2] & 0.286\textsubscript{(+14.3\%)} & 0.339\textsubscript{(+11.3\%)} & 0.326\textsubscript{(+11.7\%)} & 0.216\textsubscript{(+28.9\%)} & 0.722\textsubscript{(+3.4\%)} \\
\quad\textit{random} & \textit{0.250} & \textit{0.304} & \textit{0.292} & \textit{0.167} & \textit{0.699} \\[1pt]
GPT-5 mini [p2] & 0.261\textsubscript{(+20.3\%)} & 0.302\textsubscript{(+18.7\%)} & 0.277\textsubscript{(+22.5\%)} & 0.164\textsubscript{(+31.8\%)} & 0.727\textsubscript{(+4.3\%)} \\
\quad\textit{random} & \textit{0.217} & \textit{0.254} & \textit{0.226} & \textit{0.124} & \textit{0.697} \\[1pt]
\bottomrule
\end{tabular}%
}
\end{table}

Tab.~\ref{tab:headline-results} summarises the highest performance improvements 
over random response picking
among all possible distance metric and aggregation method combinations.
It can easily be concluded that appropriate metric and aggregation method to choose 
the closest generations to the training distribution among all candidate generated texts
can improve the overall performance substantially over randomly selection.
Similar results for Impression part is reported in 
Section \ref{sec:appendix-results} (Table \ref{tab:headline-results-impressions}) in Appendix.
along with per-model breakdowns for every evaluation metrics.
In addition, a method$\,\times\,$metric heatmap visualising the same data
(Fig.~\ref{fig:results-heatmap}) is deferred to
App.~\ref{sec:appendix-results}.


\subsection{Distance-guided pruning of generations}
\label{sec:pruning}

The inference-time selection rule of Sec.~\ref{sec:response-selection}
requires generating all $K$ candidates to completion before any of them
can be scored against the training distribution, so the per-test-image
generation cost is exactly $K$ times the cost of a single decoding. We
now describe an extension that uses the same training-distribution
distance, but applied to partial generations during decoding, to
prune unpromising candidates before they are completed and thereby reduce
total token usage.

Concretely, all $K$ candidates first generate their opening
sentence in lock-step. From the second sentence onwards, before each
new sentence is decoded we encode every still-active candidate
$\hat{y}^{(k)}$ into its sentence-embedding sets
$\mathcal{E}^{S}\!\bigl(\hat{y}^{(k)}_{:t}\bigr)$ over the first $t$
generated sentences and score it against the training distribution by
$\mathfrak{D}\!\bigl(\hat{y}^{(k)}\bigr)$
(Eq.~\eqref{eq:total-train-dist}, evaluated on the partial output).
The bottom-scoring fraction of the active candidates by this score is
then dropped from further decoding so that their token generation simply
stops and decoding continues sentence-by-sentence with this
prune-then-decode loop until a single surviving candidate remains,
which is then decoded to its end-of-sequence token and returned as
the selected response. In our experiments, we set the pruning 
fraction $\rho=0.5$

Because pruned candidates stop producing tokens at the moment they are
eliminated, this scheme strictly reduces the total number of generated
tokens compared with the full-generation-then-select pipeline of
Sec.~\ref{sec:response-selection}. We compare it on the
Findings section against (i) random selection among the $K$
candidates and (ii) the standard distance-based selection of
Sec.~\ref{sec:response-selection} on complete candidates.
Tab.~\ref{tab:pruning-headline} reports BERTScore F$_1$, RadGraph F$_1$
and CheXbert F$_1$ (the three headline metrics also used in the GRPO
tables) for every (model, distance metric) combination, together with
the percentage of tokens saved by pruning relative to full generation.
The full per-metric breakdown 
and Impression numbers are in App.~\ref{sec:pruning-appendix}.

As a proof of concept, we apply this pruning policy to five
closed-source LLMs (Mistral-Small, Gemini~2.5 Flash-Lite,
Gemini~3.1 Flash-Lite, GPT-4o\,mini and GPT-5\,mini) and report the
token savings they would yield in an equivalent open-weights
deployment with the same inference budget. The procedure is also
summarised as Alg.~\ref{alg:pruning} in App.~\ref{sec:pruning-appendix}.

Although pruning requires additional set-distance calculations per sentence addition during generation, 
which is one of the limitations of our work, 
considering that the sentence-transformer is much more smaller model with 420MB in size, 
saving tokens from larger models is still a gain.
As it can be seen, Pruning beats random selection in all 
metrics across all models with an average relative improvements 
$+12.7\%$, $+17.1\%$, $+6.2\%$ on BS-F1, RG-F1 and CXB-F1
while saving 42.1--60.1\%
of the generation tokens
It does not
always match the standard full-generation policy,
but the gap is small on the metrics and the token
saving is substantial.

\begin{table}[t]
\caption{\textbf{Distance-guided pruning of generations (\textsc{Findings}).} For every (model, distance) pair we report the percentage of generation tokens saved by the pruning policy and three headline metrics scored under three selection policies: \textit{Random} (uniform random pick among the $K$ stochastic candidates), \textit{Standard} (full-generation pipeline of Sec.~\ref{sec:response-selection}, distance-based selection on complete candidates) and \textit{Pruning} (distance-guided early-pruning during decoding, this work). Bold marks the column-best of \{random, standard, pruning\} within each metric block.}
\label{tab:pruning-headline}
\centering
\small
\setlength{\tabcolsep}{4pt}
\resizebox{\linewidth}{!}{%
\begin{tabular}{lccccccccccccc}
\toprule
 &  & wu & Tok-Save & \multicolumn{3}{c}{BERTScore F1} & \multicolumn{3}{c}{RadGraph F1} & \multicolumn{3}{c}{CheXbert F1} \\
\cmidrule(lr){5-7} \cmidrule(lr){8-10} \cmidrule(lr){11-13}
Model & $\mathcal{D}$ &  & (\%) & Random & Standard & Pruning & Random & Standard & Pruning & Random & Standard & Pruning \\
\midrule
Mistral-Small [p1] & $\mathcal{D}_{\mathrm{Cham}}$ & 1 & 53.4 & 0.201 & \textbf{0.230} & 0.228 & 0.053 & \textbf{0.061} & 0.060 & 0.515 & \textbf{0.616} & 0.595 \\
 & $\mathcal{D}_{\mathrm{Haus}}$ & 1 & 44.2 & 0.201 & \textbf{0.225} & 0.224 & 0.053 & 0.053 & \textbf{0.057} & 0.515 & \textbf{0.619} & 0.605 \\
 & $\mathcal{D}_{\mathrm{Hung}}$ & 1 & 53.4 & 0.201 & \textbf{0.230} & 0.226 & 0.053 & \textbf{0.059} & 0.058 & 0.515 & \textbf{0.616} & 0.599 \\
\midrule
Gemini 2.5 Flash-Lite [p1] & $\mathcal{D}_{\mathrm{Cham}}$ & 2 & 42.1 & 0.196 & 0.234 & \textbf{0.236} & 0.104 & 0.141 & \textbf{0.142} & 0.639 & 0.667 & \textbf{0.672} \\
 & $\mathcal{D}_{\mathrm{Haus}}$ & 1 & 55.9 & 0.196 & \textbf{0.224} & 0.218 & 0.104 & \textbf{0.124} & 0.117 & 0.639 & \textbf{0.665} & 0.648 \\
 & $\mathcal{D}_{\mathrm{Hung}}$ & 2 & 42.2 & 0.196 & 0.235 & \textbf{0.235} & 0.104 & \textbf{0.140} & 0.138 & 0.639 & \textbf{0.670} & 0.667 \\
\midrule
Gemini 3.1 Flash-Lite [p1] & $\mathcal{D}_{\mathrm{Cham}}$ & 1 & 48.9 & 0.244 & \textbf{0.255} & 0.254 & 0.112 & 0.118 & \textbf{0.123} & 0.705 & \textbf{0.706} & 0.699 \\
 & $\mathcal{D}_{\mathrm{Haus}}$ & 1 & 60.1 & 0.244 & \textbf{0.256} & 0.249 & 0.112 & 0.118 & \textbf{0.118} & 0.705 & \textbf{0.708} & \textbf{0.708} \\
 & $\mathcal{D}_{\mathrm{Hung}}$ & 1 & 49.0 & 0.244 & \textbf{0.257} & 0.252 & 0.112 & \textbf{0.119} & 0.118 & 0.705 & \textbf{0.712} & 0.710 \\
\midrule
GPT-4o mini [p1] & $\mathcal{D}_{\mathrm{Cham}}$ & 1 & 55.3 & 0.206 & \textbf{0.244} & 0.238 & 0.080 & \textbf{0.105} & 0.097 & 0.688 & \textbf{0.714} & 0.702 \\
 & $\mathcal{D}_{\mathrm{Haus}}$ & 1 & 55.0 & 0.206 & \textbf{0.240} & 0.238 & 0.080 & \textbf{0.099} & 0.098 & 0.688 & \textbf{0.713} & 0.707 \\
 & $\mathcal{D}_{\mathrm{Hung}}$ & 1 & 55.2 & 0.206 & \textbf{0.244} & 0.238 & 0.080 & \textbf{0.103} & 0.097 & 0.688 & \textbf{0.715} & 0.706 \\
\midrule
\multirow{3}{*}{\textbf{Mean}} & $\mathcal{D}_{\mathrm{Cham}}$ & -- & 49.9\,{\scriptsize $\pm$5.8} & 0.212\,{\scriptsize $\pm$0.022} & \textbf{0.241}\,{\scriptsize $\pm$0.011} & 0.239\,{\scriptsize $\pm$0.011} & 0.087\,{\scriptsize $\pm$0.027} & \textbf{0.106}\,{\scriptsize $\pm$0.034} & 0.105\,{\scriptsize $\pm$0.036} & 0.637\,{\scriptsize $\pm$0.086} & \textbf{0.676}\,{\scriptsize $\pm$0.045} & 0.667\,{\scriptsize $\pm$0.050} \\
 & $\mathcal{D}_{\mathrm{Haus}}$ & -- & 53.8\,{\scriptsize $\pm$6.8} & 0.212\,{\scriptsize $\pm$0.022} & \textbf{0.236}\,{\scriptsize $\pm$0.015} & 0.232\,{\scriptsize $\pm$0.014} & 0.087\,{\scriptsize $\pm$0.027} & \textbf{0.099}\,{\scriptsize $\pm$0.032} & 0.098\,{\scriptsize $\pm$0.029} & 0.637\,{\scriptsize $\pm$0.086} & \textbf{0.676}\,{\scriptsize $\pm$0.044} & 0.667\,{\scriptsize $\pm$0.050} \\
 & $\mathcal{D}_{\mathrm{Hung}}$ & -- & 49.9\,{\scriptsize $\pm$5.8} & 0.212\,{\scriptsize $\pm$0.022} & \textbf{0.241}\,{\scriptsize $\pm$0.012} & 0.238\,{\scriptsize $\pm$0.011} & 0.087\,{\scriptsize $\pm$0.027} & \textbf{0.105}\,{\scriptsize $\pm$0.034} & 0.103\,{\scriptsize $\pm$0.034} & 0.637\,{\scriptsize $\pm$0.086} & \textbf{0.678}\,{\scriptsize $\pm$0.046} & 0.671\,{\scriptsize $\pm$0.051} \\
\bottomrule
\end{tabular}%
}
\end{table}

\section{Conclusion}
\label{sec:conclusion}

We introduced set-distance rewards as a unified signal
for chest X-ray report generation. By representing each report as
unordered sets of sentence embeddings and scoring generated
reports against references with set-to-set distances, 
we obtain a continuous, permutation-invariant reward.
At training time, GRPO post-training with our Chamfer- and
Hausdorff-based rewards consistently outperformed both supervised
fine-tuning and discrete exact-match GRPO across three
vision--language backbones on two datasets.
At inference time, the same family of set distances
served as a gradient-free best-of-$N$ selection criterion, improving
both our GRPO-fine-tuned models and a panel of closed-source
generalist LLMs over a random-selection baseline.
Finally, evaluated on partial decodings, the
same signal supported a sentence-level pruning policy that retained
the quality benefits of full best-of-$N$ selection while cutting roughly
half of the generated tokens.

\clearpage
\bibliographystyle{unsrt}
\bibliography{refs}

\clearpage
\appendix
\raggedbottom
\section{Set-to-set distance metrics}
\label{sec:distances}

All metrics defined below operate on two finite, non-empty sets
$\mathcal{A} = \{\mathbf{a}_{1},\dots,\mathbf{a}_{n}\}$ and
$\mathcal{B} = \{\mathbf{b}_{1},\dots,\mathbf{b}_{m}\}$ of unit-norm
sentence embeddings in $\mathbb{R}^{d}$, where the cardinalities
$n = |\mathcal{A}|$ and $m = |\mathcal{B}|$ may differ. Let
$d : \mathbb{R}^{d}\!\times\!\mathbb{R}^{d} \to [0,1]$ be a base point-to-point
distance; throughout this work we use the cosine distance
\begin{equation}
    d(\mathbf{u},\mathbf{v})
    \;=\;
    \tfrac{1}{2}\!\left(1 \;-\; \frac{\mathbf{u}^{\top}\mathbf{v}}{\|\mathbf{u}\|_{2}\,\|\mathbf{v}\|_{2}}\right)
    \;\in\; [0,1],
\end{equation}
which is well matched to the unit-normalised outputs of the sentence
transformer $E_{\phi}$. The pairwise cost matrix is
$M \in \mathbb{R}^{n\times m}$ with $M_{ij} = d(\mathbf{a}_{i},\mathbf{b}_{j})$,
and for the optimal-transport based metrics we additionally rescale it to
the unit interval, $\widetilde{M} = M / \max_{ij} M_{ij}$, so that the
resulting transport cost lies in $[0,1]$. Each metric $\mathcal{D}$ is
designed so that $\mathcal{D}(\mathcal{A},\mathcal{A}) = 0$ and larger
values correspond to greater dissimilarity, which translates into the
similarity reward $1 - \mathcal{D}$ used in
Section~\ref{sec:semantic-reward}.

\paragraph{Chamfer distance.}
The (symmetric) Chamfer distance averages, over each set, the
nearest-neighbour cost to the other set:
\begin{equation}
    \mathcal{D}_{\mathrm{Chamfer}}(\mathcal{A},\mathcal{B})
    \;=\;
    \tfrac{1}{2}\!\left(
        \frac{1}{n}\sum_{i=1}^{n} \min_{1\le j\le m} M_{ij}
        \;+\;
        \frac{1}{m}\sum_{j=1}^{m} \min_{1\le i\le n} M_{ij}
    \right).
    \label{eq:chamfer}
\end{equation}
The first term rewards every generated sentence for being close to
\emph{some} reference sentence; the second term penalises reference sentences
that are not covered by any generated sentence. Chamfer does not enforce a
one-to-one correspondence and can re-use the same target sentence for
multiple sources.

\paragraph{Hausdorff distance.}
The (symmetric) Hausdorff distance replaces the averages in
Eq.~\eqref{eq:chamfer} with maxima, yielding the worst-case
nearest-neighbour cost in either direction:
\begin{equation}
    \mathcal{D}_{\mathrm{Hausdorff}}(\mathcal{A},\mathcal{B})
    \;=\;
    \max\!\left\{
        \max_{1\le i\le n} \min_{1\le j\le m} M_{ij},\;
        \max_{1\le j\le m} \min_{1\le i\le n} M_{ij}
    \right\}.
    \label{eq:hausdorff}
\end{equation}
A single uncovered sentence on either side is enough to dominate the score,
which makes Hausdorff a strictly harsher reward than Chamfer.

\paragraph{Optimal transport (Wasserstein) distance.}
We assign the uniform discrete measure to each set,
$\mathbf{p} = \tfrac{1}{n}\mathbf{1}_{n}$ and
$\mathbf{q} = \tfrac{1}{m}\mathbf{1}_{m}$, and define the transport polytope
\[
    \Gamma(\mathbf{p},\mathbf{q})
    \;=\;
    \bigl\{\,
        \gamma \in \mathbb{R}_{\ge 0}^{\,n\times m}
        \;:\;
        \gamma\,\mathbf{1}_{m} = \mathbf{p},\;\;
        \gamma^{\!\top}\mathbf{1}_{n} = \mathbf{q}
    \,\bigr\}.
\]
The optimal-transport (Wasserstein) distance is then
\begin{equation}
    \mathcal{D}_{\mathrm{OT}}(\mathcal{A},\mathcal{B})
    \;=\;
    \min_{\gamma \in \Gamma(\mathbf{p},\mathbf{q})}\;
    \sum_{i=1}^{n}\sum_{j=1}^{m} \gamma_{ij}\,\widetilde{M}_{ij}.
    \label{eq:ot}
\end{equation}
OT recovers a soft, mass-preserving many-to-many alignment between the two
sets and penalises systematic differences in their distribution even when
individual nearest-neighbour distances are small.

\paragraph{Sinkhorn (entropy-regularised) distance.}
To obtain a smoother, faster-to-compute proxy for OT we add an entropic
penalty
$H(\gamma) = -\sum_{i,j}\gamma_{ij}\bigl(\log \gamma_{ij} - 1\bigr)$:
\begin{equation}
    \mathcal{D}^{\,\varepsilon}_{\mathrm{Sinkhorn}}(\mathcal{A},\mathcal{B})
    \;=\;
    \min_{\gamma \in \Gamma(\mathbf{p},\mathbf{q})}
    \;\sum_{i,j} \gamma_{ij}\,\widetilde{M}_{ij}
    \;-\; \varepsilon\,H(\gamma),
    \qquad \varepsilon > 0,
    \label{eq:sinkhorn}
\end{equation}
solved by Sinkhorn--Knopp iterations. We evaluate three regimes
$\varepsilon \in \{0.01,\,0.1,\,0.5\}$: smaller $\varepsilon$ approaches
exact OT at higher iteration cost, while larger $\varepsilon$ yields a
smoother and more strongly regularised estimate.

\paragraph{Unbalanced optimal transport.}
When $|\mathcal{A}| \neq |\mathcal{B}|$ or when individual sentences on one
side genuinely have no counterpart on the other, the hard marginal
constraints in Eq.~\eqref{eq:ot} can be inappropriate. Unbalanced OT
relaxes them with soft Kullback--Leibler penalties of strength $\tau > 0$:
\begin{equation}
\begin{aligned}
    \mathcal{D}^{\,\varepsilon,\tau}_{\mathrm{UOT}}(\mathcal{A},\mathcal{B})
    \;=\;
    \min_{\gamma \in \mathbb{R}_{\ge 0}^{\,n\times m}}\;
        & \sum_{i,j} \gamma_{ij}\,\widetilde{M}_{ij}
        \;-\; \varepsilon\, H(\gamma) \\[-1pt]
    & \;+\; \tau\,\mathrm{KL}\!\bigl(\gamma\,\mathbf{1}_{m}\,\big\|\,\mathbf{p}\bigr)
        \;+\; \tau\,\mathrm{KL}\!\bigl(\gamma^{\!\top}\mathbf{1}_{n}\,\big\|\,\mathbf{q}\bigr).
\end{aligned}
    \label{eq:uot}
\end{equation}
We fix $\varepsilon = 0.1$ and report two relaxation regimes
$\tau \in \{0.5,\,1.0\}$: smaller $\tau$ tolerates larger discrepancies in
set size by allowing more mass to be created or destroyed.

\paragraph{Gromov--Wasserstein distance.}
Gromov--Wasserstein (GW) compares the \emph{intrinsic geometry} of the two
sets rather than their absolute coordinates: it asks how well the pairwise
distances inside $\mathcal{A}$ can be mapped to the pairwise distances
inside $\mathcal{B}$. Let
$C^{\mathcal{A}}\in\mathbb{R}^{n\times n}$ and
$C^{\mathcal{B}}\in\mathbb{R}^{m\times m}$ be the rescaled intra-set
distance matrices, with
$C^{\mathcal{A}}_{ik} = d(\mathbf{a}_{i},\mathbf{a}_{k})\,/\,\max_{i'k'} d(\mathbf{a}_{i'},\mathbf{a}_{k'})$
and $C^{\mathcal{B}}$ analogously. The (entropic) GW distance is
\begin{equation}
    \mathcal{D}^{\,\varepsilon}_{\mathrm{GW}}(\mathcal{A},\mathcal{B})
    \;=\;
    \min_{\gamma \in \Gamma(\mathbf{p},\mathbf{q})}\;
        \sum_{i,j,k,\ell}
        \bigl(C^{\mathcal{A}}_{ik} - C^{\mathcal{B}}_{j\ell}\bigr)^{\!2}\,
        \gamma_{ij}\,\gamma_{k\ell}
    \;-\; \varepsilon\, H(\gamma),
    \label{eq:gw}
\end{equation}
which we solve with $\varepsilon = 0.1$. GW is invariant under isometries of
either set and is therefore well suited to comparing reports whose absolute
embedding positions may shift while their internal sentence-to-sentence
structure is preserved. We treat the case $\min(n,m) < 2$, in which the
intra-set geometry is trivial, as undefined and fall back to a zero
similarity.

\paragraph{Hungarian matching with nearest-neighbour fallback.}
In contrast to OT-style soft alignments, Hungarian matching enforces a strict
one-to-one correspondence on the smaller set. Without loss of generality
assume $n \le m$, and let
\begin{equation}
    \pi^{\!*}
    \;=\;
    \operatorname*{arg\,min}_{\pi:\{1,\dots,n\}\hookrightarrow\{1,\dots,m\}}
    \sum_{i=1}^{n} M_{i,\pi(i)}
    \label{eq:hungarian-assignment}
\end{equation}
be the optimal injective assignment returned by the Hungarian algorithm.
Writing $\mathcal{U} = \{1,\dots,m\}\setminus \pi^{\!*}(\{1,\dots,n\})$ for
the unmatched indices on the larger side, the nearest-neighbour variant
attributes to each unmatched element only the cost of its closest neighbour
on the smaller side and normalises by $\max(n,m)$ so that the score is
comparable across set sizes:
\begin{equation}
    \mathcal{D}_{\mathrm{Hung\text{-}NN}}(\mathcal{A},\mathcal{B})
    \;=\;
    \frac{1}{\max(n,m)}
    \!\left[\;
        \sum_{i=1}^{n} M_{i,\pi^{\!*}(i)}
        \;+\;
        \sum_{j \in \mathcal{U}} \min_{1\le i\le n} M_{ij}
    \;\right].
    \label{eq:hung-nn}
\end{equation}

\paragraph{Hungarian matching with count penalty.}
An alternative treatment of unmatched elements replaces the per-element
nearest-neighbour cost in Eq.~\eqref{eq:hung-nn} with a fixed penalty
$\alpha > 0$:
\begin{equation}
    \mathcal{D}^{\,\alpha}_{\mathrm{Hung\text{-}Pen}}(\mathcal{A},\mathcal{B})
    \;=\;
    \frac{1}{\min(n,m)}\sum_{i=1}^{\min(n,m)} M_{i,\pi^{\!*}(i)}
    \;+\;
    \alpha\,\bigl|\,n - m\,\bigr|.
    \label{eq:hung-pen}
\end{equation}
Here the first term is the mean cost of the matched pairs and the second
term penalises any imbalance in cardinality, independently of the semantic
quality of the matched sentences. We evaluate $\alpha \in \{0.1,\,0.5\}$:
larger $\alpha$ penalises reports with the wrong number of sentences more
aggressively.

\paragraph{Partial optimal transport.}
Partial OT (POT) relaxes the mass-preservation constraint of OT by
transporting only a fraction $\rho \in (0, 1]$ of the total mass:
\begin{equation}
    \mathcal{D}^{\,\rho}_{\mathrm{POT}}(\mathcal{A},\mathcal{B})
    \;=\;
    \min_{\substack{
        \gamma \in \mathbb{R}_{\ge 0}^{\,n\times m} \\[1pt]
        \gamma\,\mathbf{1}_{m} \le \mathbf{p} \\[1pt]
        \gamma^{\!\top}\mathbf{1}_{n} \le \mathbf{q} \\[1pt]
        \mathbf{1}_{n}^{\!\top}\,\gamma\,\mathbf{1}_{m} = \rho
    }}
    \;\sum_{i,j}\gamma_{ij}\,\widetilde{M}_{ij}.
    \label{eq:pot}
\end{equation}
Intuitively, a fraction $1-\rho$ of the mass on each side may be
\emph{discarded} at zero cost, capturing the intuition that not every
generated sentence requires a counterpart in the reference (and vice versa).
We use three settings: an adaptive
$\rho = \min(n,m) / \max(n,m)$, which is the largest ratio that avoids
forcing spurious matches under asymmetric cardinalities, and two fixed
values $\rho \in \{0.5,\,0.8\}$.

\paragraph{Summary.}
Together, these nine families of metrics cover three complementary views of
set similarity. \emph{Nearest-neighbour} metrics (Chamfer, Hausdorff)
measure local coverage and are cheap to compute but ignore one-to-one
constraints. \emph{Transport-based} metrics (OT,
Partial OT) explicitly model how mass must be moved between the two
sets and naturally handle continuous, asymmetric or geometrically
structured discrepancies. \emph{Assignment-based} metrics (Hungarian-NN,
Hungarian-Pen) commit to a discrete one-to-one correspondence on the
smaller set and treat cardinality mismatches with an explicit penalty.
We compare all of them as candidate semantic rewards in our GRPO
fine-tuning experiments.


\section{Prompts used for closed-source LLM evaluations}
\label{sec:prompts}

For each closed-source model evaluated in the response-selection
experiments (Mistral-Small, Gemini~2.5 Flash-Lite, Gemini~3.1 Flash-Lite,
GPT-4o\,mini and GPT-5\,mini), we draw $K$ stochastic completions per test
sample from the API under one of two prompt templates, denoted
\texttt{[p1]} and \texttt{[p2]} in the result tables. The two prompts are
reproduced verbatim below; both enforce the same output schema
(\texttt{Findings:\,<text>} followed by \texttt{Impression:\,<text>}), but
\texttt{[p2]} additionally provides five in-context exemplars.

\tcbset{
    promptbox/.style={
        enhanced jigsaw,
        breakable,
        sharp corners=south,
        arc=4pt,
        colback=blue!2!white,
        colframe=blue!35!gray,
        boxrule=0.55pt,
        left=10pt, right=10pt, top=8pt, bottom=8pt,
        coltitle=white,
        colbacktitle=blue!50!gray,
        fonttitle=\bfseries\sffamily\small,
        fontupper=\ttfamily\footnotesize,
        attach boxed title to top left={yshift=-2pt, xshift=8pt},
        boxed title style={
            colframe=blue!50!gray,
            arc=2pt,
            boxrule=0pt,
            top=2pt, bottom=2pt, left=6pt, right=6pt
        },
        before upper={\setlength{\parindent}{0pt}\setlength{\parskip}{4pt}}
    }
}

\subsection{Prompt~1 (zero-shot)}
\label{sec:prompt1}

\begin{tcolorbox}[promptbox, title={Prompt~1\,---\,zero-shot}]
You are a radiology report generation model specialized in chest X-rays.

Generate a concise clinical report for the given image.

Strict requirements:

- Output ONLY in this exact format:
\\[2pt]
\hspace*{2em}\verb|Findings: <text>|
\\[1pt]
\hspace*{2em}\verb|Impression: <text>|

- Do NOT include explanations, disclaimers, or any extra text.

- Do NOT include phrases such as ``consult a doctor'' or ``this is not medical advice''.

- Use professional radiology language.

- Keep it concise and structured.

\medskip
Output:
\end{tcolorbox}

\subsection{Prompt~2 (five-shot)}
\label{sec:prompt2}

\begin{tcolorbox}[promptbox, title={Prompt~2\,---\,five-shot}]
You are a radiology report generation model. Given a chest X-ray image,
generate a concise radiology report.

Follow these strict rules:

- Output ONLY in this format:
\\[2pt]
\hspace*{2em}\verb|Findings: <text>|
\\[1pt]
\hspace*{2em}\verb|Impression: <text>|

- Do NOT include any explanations, disclaimers, or additional commentary.

- Do NOT say things like ``consult a doctor'' or ``this is not medical advice''.

- Match the writing style, tone, and structure of the examples below.

- Be concise and clinically accurate.

\medskip
Here are example reports:

\medskip
\textbf{Example~0}\\
\verb|Findings:| Mild cardiomegaly. No edema. No consolidation or effusion.
No pneumothorax.\\
\verb|Impression:| Mild cardiomegaly.

\medskip
\textbf{Example~1}\\
\verb|Findings:| No pneumonia is seen. Minimal peribronchial thickening
is noted. The heart is within normal limits in size. No bony abnormality
is seen.\\
\verb|Impression:| No pneumonia. Mild peribronchial thickening.

\medskip
\textbf{Example~2}\\
\verb|Findings:| The heart size and mediastinal contours are within normal
limits. Both lungs are clear. The visualized skeletal structures are
unremarkable.\\
\verb|Impression:| No active cardiopulmonary disease.

\medskip
\textbf{Example~3}\\
\verb|Findings:| The lungs are well-expanded. The interstitial markings
are increased bilaterally. Patchy areas of confluence are noted in the
mid to lower left lung and at the right lung base. The heart and pulmonary
vascularity are normal. The mediastinum is normal in width. There is
multilevel degenerative disc disease of the thoracic spine.\\
\verb|Impression:| Bilateral interstitial pneumonia with patchy areas of
alveolar infiltrate. No pulmonary edema. No pleural effusion. Followup PA
and lateral chest X-ray is recommended in 3--4 weeks following trial of
antibiotic therapy to ensure resolution and exclude underlying malignancy.

\medskip
\textbf{Example~4}\\
\verb|Findings:| The heart size and mediastinal contours are within normal
limits. Both lungs are clear. The visualized skeletal structures are
unremarkable.\\
\verb|Impression:| No active disease.

\medskip
Now generate the report for the given chest X-ray image.

\medskip
Output:
\end{tcolorbox}


\section{Experimental setup}
\label{sec:experimental-setup}

\subsection{Datasets}
We had 179,778 samples in training and 45,364 samples for validation for MIMIC-CXR.
We get all the (image,report) pairs in RexGradient so that we have 
238,968 training and 17,007 validation samples which is splitted in the original dataset.

\subsection{SFT - Parameters}

\begin{description}
  \item[GPU:] 1xH100
  \item[Effective Batch size:] 96
  \item[Optimiser:] AdamW
  \item[Learning rate:] 1e-4
  \item[Number of Workers:] 8
\end{description}

\subsection{GRPO post-training - Parameters}

\begin{description}
  \item[GPU:] 2xH100 Nvidia GPUs
  \item[Effective Batch size:] 64
  \item[Optimiser:] AdamW
  \item[Learning rate:] 2e-5
  \item[Group size $G$:] 8
  \item[Number of Workers:] 8
  \item[Reward weights:] $\lambda_{\mathrm{fmt}} = 1$, $\lambda_{\mathrm{sem}} = 1$
\end{description}

\subsection{Inference-time response selection}
\label{sec:appendix-selection-setup}

For every test image we draw $K = 5$ 
stochastic generations from the target policy with their default temperatures.
The training-corpus reference embeddings
$\{\mathcal{E}^{S}(r^{(t)})\}_{t,S}$ are pre-computed for $N=5000$ samples
once per dataset that are randomly selected and cached on disk. 
This makes the sampel to training distribution distance calculations faster.
Morever, we limit the number of test samples for response selection experiments 
to be $N=1000$ to limit the api costs.


\section{Full GRPO post-training results}
\label{sec:grpo-appendix}

This appendix reports, for every (dataset, model, report section), the mean
and sample standard deviation across 5 random seeds for every NLP metric we
evaluated. Bold marks the column-best reward function within each table.

\subsection{Headline tables for the \textsc{Impression} section}

We first reproduce, for the \textsc{Impression} section, the same four-metric
headline tables shown in the main paper for the \textsc{Findings} section
(Tabs.~\ref{tab:grpo-rexgradient-findings} and
\ref{tab:grpo-mimic-findings}). Each row pair reports the mean and
sample standard deviation over 5 random seeds; the final \emph{Mean} block
averages each (reward, metric) cell across the models in that dataset.

\begin{table}[H]
\caption{\textbf{GRPO post-training results on ReXGradient (Impression).} Mean over 5 random seeds with sample std in \scriptsize. The final \emph{Mean} block averages across models. \textbf{Bold} = column-best reward function per row.}
\label{tab:grpo-rexgradient-impressions}
\centering
\footnotesize
\setlength{\tabcolsep}{3pt}
\renewcommand{\arraystretch}{0.95}
\begin{tabular}{llccccc}
\toprule
Model & Metric & SFT & $R_{\mathrm{fmt}}$ & $R_{\mathrm{exact}}$ & $R_{\mathrm{Cham}}$ & $R_{\mathrm{Haus}}$ \\
\midrule
\multirow{4}{*}{Qwen3-VL-2B} & BERTScore F1 & 0.324\,{\scriptsize $\pm$0.004} & 0.326\,{\scriptsize $\pm$0.006} & 0.322\,{\scriptsize $\pm$0.004} & \textbf{0.336}\,{\scriptsize $\pm$0.002} & 0.334\,{\scriptsize $\pm$0.003} \\
 & COMET & 0.592\,{\scriptsize $\pm$0.004} & 0.594\,{\scriptsize $\pm$0.003} & 0.594\,{\scriptsize $\pm$0.002} & 0.599\,{\scriptsize $\pm$0.001} & \textbf{0.601}\,{\scriptsize $\pm$0.003} \\
 & RadGraph F1 & 0.224\,{\scriptsize $\pm$0.004} & 0.225\,{\scriptsize $\pm$0.006} & 0.220\,{\scriptsize $\pm$0.005} & 0.160\,{\scriptsize $\pm$0.003} & \textbf{0.230}\,{\scriptsize $\pm$0.005} \\
 & CheXbert F1 & 0.755\,{\scriptsize $\pm$0.004} & 0.760\,{\scriptsize $\pm$0.006} & 0.705\,{\scriptsize $\pm$0.009} & \textbf{0.778}\,{\scriptsize $\pm$0.000} & 0.733\,{\scriptsize $\pm$0.010} \\
\midrule
\multirow{4}{*}{Qwen3-VL-4B} & BERTScore F1 & 0.322\,{\scriptsize $\pm$0.003} & 0.322\,{\scriptsize $\pm$0.005} & 0.324\,{\scriptsize $\pm$0.002} & 0.343\,{\scriptsize $\pm$0.001} & \textbf{0.347}\,{\scriptsize $\pm$0.002} \\
 & COMET & 0.589\,{\scriptsize $\pm$0.001} & 0.587\,{\scriptsize $\pm$0.003} & 0.596\,{\scriptsize $\pm$0.001} & \textbf{0.611}\,{\scriptsize $\pm$0.000} & 0.604\,{\scriptsize $\pm$0.001} \\
 & RadGraph F1 & 0.211\,{\scriptsize $\pm$0.003} & 0.210\,{\scriptsize $\pm$0.005} & 0.223\,{\scriptsize $\pm$0.003} & \textbf{0.257}\,{\scriptsize $\pm$0.000} & 0.174\,{\scriptsize $\pm$0.002} \\
 & CheXbert F1 & 0.735\,{\scriptsize $\pm$0.004} & 0.740\,{\scriptsize $\pm$0.004} & 0.695\,{\scriptsize $\pm$0.010} & \textbf{0.777}\,{\scriptsize $\pm$0.001} & 0.771\,{\scriptsize $\pm$0.001} \\
\midrule
\multirow{4}{*}{Gemma3-4B} & BERTScore F1 & 0.293\,{\scriptsize $\pm$0.006} & 0.296\,{\scriptsize $\pm$0.004} & 0.321\,{\scriptsize $\pm$0.005} & \textbf{0.328}\,{\scriptsize $\pm$0.004} & 0.315\,{\scriptsize $\pm$0.003} \\
 & COMET & 0.568\,{\scriptsize $\pm$0.003} & 0.568\,{\scriptsize $\pm$0.003} & 0.596\,{\scriptsize $\pm$0.002} & \textbf{0.600}\,{\scriptsize $\pm$0.002} & 0.585\,{\scriptsize $\pm$0.002} \\
 & RadGraph F1 & 0.168\,{\scriptsize $\pm$0.006} & 0.171\,{\scriptsize $\pm$0.004} & \textbf{0.227}\,{\scriptsize $\pm$0.003} & 0.222\,{\scriptsize $\pm$0.004} & 0.200\,{\scriptsize $\pm$0.003} \\
 & CheXbert F1 & 0.712\,{\scriptsize $\pm$0.012} & 0.715\,{\scriptsize $\pm$0.004} & 0.675\,{\scriptsize $\pm$0.008} & \textbf{0.740}\,{\scriptsize $\pm$0.004} & 0.733\,{\scriptsize $\pm$0.003} \\
\midrule
\multirow{4}{*}{\textbf{Mean}} & BERTScore F1 & 0.313\,{\scriptsize $\pm$0.018} & 0.315\,{\scriptsize $\pm$0.017} & 0.322\,{\scriptsize $\pm$0.002} & \textbf{0.336}\,{\scriptsize $\pm$0.008} & 0.332\,{\scriptsize $\pm$0.016} \\
 & COMET & 0.583\,{\scriptsize $\pm$0.013} & 0.583\,{\scriptsize $\pm$0.013} & 0.595\,{\scriptsize $\pm$0.001} & \textbf{0.603}\,{\scriptsize $\pm$0.007} & 0.597\,{\scriptsize $\pm$0.011} \\
 & RadGraph F1 & 0.201\,{\scriptsize $\pm$0.029} & 0.202\,{\scriptsize $\pm$0.028} & \textbf{0.223}\,{\scriptsize $\pm$0.003} & 0.213\,{\scriptsize $\pm$0.049} & 0.202\,{\scriptsize $\pm$0.028} \\
 & CheXbert F1 & 0.734\,{\scriptsize $\pm$0.022} & 0.738\,{\scriptsize $\pm$0.022} & 0.692\,{\scriptsize $\pm$0.015} & \textbf{0.765}\,{\scriptsize $\pm$0.022} & 0.746\,{\scriptsize $\pm$0.022} \\
\bottomrule
\end{tabular}
\end{table}

\begin{table}[H]
\caption{\textbf{GRPO post-training results on MIMIC-CXR (Impression).} Mean over 5 random seeds with sample std in \scriptsize. The final \emph{Mean} block averages across models. \textbf{Bold} = column-best reward function per row.}
\label{tab:grpo-mimic-impressions}
\centering
\footnotesize
\setlength{\tabcolsep}{3pt}
\renewcommand{\arraystretch}{0.95}
\begin{tabular}{llccccc}
\toprule
Model & Metric & SFT & $R_{\mathrm{fmt}}$ & $R_{\mathrm{exact}}$ & $R_{\mathrm{Cham}}$ & $R_{\mathrm{Haus}}$ \\
\midrule
\multirow{4}{*}{Qwen3-VL-2B} & BERTScore F1 & 0.337\,{\scriptsize $\pm$0.004} & 0.339\,{\scriptsize $\pm$0.003} & \textbf{0.353}\,{\scriptsize $\pm$0.002} & 0.271\,{\scriptsize $\pm$0.001} & 0.342\,{\scriptsize $\pm$0.004} \\
 & COMET & 0.587\,{\scriptsize $\pm$0.002} & 0.588\,{\scriptsize $\pm$0.002} & 0.590\,{\scriptsize $\pm$0.001} & 0.593\,{\scriptsize $\pm$0.001} & \textbf{0.629}\,{\scriptsize $\pm$0.002} \\
 & RadGraph F1 & 0.219\,{\scriptsize $\pm$0.004} & 0.217\,{\scriptsize $\pm$0.006} & \textbf{0.264}\,{\scriptsize $\pm$0.002} & 0.250\,{\scriptsize $\pm$0.001} & 0.147\,{\scriptsize $\pm$0.004} \\
 & CheXbert F1 & 0.616\,{\scriptsize $\pm$0.006} & 0.624\,{\scriptsize $\pm$0.010} & 0.562\,{\scriptsize $\pm$0.008} & 0.573\,{\scriptsize $\pm$0.004} & \textbf{0.640}\,{\scriptsize $\pm$0.006} \\
\midrule
\multirow{4}{*}{Qwen3-VL-4B} & BERTScore F1 & 0.326\,{\scriptsize $\pm$0.003} & 0.329\,{\scriptsize $\pm$0.009} & \textbf{0.378}\,{\scriptsize $\pm$0.003} & 0.374\,{\scriptsize $\pm$0.002} & 0.298\,{\scriptsize $\pm$0.001} \\
 & COMET & 0.582\,{\scriptsize $\pm$0.001} & 0.584\,{\scriptsize $\pm$0.003} & 0.597\,{\scriptsize $\pm$0.002} & \textbf{0.605}\,{\scriptsize $\pm$0.001} & 0.564\,{\scriptsize $\pm$0.001} \\
 & RadGraph F1 & 0.199\,{\scriptsize $\pm$0.004} & 0.208\,{\scriptsize $\pm$0.010} & \textbf{0.276}\,{\scriptsize $\pm$0.004} & 0.254\,{\scriptsize $\pm$0.003} & 0.142\,{\scriptsize $\pm$0.000} \\
 & CheXbert F1 & 0.608\,{\scriptsize $\pm$0.004} & 0.628\,{\scriptsize $\pm$0.012} & 0.558\,{\scriptsize $\pm$0.005} & 0.620\,{\scriptsize $\pm$0.004} & \textbf{0.659}\,{\scriptsize $\pm$0.003} \\
\midrule
\multirow{4}{*}{Gemma3-4B} & BERTScore F1 & 0.323\,{\scriptsize $\pm$0.010} & 0.326\,{\scriptsize $\pm$0.007} & 0.346\,{\scriptsize $\pm$0.009} & \textbf{0.360}\,{\scriptsize $\pm$0.007} & 0.343\,{\scriptsize $\pm$0.005} \\
 & COMET & 0.582\,{\scriptsize $\pm$0.007} & 0.583\,{\scriptsize $\pm$0.003} & 0.593\,{\scriptsize $\pm$0.005} & \textbf{0.606}\,{\scriptsize $\pm$0.002} & 0.596\,{\scriptsize $\pm$0.003} \\
 & RadGraph F1 & 0.195\,{\scriptsize $\pm$0.011} & 0.199\,{\scriptsize $\pm$0.008} & \textbf{0.246}\,{\scriptsize $\pm$0.009} & 0.234\,{\scriptsize $\pm$0.006} & 0.225\,{\scriptsize $\pm$0.008} \\
 & CheXbert F1 & 0.593\,{\scriptsize $\pm$0.008} & 0.599\,{\scriptsize $\pm$0.014} & 0.578\,{\scriptsize $\pm$0.011} & 0.606\,{\scriptsize $\pm$0.006} & \textbf{0.623}\,{\scriptsize $\pm$0.012} \\
\midrule
\multirow{4}{*}{\textbf{Mean}} & BERTScore F1 & 0.329\,{\scriptsize $\pm$0.008} & 0.331\,{\scriptsize $\pm$0.007} & \textbf{0.359}\,{\scriptsize $\pm$0.016} & 0.335\,{\scriptsize $\pm$0.056} & 0.328\,{\scriptsize $\pm$0.026} \\
 & COMET & 0.584\,{\scriptsize $\pm$0.003} & 0.585\,{\scriptsize $\pm$0.002} & 0.594\,{\scriptsize $\pm$0.003} & \textbf{0.601}\,{\scriptsize $\pm$0.007} & 0.596\,{\scriptsize $\pm$0.032} \\
 & RadGraph F1 & 0.204\,{\scriptsize $\pm$0.013} & 0.208\,{\scriptsize $\pm$0.009} & \textbf{0.262}\,{\scriptsize $\pm$0.015} & 0.246\,{\scriptsize $\pm$0.011} & 0.171\,{\scriptsize $\pm$0.046} \\
 & CheXbert F1 & 0.606\,{\scriptsize $\pm$0.012} & 0.617\,{\scriptsize $\pm$0.016} & 0.566\,{\scriptsize $\pm$0.010} & 0.600\,{\scriptsize $\pm$0.024} & \textbf{0.641}\,{\scriptsize $\pm$0.018} \\
\bottomrule
\end{tabular}
\end{table}

\subsection{ReXGradient}


\subsubsection{Findings}

\begin{table}[H]
\caption{\textbf{GRPO post-training results on ReXGradient (Findings).} BLEU (sentence \& corpus, $n=1\dots 4$). Mean over 5 random seeds with sample std in \scriptsize. The final \emph{Mean} block averages each (reward, metric) cell across the three models. \textbf{Bold} = column-best reward function per row.}
\label{tab:grpo-app-rexgradient-findings-bleu}
\centering
\footnotesize
\setlength{\tabcolsep}{3pt}
\renewcommand{\arraystretch}{0.95}
\resizebox{\linewidth}{!}{%
\begin{tabular}{llccccc}
\toprule
Model & Metric & SFT & $R_{\mathrm{fmt}}$ & $R_{\mathrm{exact}}$ & $R_{\mathrm{Cham}}$ & $R_{\mathrm{Haus}}$ \\
\midrule
\multirow{8}{*}{Qwen3-VL-2B} & S-BLEU-1 & 0.345\,{\scriptsize $\pm$0.005} & 0.348\,{\scriptsize $\pm$0.003} & 0.352\,{\scriptsize $\pm$0.003} & \textbf{0.379}\,{\scriptsize $\pm$0.001} & 0.359\,{\scriptsize $\pm$0.004} \\
 & S-BLEU-2 & 0.249\,{\scriptsize $\pm$0.005} & 0.252\,{\scriptsize $\pm$0.004} & 0.249\,{\scriptsize $\pm$0.003} & \textbf{0.265}\,{\scriptsize $\pm$0.001} & 0.258\,{\scriptsize $\pm$0.003} \\
 & S-BLEU-3 & 0.198\,{\scriptsize $\pm$0.005} & 0.200\,{\scriptsize $\pm$0.004} & 0.197\,{\scriptsize $\pm$0.004} & 0.191\,{\scriptsize $\pm$0.002} & \textbf{0.206}\,{\scriptsize $\pm$0.003} \\
 & S-BLEU-4 & 0.171\,{\scriptsize $\pm$0.005} & 0.173\,{\scriptsize $\pm$0.004} & 0.169\,{\scriptsize $\pm$0.004} & 0.146\,{\scriptsize $\pm$0.002} & \textbf{0.178}\,{\scriptsize $\pm$0.003} \\
 & C-BLEU-1 & 0.311\,{\scriptsize $\pm$0.005} & 0.313\,{\scriptsize $\pm$0.003} & 0.337\,{\scriptsize $\pm$0.004} & \textbf{0.386}\,{\scriptsize $\pm$0.001} & 0.344\,{\scriptsize $\pm$0.005} \\
 & C-BLEU-2 & 0.223\,{\scriptsize $\pm$0.004} & 0.225\,{\scriptsize $\pm$0.003} & 0.236\,{\scriptsize $\pm$0.003} & \textbf{0.272}\,{\scriptsize $\pm$0.002} & 0.244\,{\scriptsize $\pm$0.004} \\
 & C-BLEU-3 & 0.177\,{\scriptsize $\pm$0.004} & 0.179\,{\scriptsize $\pm$0.003} & 0.185\,{\scriptsize $\pm$0.003} & \textbf{0.203}\,{\scriptsize $\pm$0.002} & 0.193\,{\scriptsize $\pm$0.003} \\
\bottomrule
\end{tabular}
}
\end{table}

\begin{table}[H]
\caption{\textbf{GRPO post-training results on ReXGradient (Findings).} ROUGE (F$_1$ for $n=1$, $n=2$, $L$). Mean over 5 random seeds with sample std in \scriptsize. The final \emph{Mean} block averages each (reward, metric) cell across the three models. \textbf{Bold} = column-best reward function per row.}
\label{tab:grpo-app-rexgradient-findings-rouge}
\centering
\footnotesize
\setlength{\tabcolsep}{3pt}
\renewcommand{\arraystretch}{0.95}
\begin{tabular}{llccccc}
\toprule
Model & Metric & SFT & $R_{\mathrm{fmt}}$ & $R_{\mathrm{exact}}$ & $R_{\mathrm{Cham}}$ & $R_{\mathrm{Haus}}$ \\
\midrule
\multirow{3}{*}{Qwen3-VL-2B} & R1-F & 0.374\,{\scriptsize $\pm$0.006} & 0.377\,{\scriptsize $\pm$0.002} & 0.375\,{\scriptsize $\pm$0.003} & \textbf{0.409}\,{\scriptsize $\pm$0.002} & 0.381\,{\scriptsize $\pm$0.003} \\
 & R2-F & 0.208\,{\scriptsize $\pm$0.006} & 0.210\,{\scriptsize $\pm$0.004} & 0.205\,{\scriptsize $\pm$0.003} & 0.188\,{\scriptsize $\pm$0.002} & \textbf{0.213}\,{\scriptsize $\pm$0.003} \\
 & RL-F & 0.312\,{\scriptsize $\pm$0.006} & \textbf{0.314}\,{\scriptsize $\pm$0.002} & 0.310\,{\scriptsize $\pm$0.003} & 0.305\,{\scriptsize $\pm$0.002} & \textbf{0.314}\,{\scriptsize $\pm$0.003} \\
\midrule
\multirow{3}{*}{Qwen3-VL-4B} & R1-F & 0.370\,{\scriptsize $\pm$0.001} & 0.372\,{\scriptsize $\pm$0.002} & 0.375\,{\scriptsize $\pm$0.002} & \textbf{0.425}\,{\scriptsize $\pm$0.001} & 0.386\,{\scriptsize $\pm$0.002} \\
 & R2-F & 0.196\,{\scriptsize $\pm$0.003} & 0.196\,{\scriptsize $\pm$0.003} & 0.206\,{\scriptsize $\pm$0.001} & \textbf{0.229}\,{\scriptsize $\pm$0.001} & 0.179\,{\scriptsize $\pm$0.002} \\
 & RL-F & 0.302\,{\scriptsize $\pm$0.002} & 0.303\,{\scriptsize $\pm$0.003} & 0.309\,{\scriptsize $\pm$0.001} & \textbf{0.341}\,{\scriptsize $\pm$0.000} & 0.293\,{\scriptsize $\pm$0.002} \\
\midrule
\multirow{3}{*}{Gemma3-4B} & R1-F & 0.364\,{\scriptsize $\pm$0.005} & 0.371\,{\scriptsize $\pm$0.003} & 0.383\,{\scriptsize $\pm$0.003} & \textbf{0.387}\,{\scriptsize $\pm$0.002} & 0.381\,{\scriptsize $\pm$0.002} \\
\bottomrule
\end{tabular}
\end{table}

\begin{table}[H]
\caption{\textbf{GRPO post-training results on ReXGradient (Findings).} Embedding-based / lexical (METEOR, BERTScore F$_1$, COMET, ChrF$++$). Mean over 5 random seeds with sample std in \scriptsize. The final \emph{Mean} block averages each (reward, metric) cell across the three models. \textbf{Bold} = column-best reward function per row.}
\label{tab:grpo-app-rexgradient-findings-embedding}
\centering
\footnotesize
\setlength{\tabcolsep}{3pt}
\renewcommand{\arraystretch}{0.95}
\begin{tabular}{llccccc}
\toprule
Model & Metric & SFT & $R_{\mathrm{fmt}}$ & $R_{\mathrm{exact}}$ & $R_{\mathrm{Cham}}$ & $R_{\mathrm{Haus}}$ \\
\midrule
\multirow{4}{*}{Qwen3-VL-2B} & METEOR & 0.324\,{\scriptsize $\pm$0.005} & 0.326\,{\scriptsize $\pm$0.003} & 0.326\,{\scriptsize $\pm$0.004} & \textbf{0.356}\,{\scriptsize $\pm$0.002} & 0.333\,{\scriptsize $\pm$0.002} \\
 & BERTScore F1 & 0.339\,{\scriptsize $\pm$0.004} & 0.342\,{\scriptsize $\pm$0.002} & 0.332\,{\scriptsize $\pm$0.002} & \textbf{0.367}\,{\scriptsize $\pm$0.002} & 0.339\,{\scriptsize $\pm$0.003} \\
 & COMET & 0.619\,{\scriptsize $\pm$0.002} & 0.620\,{\scriptsize $\pm$0.001} & 0.619\,{\scriptsize $\pm$0.001} & \textbf{0.657}\,{\scriptsize $\pm$0.001} & 0.623\,{\scriptsize $\pm$0.001} \\
 & ChrF++ & 30.727\,{\scriptsize $\pm$0.382} & 30.846\,{\scriptsize $\pm$0.245} & 31.041\,{\scriptsize $\pm$0.297} & \textbf{32.875}\,{\scriptsize $\pm$0.163} & 31.802\,{\scriptsize $\pm$0.268} \\
\midrule
\multirow{4}{*}{Qwen3-VL-4B} & METEOR & 0.322\,{\scriptsize $\pm$0.001} & 0.322\,{\scriptsize $\pm$0.002} & 0.330\,{\scriptsize $\pm$0.002} & \textbf{0.381}\,{\scriptsize $\pm$0.001} & 0.300\,{\scriptsize $\pm$0.002} \\
 & BERTScore F1 & 0.332\,{\scriptsize $\pm$0.002} & 0.333\,{\scriptsize $\pm$0.003} & 0.331\,{\scriptsize $\pm$0.002} & \textbf{0.351}\,{\scriptsize $\pm$0.000} & 0.333\,{\scriptsize $\pm$0.002} \\
 & COMET & 0.620\,{\scriptsize $\pm$0.001} & 0.618\,{\scriptsize $\pm$0.001} & 0.617\,{\scriptsize $\pm$0.001} & \textbf{0.677}\,{\scriptsize $\pm$0.000} & 0.572\,{\scriptsize $\pm$0.001} \\
\bottomrule
\end{tabular}
\end{table}

\begin{table}[H]
\caption{\textbf{GRPO post-training results on ReXGradient (Findings).} Clinical (RadGraph, CheXbert). Mean over 5 random seeds with sample std in \scriptsize. The final \emph{Mean} block averages each (reward, metric) cell across the three models. \textbf{Bold} = column-best reward function per row.}
\label{tab:grpo-app-rexgradient-findings-clinical}
\centering
\footnotesize
\setlength{\tabcolsep}{3pt}
\renewcommand{\arraystretch}{0.95}
\begin{tabular}{llccccc}
\toprule
Model & Metric & SFT & $R_{\mathrm{fmt}}$ & $R_{\mathrm{exact}}$ & $R_{\mathrm{Cham}}$ & $R_{\mathrm{Haus}}$ \\
\midrule
\multirow{4}{*}{Qwen3-VL-2B} & RG-E & 0.295\,{\scriptsize $\pm$0.006} & 0.299\,{\scriptsize $\pm$0.003} & 0.293\,{\scriptsize $\pm$0.002} & 0.303\,{\scriptsize $\pm$0.002} & \textbf{0.305}\,{\scriptsize $\pm$0.003} \\
 & RG-ER & 0.280\,{\scriptsize $\pm$0.006} & 0.284\,{\scriptsize $\pm$0.004} & 0.277\,{\scriptsize $\pm$0.002} & 0.285\,{\scriptsize $\pm$0.002} & \textbf{0.290}\,{\scriptsize $\pm$0.003} \\
 & RadGraph F1 & 0.247\,{\scriptsize $\pm$0.006} & 0.251\,{\scriptsize $\pm$0.003} & 0.244\,{\scriptsize $\pm$0.002} & 0.224\,{\scriptsize $\pm$0.003} & \textbf{0.258}\,{\scriptsize $\pm$0.003} \\
 & CheXbert F1 & 0.705\,{\scriptsize $\pm$0.003} & 0.708\,{\scriptsize $\pm$0.007} & 0.663\,{\scriptsize $\pm$0.008} & \textbf{0.734}\,{\scriptsize $\pm$0.000} & 0.686\,{\scriptsize $\pm$0.009} \\
\midrule
\multirow{4}{*}{Qwen3-VL-4B} & RG-E & 0.288\,{\scriptsize $\pm$0.003} & 0.289\,{\scriptsize $\pm$0.002} & 0.292\,{\scriptsize $\pm$0.002} & \textbf{0.344}\,{\scriptsize $\pm$0.000} & 0.288\,{\scriptsize $\pm$0.002} \\
 & RG-ER & 0.273\,{\scriptsize $\pm$0.003} & 0.274\,{\scriptsize $\pm$0.003} & 0.278\,{\scriptsize $\pm$0.002} & \textbf{0.310}\,{\scriptsize $\pm$0.000} & 0.278\,{\scriptsize $\pm$0.002} \\
 & RadGraph F1 & 0.237\,{\scriptsize $\pm$0.004} & 0.239\,{\scriptsize $\pm$0.004} & 0.242\,{\scriptsize $\pm$0.001} & \textbf{0.282}\,{\scriptsize $\pm$0.000} & 0.242\,{\scriptsize $\pm$0.002} \\
\bottomrule
\end{tabular}
\end{table}

\subsubsection{Impression}

\begin{table}[H]
\caption{\textbf{GRPO post-training results on ReXGradient (Impression).} BLEU (sentence \& corpus, $n=1\dots 4$). Mean over 5 random seeds with sample std in \scriptsize. The final \emph{Mean} block averages each (reward, metric) cell across the three models. \textbf{Bold} = column-best reward function per row.}
\label{tab:grpo-app-rexgradient-impression-bleu}
\centering
\footnotesize
\setlength{\tabcolsep}{3pt}
\renewcommand{\arraystretch}{0.95}
\resizebox{\linewidth}{!}{%
\begin{tabular}{llccccc}
\toprule
Model & Metric & SFT & $R_{\mathrm{fmt}}$ & $R_{\mathrm{exact}}$ & $R_{\mathrm{Cham}}$ & $R_{\mathrm{Haus}}$ \\
\midrule
\multirow{8}{*}{Qwen3-VL-2B} & S-BLEU-1 & 0.317\,{\scriptsize $\pm$0.003} & 0.321\,{\scriptsize $\pm$0.003} & 0.319\,{\scriptsize $\pm$0.003} & 0.313\,{\scriptsize $\pm$0.000} & \textbf{0.328}\,{\scriptsize $\pm$0.003} \\
 & S-BLEU-2 & 0.231\,{\scriptsize $\pm$0.003} & 0.234\,{\scriptsize $\pm$0.004} & 0.228\,{\scriptsize $\pm$0.004} & 0.213\,{\scriptsize $\pm$0.001} & \textbf{0.236}\,{\scriptsize $\pm$0.003} \\
 & S-BLEU-3 & 0.202\,{\scriptsize $\pm$0.003} & 0.204\,{\scriptsize $\pm$0.005} & 0.198\,{\scriptsize $\pm$0.004} & 0.144\,{\scriptsize $\pm$0.004} & \textbf{0.206}\,{\scriptsize $\pm$0.003} \\
 & S-BLEU-4 & 0.189\,{\scriptsize $\pm$0.003} & 0.191\,{\scriptsize $\pm$0.005} & 0.184\,{\scriptsize $\pm$0.004} & 0.129\,{\scriptsize $\pm$0.004} & \textbf{0.192}\,{\scriptsize $\pm$0.002} \\
 & C-BLEU-1 & 0.136\,{\scriptsize $\pm$0.002} & 0.136\,{\scriptsize $\pm$0.003} & 0.227\,{\scriptsize $\pm$0.003} & 0.113\,{\scriptsize $\pm$0.000} & \textbf{0.237}\,{\scriptsize $\pm$0.007} \\
 & C-BLEU-2 & 0.091\,{\scriptsize $\pm$0.001} & 0.092\,{\scriptsize $\pm$0.002} & 0.142\,{\scriptsize $\pm$0.003} & 0.072\,{\scriptsize $\pm$0.000} & \textbf{0.150}\,{\scriptsize $\pm$0.005} \\
 & C-BLEU-3 & 0.072\,{\scriptsize $\pm$0.001} & 0.073\,{\scriptsize $\pm$0.002} & 0.105\,{\scriptsize $\pm$0.003} & 0.049\,{\scriptsize $\pm$0.001} & \textbf{0.111}\,{\scriptsize $\pm$0.004} \\
\bottomrule
\end{tabular}
}
\end{table}

\begin{table}[H]
\caption{\textbf{GRPO post-training results on ReXGradient (Impression).} ROUGE (F$_1$ for $n=1$, $n=2$, $L$). Mean over 5 random seeds with sample std in \scriptsize. The final \emph{Mean} block averages each (reward, metric) cell across the three models. \textbf{Bold} = column-best reward function per row.}
\label{tab:grpo-app-rexgradient-impression-rouge}
\centering
\footnotesize
\setlength{\tabcolsep}{3pt}
\renewcommand{\arraystretch}{0.95}
\begin{tabular}{llccccc}
\toprule
Model & Metric & SFT & $R_{\mathrm{fmt}}$ & $R_{\mathrm{exact}}$ & $R_{\mathrm{Cham}}$ & $R_{\mathrm{Haus}}$ \\
\midrule
\multirow{3}{*}{Qwen3-VL-2B} & R1-F & 0.306\,{\scriptsize $\pm$0.004} & 0.309\,{\scriptsize $\pm$0.004} & 0.305\,{\scriptsize $\pm$0.004} & 0.303\,{\scriptsize $\pm$0.001} & \textbf{0.319}\,{\scriptsize $\pm$0.004} \\
 & R2-F & 0.190\,{\scriptsize $\pm$0.004} & 0.193\,{\scriptsize $\pm$0.005} & 0.189\,{\scriptsize $\pm$0.005} & 0.139\,{\scriptsize $\pm$0.003} & \textbf{0.196}\,{\scriptsize $\pm$0.003} \\
 & RL-F & 0.303\,{\scriptsize $\pm$0.004} & 0.307\,{\scriptsize $\pm$0.004} & 0.299\,{\scriptsize $\pm$0.005} & 0.300\,{\scriptsize $\pm$0.001} & \textbf{0.312}\,{\scriptsize $\pm$0.004} \\
\midrule
\multirow{3}{*}{Qwen3-VL-4B} & R1-F & 0.299\,{\scriptsize $\pm$0.003} & 0.300\,{\scriptsize $\pm$0.006} & 0.307\,{\scriptsize $\pm$0.002} & \textbf{0.341}\,{\scriptsize $\pm$0.000} & 0.314\,{\scriptsize $\pm$0.002} \\
 & R2-F & 0.178\,{\scriptsize $\pm$0.005} & 0.178\,{\scriptsize $\pm$0.005} & 0.186\,{\scriptsize $\pm$0.003} & \textbf{0.215}\,{\scriptsize $\pm$0.000} & 0.152\,{\scriptsize $\pm$0.002} \\
 & RL-F & 0.296\,{\scriptsize $\pm$0.003} & 0.297\,{\scriptsize $\pm$0.006} & 0.301\,{\scriptsize $\pm$0.002} & \textbf{0.340}\,{\scriptsize $\pm$0.000} & 0.311\,{\scriptsize $\pm$0.002} \\
\midrule
\multirow{3}{*}{Gemma3-4B} & R1-F & 0.262\,{\scriptsize $\pm$0.004} & 0.262\,{\scriptsize $\pm$0.005} & 0.310\,{\scriptsize $\pm$0.004} & \textbf{0.311}\,{\scriptsize $\pm$0.003} & 0.290\,{\scriptsize $\pm$0.003} \\
\bottomrule
\end{tabular}
\end{table}

\begin{table}[H]
\caption{\textbf{GRPO post-training results on ReXGradient (Impression).} Embedding-based / lexical (METEOR, BERTScore F$_1$, COMET, ChrF$++$). Mean over 5 random seeds with sample std in \scriptsize. The final \emph{Mean} block averages each (reward, metric) cell across the three models. \textbf{Bold} = column-best reward function per row.}
\label{tab:grpo-app-rexgradient-impression-embedding}
\centering
\footnotesize
\setlength{\tabcolsep}{3pt}
\renewcommand{\arraystretch}{0.95}
\begin{tabular}{llccccc}
\toprule
Model & Metric & SFT & $R_{\mathrm{fmt}}$ & $R_{\mathrm{exact}}$ & $R_{\mathrm{Cham}}$ & $R_{\mathrm{Haus}}$ \\
\midrule
\multirow{4}{*}{Qwen3-VL-2B} & METEOR & 0.297\,{\scriptsize $\pm$0.003} & 0.301\,{\scriptsize $\pm$0.004} & 0.308\,{\scriptsize $\pm$0.005} & 0.289\,{\scriptsize $\pm$0.001} & \textbf{0.314}\,{\scriptsize $\pm$0.004} \\
 & BERTScore F1 & 0.324\,{\scriptsize $\pm$0.004} & 0.326\,{\scriptsize $\pm$0.006} & 0.322\,{\scriptsize $\pm$0.004} & \textbf{0.336}\,{\scriptsize $\pm$0.002} & 0.334\,{\scriptsize $\pm$0.003} \\
 & COMET & 0.592\,{\scriptsize $\pm$0.004} & 0.594\,{\scriptsize $\pm$0.003} & 0.594\,{\scriptsize $\pm$0.002} & 0.599\,{\scriptsize $\pm$0.001} & \textbf{0.601}\,{\scriptsize $\pm$0.003} \\
 & ChrF++ & 19.531\,{\scriptsize $\pm$0.209} & 19.701\,{\scriptsize $\pm$0.255} & 22.081\,{\scriptsize $\pm$0.338} & 15.199\,{\scriptsize $\pm$0.145} & \textbf{22.903}\,{\scriptsize $\pm$0.409} \\
\midrule
\multirow{4}{*}{Qwen3-VL-4B} & METEOR & 0.293\,{\scriptsize $\pm$0.003} & 0.294\,{\scriptsize $\pm$0.004} & 0.312\,{\scriptsize $\pm$0.003} & \textbf{0.328}\,{\scriptsize $\pm$0.001} & 0.301\,{\scriptsize $\pm$0.001} \\
 & BERTScore F1 & 0.322\,{\scriptsize $\pm$0.003} & 0.322\,{\scriptsize $\pm$0.005} & 0.324\,{\scriptsize $\pm$0.002} & 0.343\,{\scriptsize $\pm$0.001} & \textbf{0.347}\,{\scriptsize $\pm$0.002} \\
 & COMET & 0.589\,{\scriptsize $\pm$0.001} & 0.587\,{\scriptsize $\pm$0.003} & 0.596\,{\scriptsize $\pm$0.001} & \textbf{0.611}\,{\scriptsize $\pm$0.000} & 0.604\,{\scriptsize $\pm$0.001} \\
\bottomrule
\end{tabular}
\end{table}

\begin{table}[H]
\caption{\textbf{GRPO post-training results on ReXGradient (Impression).} Clinical (RadGraph, CheXbert). Mean over 5 random seeds with sample std in \scriptsize. The final \emph{Mean} block averages each (reward, metric) cell across the three models. \textbf{Bold} = column-best reward function per row.}
\label{tab:grpo-app-rexgradient-impression-clinical}
\centering
\footnotesize
\setlength{\tabcolsep}{3pt}
\renewcommand{\arraystretch}{0.95}
\begin{tabular}{llccccc}
\toprule
Model & Metric & SFT & $R_{\mathrm{fmt}}$ & $R_{\mathrm{exact}}$ & $R_{\mathrm{Cham}}$ & $R_{\mathrm{Haus}}$ \\
\midrule
\multirow{4}{*}{Qwen3-VL-2B} & RG-E & 0.252\,{\scriptsize $\pm$0.005} & 0.255\,{\scriptsize $\pm$0.005} & 0.251\,{\scriptsize $\pm$0.004} & 0.233\,{\scriptsize $\pm$0.001} & \textbf{0.258}\,{\scriptsize $\pm$0.006} \\
 & RG-ER & 0.237\,{\scriptsize $\pm$0.006} & 0.239\,{\scriptsize $\pm$0.005} & 0.237\,{\scriptsize $\pm$0.004} & 0.214\,{\scriptsize $\pm$0.001} & \textbf{0.247}\,{\scriptsize $\pm$0.005} \\
 & RadGraph F1 & 0.224\,{\scriptsize $\pm$0.004} & 0.225\,{\scriptsize $\pm$0.006} & 0.220\,{\scriptsize $\pm$0.005} & 0.160\,{\scriptsize $\pm$0.003} & \textbf{0.230}\,{\scriptsize $\pm$0.005} \\
 & CheXbert F1 & 0.755\,{\scriptsize $\pm$0.004} & 0.760\,{\scriptsize $\pm$0.006} & 0.705\,{\scriptsize $\pm$0.009} & \textbf{0.778}\,{\scriptsize $\pm$0.000} & 0.733\,{\scriptsize $\pm$0.010} \\
\midrule
\multirow{4}{*}{Qwen3-VL-4B} & RG-E & 0.246\,{\scriptsize $\pm$0.003} & 0.247\,{\scriptsize $\pm$0.006} & 0.255\,{\scriptsize $\pm$0.003} & \textbf{0.291}\,{\scriptsize $\pm$0.000} & 0.247\,{\scriptsize $\pm$0.002} \\
 & RG-ER & 0.225\,{\scriptsize $\pm$0.004} & 0.225\,{\scriptsize $\pm$0.004} & 0.239\,{\scriptsize $\pm$0.003} & \textbf{0.268}\,{\scriptsize $\pm$0.000} & 0.224\,{\scriptsize $\pm$0.002} \\
 & RadGraph F1 & 0.211\,{\scriptsize $\pm$0.003} & 0.210\,{\scriptsize $\pm$0.005} & 0.223\,{\scriptsize $\pm$0.003} & \textbf{0.257}\,{\scriptsize $\pm$0.000} & 0.174\,{\scriptsize $\pm$0.002} \\
\bottomrule
\end{tabular}
\end{table}

\subsection{MIMIC-CXR}


\subsubsection{Findings}

\begin{table}[H]
\caption{\textbf{GRPO post-training results on MIMIC-CXR (Findings).} BLEU (sentence \& corpus, $n=1\dots 4$). Mean over 5 random seeds with sample std in \scriptsize. The final \emph{Mean} block averages each (reward, metric) cell across the three models. \textbf{Bold} = column-best reward function per row.}
\label{tab:grpo-app-mimiccxr-findings-bleu}
\centering
\footnotesize
\setlength{\tabcolsep}{3pt}
\renewcommand{\arraystretch}{0.95}
\resizebox{\linewidth}{!}{%
\begin{tabular}{llccccc}
\toprule
Model & Metric & SFT & $R_{\mathrm{fmt}}$ & $R_{\mathrm{exact}}$ & $R_{\mathrm{Cham}}$ & $R_{\mathrm{Haus}}$ \\
\midrule
\multirow{8}{*}{Qwen3-VL-2B} & S-BLEU-1 & 0.338\,{\scriptsize $\pm$0.004} & 0.328\,{\scriptsize $\pm$0.003} & 0.301\,{\scriptsize $\pm$0.002} & 0.252\,{\scriptsize $\pm$0.001} & \textbf{0.382}\,{\scriptsize $\pm$0.002} \\
 & S-BLEU-2 & 0.222\,{\scriptsize $\pm$0.004} & 0.214\,{\scriptsize $\pm$0.003} & 0.184\,{\scriptsize $\pm$0.001} & 0.161\,{\scriptsize $\pm$0.000} & \textbf{0.251}\,{\scriptsize $\pm$0.002} \\
 & S-BLEU-3 & 0.155\,{\scriptsize $\pm$0.004} & 0.149\,{\scriptsize $\pm$0.004} & 0.122\,{\scriptsize $\pm$0.001} & 0.110\,{\scriptsize $\pm$0.000} & \textbf{0.175}\,{\scriptsize $\pm$0.002} \\
 & S-BLEU-4 & 0.113\,{\scriptsize $\pm$0.003} & 0.108\,{\scriptsize $\pm$0.004} & 0.085\,{\scriptsize $\pm$0.001} & 0.077\,{\scriptsize $\pm$0.000} & \textbf{0.124}\,{\scriptsize $\pm$0.002} \\
 & C-BLEU-1 & 0.353\,{\scriptsize $\pm$0.005} & 0.339\,{\scriptsize $\pm$0.002} & 0.299\,{\scriptsize $\pm$0.002} & 0.299\,{\scriptsize $\pm$0.002} & \textbf{0.397}\,{\scriptsize $\pm$0.002} \\
 & C-BLEU-2 & 0.231\,{\scriptsize $\pm$0.005} & 0.222\,{\scriptsize $\pm$0.002} & 0.183\,{\scriptsize $\pm$0.001} & 0.191\,{\scriptsize $\pm$0.001} & \textbf{0.260}\,{\scriptsize $\pm$0.003} \\
 & C-BLEU-3 & 0.164\,{\scriptsize $\pm$0.004} & 0.157\,{\scriptsize $\pm$0.002} & 0.123\,{\scriptsize $\pm$0.001} & 0.130\,{\scriptsize $\pm$0.001} & \textbf{0.182}\,{\scriptsize $\pm$0.002} \\
\bottomrule
\end{tabular}
}
\end{table}

\begin{table}[H]
\caption{\textbf{GRPO post-training results on MIMIC-CXR (Findings).} ROUGE (F$_1$ for $n=1$, $n=2$, $L$). Mean over 5 random seeds with sample std in \scriptsize. The final \emph{Mean} block averages each (reward, metric) cell across the three models. \textbf{Bold} = column-best reward function per row.}
\label{tab:grpo-app-mimiccxr-findings-rouge}
\centering
\footnotesize
\setlength{\tabcolsep}{3pt}
\renewcommand{\arraystretch}{0.95}
\begin{tabular}{llccccc}
\toprule
Model & Metric & SFT & $R_{\mathrm{fmt}}$ & $R_{\mathrm{exact}}$ & $R_{\mathrm{Cham}}$ & $R_{\mathrm{Haus}}$ \\
\midrule
\multirow{3}{*}{Qwen3-VL-2B} & R1-F & 0.408\,{\scriptsize $\pm$0.003} & 0.403\,{\scriptsize $\pm$0.001} & 0.372\,{\scriptsize $\pm$0.001} & 0.396\,{\scriptsize $\pm$0.001} & \textbf{0.445}\,{\scriptsize $\pm$0.002} \\
 & R2-F & 0.172\,{\scriptsize $\pm$0.003} & 0.168\,{\scriptsize $\pm$0.003} & 0.145\,{\scriptsize $\pm$0.001} & 0.162\,{\scriptsize $\pm$0.000} & \textbf{0.184}\,{\scriptsize $\pm$0.002} \\
 & RL-F & 0.294\,{\scriptsize $\pm$0.002} & 0.292\,{\scriptsize $\pm$0.003} & 0.279\,{\scriptsize $\pm$0.002} & 0.275\,{\scriptsize $\pm$0.000} & \textbf{0.295}\,{\scriptsize $\pm$0.003} \\
\midrule
\multirow{3}{*}{Qwen3-VL-4B} & R1-F & 0.405\,{\scriptsize $\pm$0.002} & 0.401\,{\scriptsize $\pm$0.002} & 0.330\,{\scriptsize $\pm$0.002} & 0.398\,{\scriptsize $\pm$0.001} & \textbf{0.438}\,{\scriptsize $\pm$0.001} \\
 & R2-F & 0.170\,{\scriptsize $\pm$0.003} & 0.169\,{\scriptsize $\pm$0.003} & 0.125\,{\scriptsize $\pm$0.001} & 0.194\,{\scriptsize $\pm$0.001} & \textbf{0.204}\,{\scriptsize $\pm$0.001} \\
 & RL-F & 0.292\,{\scriptsize $\pm$0.002} & 0.294\,{\scriptsize $\pm$0.002} & 0.250\,{\scriptsize $\pm$0.001} & 0.300\,{\scriptsize $\pm$0.001} & \textbf{0.310}\,{\scriptsize $\pm$0.001} \\
\midrule
\multirow{3}{*}{Gemma3-4B} & R1-F & 0.422\,{\scriptsize $\pm$0.002} & 0.424\,{\scriptsize $\pm$0.005} & 0.416\,{\scriptsize $\pm$0.003} & \textbf{0.433}\,{\scriptsize $\pm$0.003} & 0.426\,{\scriptsize $\pm$0.002} \\
\bottomrule
\end{tabular}
\end{table}

\begin{table}[H]
\caption{\textbf{GRPO post-training results on MIMIC-CXR (Findings).} Embedding-based / lexical (METEOR, BERTScore F$_1$, COMET, ChrF$++$). Mean over 5 random seeds with sample std in \scriptsize. The final \emph{Mean} block averages each (reward, metric) cell across the three models. \textbf{Bold} = column-best reward function per row.}
\label{tab:grpo-app-mimiccxr-findings-embedding}
\centering
\footnotesize
\setlength{\tabcolsep}{3pt}
\renewcommand{\arraystretch}{0.95}
\begin{tabular}{llccccc}
\toprule
Model & Metric & SFT & $R_{\mathrm{fmt}}$ & $R_{\mathrm{exact}}$ & $R_{\mathrm{Cham}}$ & $R_{\mathrm{Haus}}$ \\
\midrule
\multirow{4}{*}{Qwen3-VL-2B} & METEOR & 0.315\,{\scriptsize $\pm$0.004} & 0.308\,{\scriptsize $\pm$0.003} & 0.276\,{\scriptsize $\pm$0.001} & 0.277\,{\scriptsize $\pm$0.001} & \textbf{0.341}\,{\scriptsize $\pm$0.003} \\
 & BERTScore F1 & 0.313\,{\scriptsize $\pm$0.003} & 0.312\,{\scriptsize $\pm$0.003} & 0.276\,{\scriptsize $\pm$0.002} & 0.320\,{\scriptsize $\pm$0.001} & \textbf{0.335}\,{\scriptsize $\pm$0.002} \\
 & COMET & 0.621\,{\scriptsize $\pm$0.002} & 0.618\,{\scriptsize $\pm$0.002} & 0.602\,{\scriptsize $\pm$0.001} & 0.604\,{\scriptsize $\pm$0.000} & \textbf{0.635}\,{\scriptsize $\pm$0.001} \\
 & ChrF++ & 32.972\,{\scriptsize $\pm$0.373} & 32.329\,{\scriptsize $\pm$0.201} & 29.064\,{\scriptsize $\pm$0.163} & 29.808\,{\scriptsize $\pm$0.074} & \textbf{34.595}\,{\scriptsize $\pm$0.245} \\
\midrule
\multirow{4}{*}{Qwen3-VL-4B} & METEOR & 0.315\,{\scriptsize $\pm$0.002} & 0.308\,{\scriptsize $\pm$0.001} & 0.226\,{\scriptsize $\pm$0.001} & 0.274\,{\scriptsize $\pm$0.001} & \textbf{0.362}\,{\scriptsize $\pm$0.001} \\
 & BERTScore F1 & 0.308\,{\scriptsize $\pm$0.004} & 0.310\,{\scriptsize $\pm$0.004} & 0.263\,{\scriptsize $\pm$0.002} & 0.341\,{\scriptsize $\pm$0.001} & \textbf{0.343}\,{\scriptsize $\pm$0.001} \\
 & COMET & 0.619\,{\scriptsize $\pm$0.002} & 0.619\,{\scriptsize $\pm$0.001} & 0.574\,{\scriptsize $\pm$0.001} & 0.606\,{\scriptsize $\pm$0.001} & \textbf{0.635}\,{\scriptsize $\pm$0.000} \\
\bottomrule
\end{tabular}
\end{table}

\begin{table}[H]
\caption{\textbf{GRPO post-training results on MIMIC-CXR (Findings).} Clinical (RadGraph, CheXbert). Mean over 5 random seeds with sample std in \scriptsize. The final \emph{Mean} block averages each (reward, metric) cell across the three models. \textbf{Bold} = column-best reward function per row.}
\label{tab:grpo-app-mimiccxr-findings-clinical}
\centering
\footnotesize
\setlength{\tabcolsep}{3pt}
\renewcommand{\arraystretch}{0.95}
\begin{tabular}{llccccc}
\toprule
Model & Metric & SFT & $R_{\mathrm{fmt}}$ & $R_{\mathrm{exact}}$ & $R_{\mathrm{Cham}}$ & $R_{\mathrm{Haus}}$ \\
\midrule
\multirow{4}{*}{Qwen3-VL-2B} & RG-E & 0.279\,{\scriptsize $\pm$0.001} & 0.277\,{\scriptsize $\pm$0.005} & 0.270\,{\scriptsize $\pm$0.003} & 0.276\,{\scriptsize $\pm$0.001} & \textbf{0.333}\,{\scriptsize $\pm$0.003} \\
 & RG-ER & 0.252\,{\scriptsize $\pm$0.001} & 0.251\,{\scriptsize $\pm$0.004} & 0.204\,{\scriptsize $\pm$0.003} & 0.258\,{\scriptsize $\pm$0.001} & \textbf{0.295}\,{\scriptsize $\pm$0.003} \\
 & RadGraph F1 & 0.198\,{\scriptsize $\pm$0.001} & 0.197\,{\scriptsize $\pm$0.004} & 0.153\,{\scriptsize $\pm$0.002} & 0.191\,{\scriptsize $\pm$0.001} & \textbf{0.231}\,{\scriptsize $\pm$0.003} \\
 & CheXbert F1 & 0.499\,{\scriptsize $\pm$0.007} & 0.502\,{\scriptsize $\pm$0.006} & 0.469\,{\scriptsize $\pm$0.008} & 0.481\,{\scriptsize $\pm$0.003} & \textbf{0.508}\,{\scriptsize $\pm$0.005} \\
\midrule
\multirow{4}{*}{Qwen3-VL-4B} & RG-E & 0.277\,{\scriptsize $\pm$0.003} & 0.275\,{\scriptsize $\pm$0.003} & 0.225\,{\scriptsize $\pm$0.001} & \textbf{0.328}\,{\scriptsize $\pm$0.001} & 0.324\,{\scriptsize $\pm$0.001} \\
 & RG-ER & 0.250\,{\scriptsize $\pm$0.004} & 0.249\,{\scriptsize $\pm$0.005} & 0.214\,{\scriptsize $\pm$0.001} & \textbf{0.304}\,{\scriptsize $\pm$0.001} & 0.284\,{\scriptsize $\pm$0.000} \\
 & RadGraph F1 & 0.195\,{\scriptsize $\pm$0.004} & 0.197\,{\scriptsize $\pm$0.004} & 0.157\,{\scriptsize $\pm$0.002} & \textbf{0.240}\,{\scriptsize $\pm$0.001} & 0.212\,{\scriptsize $\pm$0.002} \\
\bottomrule
\end{tabular}
\end{table}

\subsubsection{Impression}

\begin{table}[H]
\caption{\textbf{GRPO post-training results on MIMIC-CXR (Impression).} BLEU (sentence \& corpus, $n=1\dots 4$). Mean over 5 random seeds with sample std in \scriptsize. The final \emph{Mean} block averages each (reward, metric) cell across the three models. \textbf{Bold} = column-best reward function per row.}
\label{tab:grpo-app-mimiccxr-impression-bleu}
\centering
\footnotesize
\setlength{\tabcolsep}{3pt}
\renewcommand{\arraystretch}{0.95}
\resizebox{\linewidth}{!}{%
\begin{tabular}{llccccc}
\toprule
Model & Metric & SFT & $R_{\mathrm{fmt}}$ & $R_{\mathrm{exact}}$ & $R_{\mathrm{Cham}}$ & $R_{\mathrm{Haus}}$ \\
\midrule
\multirow{8}{*}{Qwen3-VL-2B} & S-BLEU-1 & 0.321\,{\scriptsize $\pm$0.001} & 0.322\,{\scriptsize $\pm$0.003} & \textbf{0.352}\,{\scriptsize $\pm$0.002} & 0.297\,{\scriptsize $\pm$0.001} & 0.310\,{\scriptsize $\pm$0.003} \\
 & S-BLEU-2 & 0.239\,{\scriptsize $\pm$0.001} & 0.239\,{\scriptsize $\pm$0.004} & \textbf{0.273}\,{\scriptsize $\pm$0.002} & 0.191\,{\scriptsize $\pm$0.001} & 0.214\,{\scriptsize $\pm$0.003} \\
 & S-BLEU-3 & 0.189\,{\scriptsize $\pm$0.002} & 0.187\,{\scriptsize $\pm$0.006} & \textbf{0.228}\,{\scriptsize $\pm$0.002} & 0.140\,{\scriptsize $\pm$0.000} & 0.160\,{\scriptsize $\pm$0.003} \\
 & S-BLEU-4 & 0.168\,{\scriptsize $\pm$0.003} & 0.166\,{\scriptsize $\pm$0.006} & \textbf{0.207}\,{\scriptsize $\pm$0.001} & 0.104\,{\scriptsize $\pm$0.000} & 0.115\,{\scriptsize $\pm$0.002} \\
 & C-BLEU-1 & 0.152\,{\scriptsize $\pm$0.004} & 0.137\,{\scriptsize $\pm$0.005} & \textbf{0.248}\,{\scriptsize $\pm$0.002} & 0.231\,{\scriptsize $\pm$0.003} & 0.135\,{\scriptsize $\pm$0.003} \\
 & C-BLEU-2 & 0.099\,{\scriptsize $\pm$0.002} & 0.090\,{\scriptsize $\pm$0.004} & \textbf{0.160}\,{\scriptsize $\pm$0.002} & 0.127\,{\scriptsize $\pm$0.002} & 0.082\,{\scriptsize $\pm$0.002} \\
 & C-BLEU-3 & 0.072\,{\scriptsize $\pm$0.002} & 0.066\,{\scriptsize $\pm$0.003} & \textbf{0.115}\,{\scriptsize $\pm$0.001} & 0.080\,{\scriptsize $\pm$0.001} & 0.056\,{\scriptsize $\pm$0.002} \\
\bottomrule
\end{tabular}
}
\end{table}

\begin{table}[H]
\caption{\textbf{GRPO post-training results on MIMIC-CXR (Impression).} ROUGE (F$_1$ for $n=1$, $n=2$, $L$). Mean over 5 random seeds with sample std in \scriptsize. The final \emph{Mean} block averages each (reward, metric) cell across the three models. \textbf{Bold} = column-best reward function per row.}
\label{tab:grpo-app-mimiccxr-impression-rouge}
\centering
\footnotesize
\setlength{\tabcolsep}{3pt}
\renewcommand{\arraystretch}{0.95}
\begin{tabular}{llccccc}
\toprule
Model & Metric & SFT & $R_{\mathrm{fmt}}$ & $R_{\mathrm{exact}}$ & $R_{\mathrm{Cham}}$ & $R_{\mathrm{Haus}}$ \\
\midrule
\multirow{3}{*}{Qwen3-VL-2B} & R1-F & 0.330\,{\scriptsize $\pm$0.002} & 0.331\,{\scriptsize $\pm$0.003} & \textbf{0.359}\,{\scriptsize $\pm$0.002} & 0.316\,{\scriptsize $\pm$0.001} & 0.313\,{\scriptsize $\pm$0.004} \\
 & R2-F & 0.212\,{\scriptsize $\pm$0.002} & 0.212\,{\scriptsize $\pm$0.006} & \textbf{0.252}\,{\scriptsize $\pm$0.002} & 0.198\,{\scriptsize $\pm$0.001} & 0.209\,{\scriptsize $\pm$0.003} \\
 & RL-F & 0.323\,{\scriptsize $\pm$0.001} & 0.326\,{\scriptsize $\pm$0.003} & \textbf{0.349}\,{\scriptsize $\pm$0.002} & 0.302\,{\scriptsize $\pm$0.001} & 0.308\,{\scriptsize $\pm$0.003} \\
\midrule
\multirow{3}{*}{Qwen3-VL-4B} & R1-F & 0.320\,{\scriptsize $\pm$0.002} & 0.323\,{\scriptsize $\pm$0.008} & \textbf{0.367}\,{\scriptsize $\pm$0.004} & 0.361\,{\scriptsize $\pm$0.002} & 0.306\,{\scriptsize $\pm$0.001} \\
 & R2-F & 0.194\,{\scriptsize $\pm$0.004} & 0.205\,{\scriptsize $\pm$0.008} & \textbf{0.260}\,{\scriptsize $\pm$0.003} & 0.241\,{\scriptsize $\pm$0.002} & 0.146\,{\scriptsize $\pm$0.001} \\
 & RL-F & 0.313\,{\scriptsize $\pm$0.002} & 0.320\,{\scriptsize $\pm$0.007} & \textbf{0.358}\,{\scriptsize $\pm$0.004} & 0.355\,{\scriptsize $\pm$0.002} & 0.303\,{\scriptsize $\pm$0.001} \\
\midrule
\multirow{3}{*}{Gemma3-4B} & R1-F & 0.316\,{\scriptsize $\pm$0.012} & 0.320\,{\scriptsize $\pm$0.007} & 0.350\,{\scriptsize $\pm$0.009} & \textbf{0.358}\,{\scriptsize $\pm$0.004} & 0.345\,{\scriptsize $\pm$0.007} \\
\bottomrule
\end{tabular}
\end{table}

\begin{table}[H]
\caption{\textbf{GRPO post-training results on MIMIC-CXR (Impression).} Embedding-based / lexical (METEOR, BERTScore F$_1$, COMET, ChrF$++$). Mean over 5 random seeds with sample std in \scriptsize. The final \emph{Mean} block averages each (reward, metric) cell across the three models. \textbf{Bold} = column-best reward function per row.}
\label{tab:grpo-app-mimiccxr-impression-embedding}
\centering
\footnotesize
\setlength{\tabcolsep}{3pt}
\renewcommand{\arraystretch}{0.95}
\begin{tabular}{llccccc}
\toprule
Model & Metric & SFT & $R_{\mathrm{fmt}}$ & $R_{\mathrm{exact}}$ & $R_{\mathrm{Cham}}$ & $R_{\mathrm{Haus}}$ \\
\midrule
\multirow{4}{*}{Qwen3-VL-2B} & METEOR & 0.316\,{\scriptsize $\pm$0.001} & 0.317\,{\scriptsize $\pm$0.003} & \textbf{0.352}\,{\scriptsize $\pm$0.001} & 0.348\,{\scriptsize $\pm$0.001} & 0.299\,{\scriptsize $\pm$0.003} \\
 & BERTScore F1 & 0.337\,{\scriptsize $\pm$0.004} & 0.339\,{\scriptsize $\pm$0.003} & \textbf{0.353}\,{\scriptsize $\pm$0.002} & 0.271\,{\scriptsize $\pm$0.001} & 0.342\,{\scriptsize $\pm$0.004} \\
 & COMET & 0.587\,{\scriptsize $\pm$0.002} & 0.588\,{\scriptsize $\pm$0.002} & 0.590\,{\scriptsize $\pm$0.001} & 0.593\,{\scriptsize $\pm$0.001} & \textbf{0.629}\,{\scriptsize $\pm$0.002} \\
 & ChrF++ & 19.433\,{\scriptsize $\pm$0.208} & 18.913\,{\scriptsize $\pm$0.315} & 23.500\,{\scriptsize $\pm$0.127} & \textbf{25.311}\,{\scriptsize $\pm$0.112} & 19.060\,{\scriptsize $\pm$0.225} \\
\midrule
\multirow{4}{*}{Qwen3-VL-4B} & METEOR & 0.306\,{\scriptsize $\pm$0.001} & 0.309\,{\scriptsize $\pm$0.005} & \textbf{0.357}\,{\scriptsize $\pm$0.003} & 0.346\,{\scriptsize $\pm$0.002} & 0.290\,{\scriptsize $\pm$0.001} \\
 & BERTScore F1 & 0.326\,{\scriptsize $\pm$0.003} & 0.329\,{\scriptsize $\pm$0.009} & \textbf{0.378}\,{\scriptsize $\pm$0.003} & 0.374\,{\scriptsize $\pm$0.002} & 0.298\,{\scriptsize $\pm$0.001} \\
 & COMET & 0.582\,{\scriptsize $\pm$0.001} & 0.584\,{\scriptsize $\pm$0.003} & 0.597\,{\scriptsize $\pm$0.002} & \textbf{0.605}\,{\scriptsize $\pm$0.001} & 0.564\,{\scriptsize $\pm$0.001} \\
\bottomrule
\end{tabular}
\end{table}

\begin{table}[H]
\caption{\textbf{GRPO post-training results on MIMIC-CXR (Impression).} Clinical (RadGraph, CheXbert). Mean over 5 random seeds with sample std in \scriptsize. The final \emph{Mean} block averages each (reward, metric) cell across the three models. \textbf{Bold} = column-best reward function per row.}
\label{tab:grpo-app-mimiccxr-impression-clinical}
\centering
\footnotesize
\setlength{\tabcolsep}{3pt}
\renewcommand{\arraystretch}{0.95}
\begin{tabular}{llccccc}
\toprule
Model & Metric & SFT & $R_{\mathrm{fmt}}$ & $R_{\mathrm{exact}}$ & $R_{\mathrm{Cham}}$ & $R_{\mathrm{Haus}}$ \\
\midrule
\multirow{4}{*}{Qwen3-VL-2B} & RG-E & 0.288\,{\scriptsize $\pm$0.005} & 0.290\,{\scriptsize $\pm$0.004} & \textbf{0.323}\,{\scriptsize $\pm$0.004} & 0.304\,{\scriptsize $\pm$0.001} & 0.262\,{\scriptsize $\pm$0.004} \\
 & RG-ER & 0.280\,{\scriptsize $\pm$0.005} & 0.282\,{\scriptsize $\pm$0.005} & \textbf{0.315}\,{\scriptsize $\pm$0.004} & 0.297\,{\scriptsize $\pm$0.001} & 0.254\,{\scriptsize $\pm$0.004} \\
 & RadGraph F1 & 0.219\,{\scriptsize $\pm$0.004} & 0.217\,{\scriptsize $\pm$0.006} & \textbf{0.264}\,{\scriptsize $\pm$0.002} & 0.250\,{\scriptsize $\pm$0.001} & 0.147\,{\scriptsize $\pm$0.004} \\
 & CheXbert F1 & 0.616\,{\scriptsize $\pm$0.006} & 0.624\,{\scriptsize $\pm$0.010} & 0.562\,{\scriptsize $\pm$0.008} & 0.573\,{\scriptsize $\pm$0.004} & \textbf{0.640}\,{\scriptsize $\pm$0.006} \\
\midrule
\multirow{4}{*}{Qwen3-VL-4B} & RG-E & 0.272\,{\scriptsize $\pm$0.003} & 0.278\,{\scriptsize $\pm$0.007} & \textbf{0.338}\,{\scriptsize $\pm$0.003} & 0.333\,{\scriptsize $\pm$0.002} & 0.258\,{\scriptsize $\pm$0.001} \\
 & RG-ER & 0.262\,{\scriptsize $\pm$0.004} & 0.271\,{\scriptsize $\pm$0.007} & \textbf{0.328}\,{\scriptsize $\pm$0.003} & 0.321\,{\scriptsize $\pm$0.002} & 0.253\,{\scriptsize $\pm$0.001} \\
 & RadGraph F1 & 0.199\,{\scriptsize $\pm$0.004} & 0.208\,{\scriptsize $\pm$0.010} & \textbf{0.276}\,{\scriptsize $\pm$0.004} & 0.254\,{\scriptsize $\pm$0.003} & 0.142\,{\scriptsize $\pm$0.000} \\
\bottomrule
\end{tabular}
\end{table}





%

\section{Full selection-results breakdown}
\label{sec:appendix-results}

This appendix reports, for every model in our experiments and for every NLP
metric we computed, the absolute score of every (distance metric, aggregation)
selection policy alongside the random-selection baseline (13 runs total).
We first reproduce the headline-summary table for the \textsc{Impression}
section that complements Tab.~\ref{tab:headline-results} of the main paper
(Findings); the cross-model overall ranking and per-model breakdowns follow.

\begin{table}[H]
\caption{\textbf{Headline results (Impressions).} For every model and every of five clinically meaningful NLP metrics we report the best score obtained by any (distance metric, aggregation) selection policy. The matched random-selection baseline is shown in italics under each model row, and the percentage improvement over random is given in parentheses. Bold marks the best value per column.}
\label{tab:headline-results-impressions}
\centering
\resizebox{\linewidth}{!}{%
\begin{tabular}{lccccc}
\toprule
Model & RL-F & METEOR & BS-F1 & RG-F1 & CXB-14 \\
\midrule
Qwen3-VL-4B GRPO ($R_{\mathrm{exact}}$) & 0.325\textsubscript{(+7.9\%)} & 0.326\textsubscript{(+4.2\%)} & 0.354\textsubscript{(+9.0\%)} & 0.248\textsubscript{(+11.1\%)} & 0.770\textsubscript{(+11.3\%)} \\
\quad\textit{random} & \textit{0.301} & \textit{0.313} & \textit{0.325} & \textit{0.223} & \textit{0.693} \\[1pt]
Qwen3-VL-4B GRPO ($R_{\mathrm{fmt}}$) & 0.334\textsubscript{(+13.0\%)} & 0.324\textsubscript{(+10.6\%)} & 0.349\textsubscript{(+8.3\%)} & 0.248\textsubscript{(+18.2\%)} & 0.776\textsubscript{(+4.5\%)} \\
\quad\textit{random} & \textit{0.296} & \textit{0.293} & \textit{0.323} & \textit{0.210} & \textit{0.743} \\[1pt]
Qwen3-VL-4B GRPO ($R_{\mathrm{Cham}}$) & \textbf{0.341}\textsubscript{(+0.3\%)} & \textbf{0.329}\textsubscript{(+0.3\%)} & 0.344\textsubscript{(+0.1\%)} & \textbf{0.257}\textsubscript{(+0.0\%)} & 0.776\textsubscript{(-0.2\%)} \\
\quad\textit{random} & \textit{0.339} & \textit{0.328} & \textit{0.343} & \textit{0.257} & \textit{0.778} \\[1pt]
Qwen3-VL-4B GRPO ($R_{\mathrm{Haus}}$) & 0.330\textsubscript{(+6.4\%)} & 0.319\textsubscript{(+6.1\%)} & \textbf{0.359}\textsubscript{(+4.1\%)} & 0.225\textsubscript{(+30.8\%)} & 0.776\textsubscript{(+0.5\%)} \\
\quad\textit{random} & \textit{0.310} & \textit{0.300} & \textit{0.345} & \textit{0.172} & \textit{0.772} \\[1pt]
Mistral-Small [p1] & 0.052\textsubscript{(+34.0\%)} & 0.113\textsubscript{(+16.3\%)} & 0.167\textsubscript{(+11.2\%)} & 0.026\textsubscript{(+21.2\%)} & 0.682\textsubscript{(+18.0\%)} \\
\quad\textit{random} & \textit{0.039} & \textit{0.098} & \textit{0.151} & \textit{0.021} & \textit{0.578} \\[1pt]
Mistral-Small [p2] & 0.299\textsubscript{(+34.5\%)} & 0.280\textsubscript{(+26.4\%)} & 0.322\textsubscript{(+15.0\%)} & 0.175\textsubscript{(+48.9\%)} & 0.779\textsubscript{(+2.0\%)} \\
\quad\textit{random} & \textit{0.222} & \textit{0.221} & \textit{0.280} & \textit{0.118} & \textit{0.764} \\[1pt]
Gemini 2.5 Flash-Lite [p1] & 0.157\textsubscript{(+78.1\%)} & 0.172\textsubscript{(+31.2\%)} & 0.228\textsubscript{(+36.2\%)} & 0.077\textsubscript{(+83.6\%)} & 0.696\textsubscript{(+5.7\%)} \\
\quad\textit{random} & \textit{0.088} & \textit{0.131} & \textit{0.168} & \textit{0.042} & \textit{0.658} \\[1pt]
Gemini 2.5 Flash-Lite [p2] & 0.271\textsubscript{(+45.4\%)} & 0.259\textsubscript{(+29.7\%)} & 0.299\textsubscript{(+28.0\%)} & 0.140\textsubscript{(+52.4\%)} & 0.734\textsubscript{(+8.0\%)} \\
\quad\textit{random} & \textit{0.187} & \textit{0.200} & \textit{0.234} & \textit{0.092} & \textit{0.680} \\[1pt]
Gemini 3.1 Flash-Lite [p1] & 0.195\textsubscript{(+60.3\%)} & 0.204\textsubscript{(+28.0\%)} & 0.259\textsubscript{(+24.4\%)} & 0.084\textsubscript{(+49.0\%)} & 0.747\textsubscript{(+3.0\%)} \\
\quad\textit{random} & \textit{0.121} & \textit{0.159} & \textit{0.208} & \textit{0.056} & \textit{0.725} \\[1pt]
Gemini 3.1 Flash-Lite [p2] & 0.288\textsubscript{(+18.3\%)} & 0.277\textsubscript{(+13.6\%)} & 0.325\textsubscript{(+12.1\%)} & 0.154\textsubscript{(+23.4\%)} & 0.738\textsubscript{(+2.5\%)} \\
\quad\textit{random} & \textit{0.243} & \textit{0.244} & \textit{0.290} & \textit{0.125} & \textit{0.720} \\[1pt]
GPT-4o mini [p1] & 0.099\textsubscript{(+82.3\%)} & 0.142\textsubscript{(+34.1\%)} & 0.175\textsubscript{(+17.6\%)} & 0.026\textsubscript{(+40.5\%)} & 0.771\textsubscript{(+2.9\%)} \\
\quad\textit{random} & \textit{0.054} & \textit{0.106} & \textit{0.149} & \textit{0.018} & \textit{0.750} \\[1pt]
GPT-4o mini [p2] & 0.293\textsubscript{(+33.5\%)} & 0.285\textsubscript{(+24.0\%)} & 0.305\textsubscript{(+17.1\%)} & 0.177\textsubscript{(+57.4\%)} & 0.777\textsubscript{(+3.6\%)} \\
\quad\textit{random} & \textit{0.220} & \textit{0.230} & \textit{0.260} & \textit{0.113} & \textit{0.750} \\[1pt]
GPT-5 mini [p2] & 0.316\textsubscript{(+24.5\%)} & 0.311\textsubscript{(+14.7\%)} & 0.306\textsubscript{(+20.1\%)} & 0.193\textsubscript{(+24.0\%)} & \textbf{0.780}\textsubscript{(+1.4\%)} \\
\quad\textit{random} & \textit{0.254} & \textit{0.271} & \textit{0.255} & \textit{0.156} & \textit{0.769} \\[1pt]
\bottomrule
\end{tabular}%
}
\end{table}

\subsection{Visualisation of the headline data}

\begin{figure}[p]
    \centering
    \includegraphics[width=0.86\linewidth]{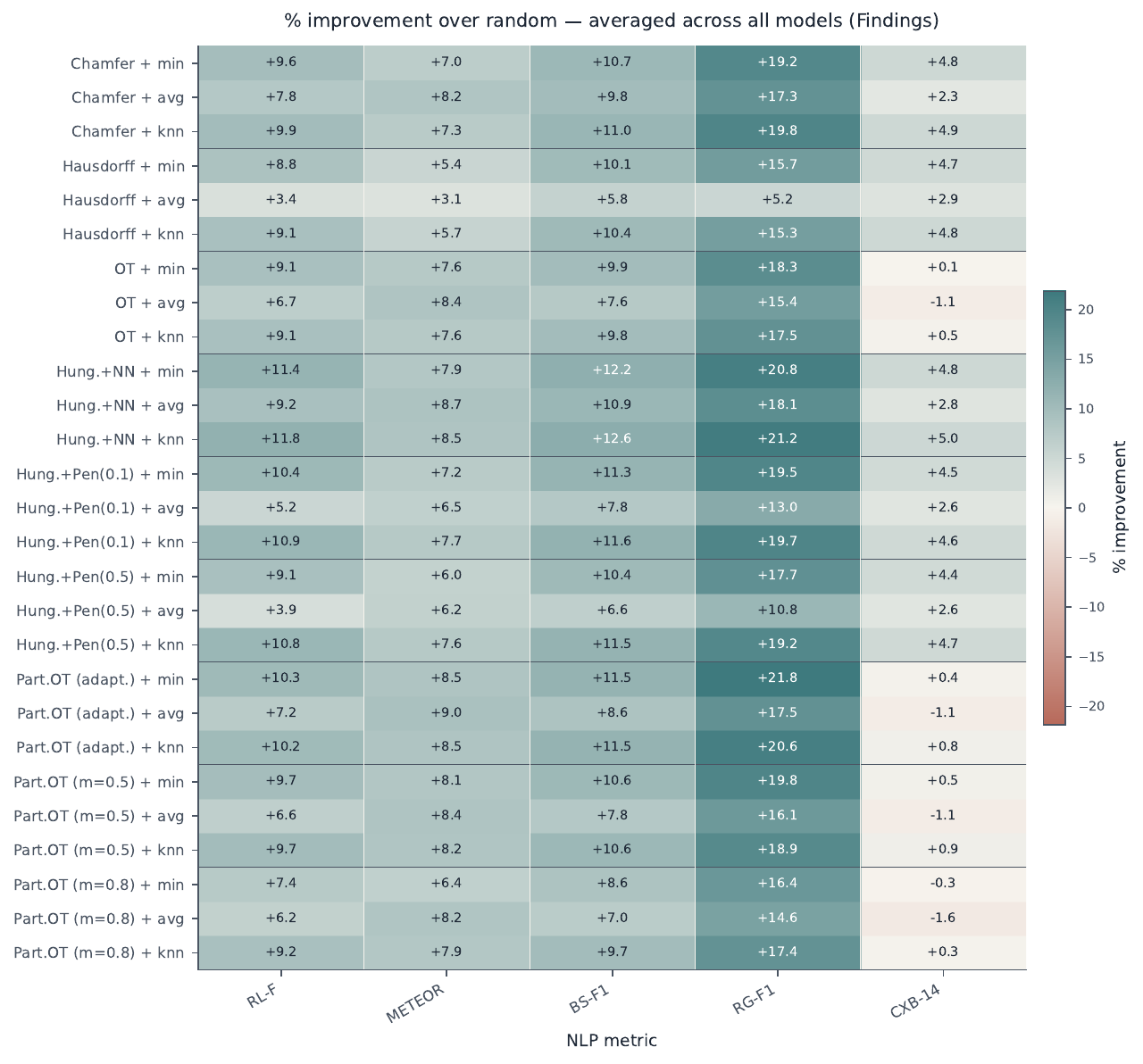}\\[4pt]
    \includegraphics[width=0.86\linewidth]{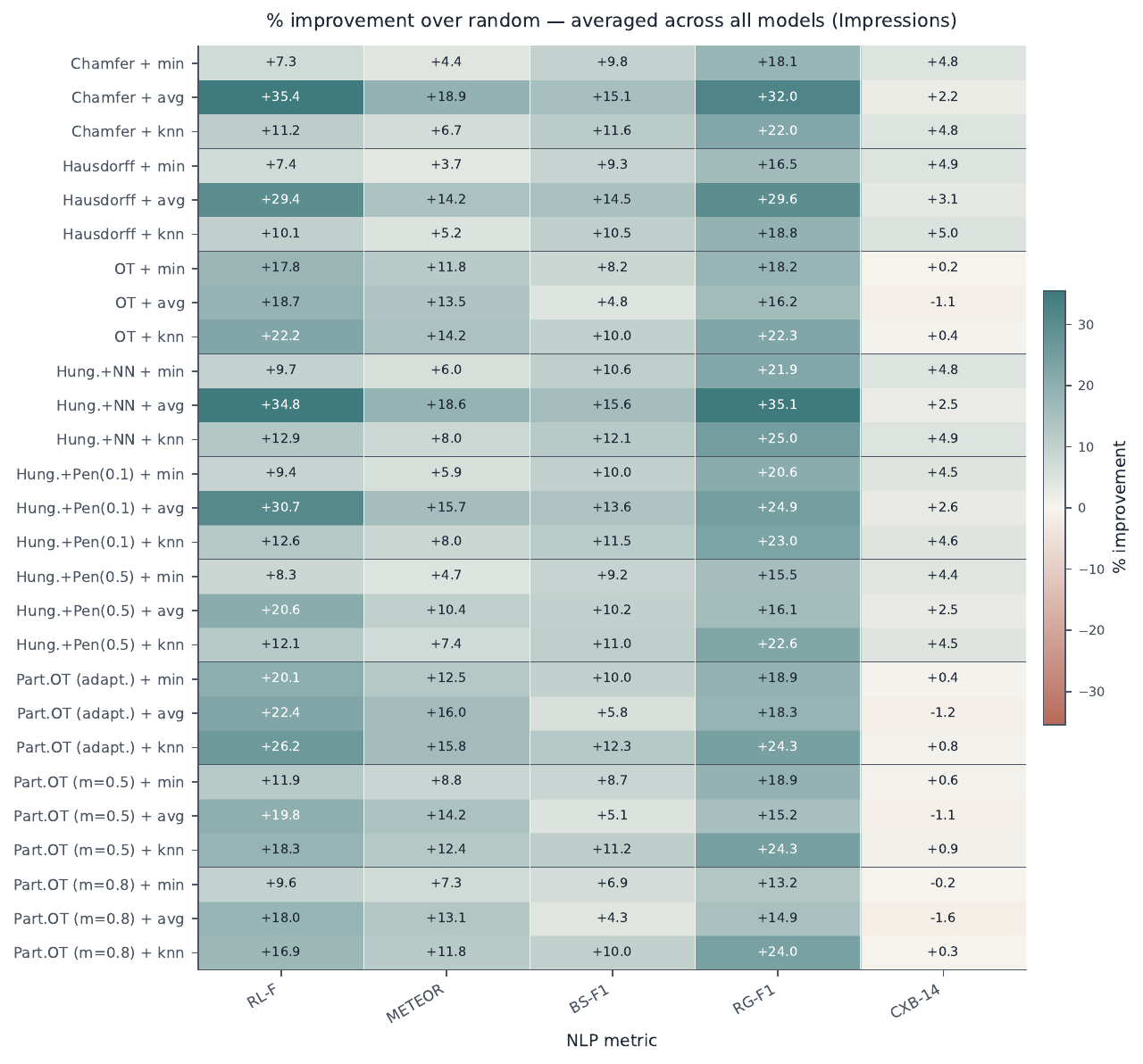}
    \caption{\textbf{Method$\,\times\,$metric heatmap.} Mean percentage
    improvement over random selection averaged across all 13 models,
    on five clinically meaningful metrics. Rows are (distance metric,
    aggregation) pairs grouped by distance-metric family; columns are NLP
    metrics. Teal cells beat random, coral cells lose to it.}
    \label{fig:results-heatmap}
\end{figure}

Fig.~\ref{fig:results-heatmap} renders the same data as
Tab.~\ref{tab:headline-results} (and its Impression counterpart
Tab.~\ref{tab:headline-results-impressions}) as a
method$\,\times\,$metric heatmap.

\clearpage
\clearpage

\section{Stratified clinical-meaningfulness analysis}
\label{sec:clinical-appendix}

This appendix reports, for every model, every (distance metric, aggregation)
selection policy, every ground-truth-defined stratum and every per-sample
metric, both the absolute score and the difference vs.\ the random baseline
of the same stratum. Joined-report (CheXbert macro F1 and per-pathology F1)
tables follow at the end. Random / oracle / naive baselines are banded for
reference (random is computed on the same per-sample metrics used here, which
are pooled across the joined \textsc{Findings}~$\cup$~\textsc{Impression}
report).

\subsection{Stratified tables (four metrics)}
\label{sec:clinical-stratified-tables}

The five tables in this subsection report the per-model best-policy view and the matched percent-improvement-over-random tables across all patients and within each ground-truth-defined stratum, on the four main metrics (BS-F1, ROUGE-L F1, METEOR and RG-F1). Per-model and per-pathology breakdowns follow.
\begin{table}[H]
\caption{\textbf{Stratified results.} Each row pair shows, for one model, the score of its best selection policy and the matched random baseline (italic), broken down by ground-truth abnormality stratum (no-finding / single finding / multiple findings; $n_A{=}504$, $n_B{=}206$, $n_C{=}290$). The best policy is the (distance, aggregation) combination that maximises mean improvement over random across all (stratum$\times$metric) cells. Subscripts: \% improvement over random in that stratum. RG-F1 is per-sample RadGraph entity F1; the corpus-level relation-aware variant reported in Tab.~\ref{tab:headline-results} is not defined per-sample.\strut}
\label{tab:clinical-headline-4m}
\centering
\resizebox{\linewidth}{!}{%
\begin{tabular}{llcccccccccccc}
\toprule
 &  & \multicolumn{4}{c}{No-Finding} & \multicolumn{4}{c}{Single} & \multicolumn{4}{c}{Multi} \\
\cmidrule(lr){3-6} \cmidrule(lr){7-10} \cmidrule(lr){11-14}
Model & Policy & BS-F1 & RL-F & METEOR & RG-F1 & BS-F1 & RL-F & METEOR & RG-F1 & BS-F1 & RL-F & METEOR & RG-F1 \\
\midrule
Qwen3-VL-4B GRPO ($R_{\mathrm{exact}}$) & Hung.\,+Pen($\alpha$=0.1)+avg & 0.340\textsubscript{(+6.2\%)} & 0.420\textsubscript{(+4.4\%)} & 0.456\textsubscript{(+3.6\%)} & 0.497\textsubscript{(+5.1\%)} & 0.143\textsubscript{(+12.9\%)} & 0.211\textsubscript{(+5.2\%)} & 0.226\textsubscript{(+2.2\%)} & 0.264\textsubscript{(+6.0\%)} & 0.062\textsubscript{(+18.5\%)} & 0.156\textsubscript{(+3.9\%)} & 0.158\textsubscript{(-2.1\%)} & 0.157\textsubscript{(+1.3\%)} \\
\textit{random} & -- & \textit{0.320} & \textit{0.402} & \textit{0.440} & \textit{0.473} & \textit{0.127} & \textit{0.200} & \textit{0.222} & \textit{0.249} & \textit{0.052} & \textit{0.151} & \textit{0.161} & \textit{0.155} \\[1pt]
Qwen3-VL-4B GRPO ($R_{\mathrm{fmt}}$) & Part.\,OT ($\rho$=0.5)+avg & 0.349\textsubscript{(+9.3\%)} & 0.427\textsubscript{(+10.7\%)} & 0.463\textsubscript{(+8.5\%)} & 0.512\textsubscript{(+10.2\%)} & 0.128\textsubscript{(+0.1\%)} & 0.194\textsubscript{(-0.6\%)} & 0.209\textsubscript{(+0.4\%)} & 0.245\textsubscript{(+2.8\%)} & 0.059\textsubscript{(+8.6\%)} & 0.154\textsubscript{(+4.7\%)} & 0.164\textsubscript{(+13.7\%)} & 0.162\textsubscript{(+9.1\%)} \\
\textit{random} & -- & \textit{0.320} & \textit{0.386} & \textit{0.427} & \textit{0.465} & \textit{0.128} & \textit{0.195} & \textit{0.208} & \textit{0.238} & \textit{0.055} & \textit{0.147} & \textit{0.145} & \textit{0.148} \\[1pt]
Qwen3-VL-4B GRPO ($R_{\mathrm{Cham}}$) & Hausdorff+knn & 0.358\textsubscript{(+1.2\%)} & 0.435\textsubscript{(+0.4\%)} & 0.488\textsubscript{(+0.4\%)} & 0.532\textsubscript{(+0.5\%)} & 0.183\textsubscript{(+1.5\%)} & 0.254\textsubscript{(+0.4\%)} & 0.270\textsubscript{(+0.4\%)} & 0.336\textsubscript{(+0.3\%)} & 0.083\textsubscript{(+17.5\%)} & 0.167\textsubscript{(+1.2\%)} & 0.168\textsubscript{(+1.0\%)} & 0.202\textsubscript{(+1.2\%)} \\
\textit{random} & -- & \textit{0.354} & \textit{0.433} & \textit{0.487} & \textit{0.529} & \textit{0.180} & \textit{0.253} & \textit{0.269} & \textit{0.335} & \textit{0.071} & \textit{0.165} & \textit{0.167} & \textit{0.199} \\[1pt]
Qwen3-VL-4B GRPO ($R_{\mathrm{Haus}}$) & Hung.\,+NN+min & 0.335\textsubscript{(+13.8\%)} & 0.422\textsubscript{(+14.0\%)} & 0.449\textsubscript{(+11.5\%)} & 0.513\textsubscript{(+6.6\%)} & 0.124\textsubscript{(+1.7\%)} & 0.210\textsubscript{(-4.1\%)} & 0.204\textsubscript{(-1.7\%)} & 0.263\textsubscript{(-0.7\%)} & 0.040\textsubscript{(+1.1\%)} & 0.146\textsubscript{(-1.8\%)} & 0.131\textsubscript{(-1.7\%)} & 0.150\textsubscript{(-1.1\%)} \\
\textit{random} & -- & \textit{0.295} & \textit{0.370} & \textit{0.403} & \textit{0.482} & \textit{0.122} & \textit{0.219} & \textit{0.207} & \textit{0.265} & \textit{0.040} & \textit{0.149} & \textit{0.133} & \textit{0.152} \\[1pt]
Mistral-Small [p1] & Chamfer+avg & 0.215\textsubscript{(+17.0\%)} & 0.182\textsubscript{(+10.3\%)} & 0.276\textsubscript{(+13.2\%)} & 0.226\textsubscript{(+13.7\%)} & 0.157\textsubscript{(+26.2\%)} & 0.162\textsubscript{(+10.6\%)} & 0.221\textsubscript{(+15.2\%)} & 0.210\textsubscript{(+24.2\%)} & 0.094\textsubscript{(+22.9\%)} & 0.140\textsubscript{(+7.3\%)} & 0.169\textsubscript{(+11.3\%)} & 0.152\textsubscript{(+7.1\%)} \\
\textit{random} & -- & \textit{0.184} & \textit{0.165} & \textit{0.244} & \textit{0.199} & \textit{0.124} & \textit{0.146} & \textit{0.192} & \textit{0.169} & \textit{0.076} & \textit{0.131} & \textit{0.152} & \textit{0.142} \\[1pt]
Mistral-Small [p2] & Hung.\,+NN+avg & 0.365\textsubscript{(+13.9\%)} & 0.357\textsubscript{(+18.7\%)} & 0.437\textsubscript{(+17.9\%)} & 0.440\textsubscript{(+18.1\%)} & 0.212\textsubscript{(+11.5\%)} & 0.223\textsubscript{(+13.3\%)} & 0.257\textsubscript{(+12.3\%)} & 0.290\textsubscript{(+14.4\%)} & 0.111\textsubscript{(+5.7\%)} & 0.156\textsubscript{(+3.7\%)} & 0.161\textsubscript{(+0.8\%)} & 0.163\textsubscript{(+1.4\%)} \\
\textit{random} & -- & \textit{0.321} & \textit{0.300} & \textit{0.371} & \textit{0.373} & \textit{0.190} & \textit{0.197} & \textit{0.228} & \textit{0.253} & \textit{0.105} & \textit{0.150} & \textit{0.160} & \textit{0.161} \\[1pt]
Gemini 2.5 Flash-Lite [p1] & Hung.\,+Pen($\alpha$=0.5)+avg & 0.240\textsubscript{(+19.5\%)} & 0.248\textsubscript{(+12.7\%)} & 0.328\textsubscript{(+14.5\%)} & 0.312\textsubscript{(+19.1\%)} & 0.143\textsubscript{(+22.8\%)} & 0.180\textsubscript{(+15.9\%)} & 0.224\textsubscript{(+16.5\%)} & 0.219\textsubscript{(+27.5\%)} & 0.097\textsubscript{(+13.5\%)} & 0.159\textsubscript{(+4.3\%)} & 0.183\textsubscript{(+5.2\%)} & 0.181\textsubscript{(+10.1\%)} \\
\textit{random} & -- & \textit{0.201} & \textit{0.220} & \textit{0.286} & \textit{0.262} & \textit{0.116} & \textit{0.156} & \textit{0.192} & \textit{0.172} & \textit{0.086} & \textit{0.152} & \textit{0.174} & \textit{0.164} \\[1pt]
Gemini 2.5 Flash-Lite [p2] & Hung.\,+Pen($\alpha$=0.1)+avg & 0.312\textsubscript{(+23.4\%)} & 0.317\textsubscript{(+21.3\%)} & 0.376\textsubscript{(+17.1\%)} & 0.376\textsubscript{(+22.3\%)} & 0.186\textsubscript{(+30.3\%)} & 0.219\textsubscript{(+19.1\%)} & 0.251\textsubscript{(+17.6\%)} & 0.265\textsubscript{(+29.0\%)} & 0.120\textsubscript{(+17.3\%)} & 0.176\textsubscript{(+9.4\%)} & 0.191\textsubscript{(+6.1\%)} & 0.195\textsubscript{(+11.7\%)} \\
\textit{random} & -- & \textit{0.253} & \textit{0.261} & \textit{0.321} & \textit{0.307} & \textit{0.143} & \textit{0.184} & \textit{0.214} & \textit{0.205} & \textit{0.103} & \textit{0.161} & \textit{0.180} & \textit{0.174} \\[1pt]
Gemini 3.1 Flash-Lite [p1] & Hung.\,+Pen($\alpha$=0.1)+avg & 0.285\textsubscript{(+12.2\%)} & 0.272\textsubscript{(+11.2\%)} & 0.380\textsubscript{(+10.1\%)} & 0.333\textsubscript{(+9.8\%)} & 0.165\textsubscript{(+6.3\%)} & 0.198\textsubscript{(+3.7\%)} & 0.276\textsubscript{(+6.5\%)} & 0.250\textsubscript{(+4.5\%)} & 0.107\textsubscript{(+4.6\%)} & 0.173\textsubscript{(+0.0\%)} & 0.225\textsubscript{(+1.4\%)} & 0.198\textsubscript{(+0.4\%)} \\
\textit{random} & -- & \textit{0.254} & \textit{0.244} & \textit{0.345} & \textit{0.303} & \textit{0.155} & \textit{0.190} & \textit{0.259} & \textit{0.239} & \textit{0.103} & \textit{0.173} & \textit{0.222} & \textit{0.197} \\[1pt]
Gemini 3.1 Flash-Lite [p2] & Chamfer+avg & 0.352\textsubscript{(+9.7\%)} & 0.355\textsubscript{(+9.8\%)} & 0.455\textsubscript{(+7.6\%)} & 0.432\textsubscript{(+9.6\%)} & 0.206\textsubscript{(+6.7\%)} & 0.227\textsubscript{(+5.5\%)} & 0.291\textsubscript{(+4.3\%)} & 0.298\textsubscript{(+3.2\%)} & 0.126\textsubscript{(+1.6\%)} & 0.186\textsubscript{(+3.4\%)} & 0.224\textsubscript{(-0.4\%)} & 0.219\textsubscript{(-0.7\%)} \\
\textit{random} & -- & \textit{0.321} & \textit{0.323} & \textit{0.423} & \textit{0.394} & \textit{0.193} & \textit{0.216} & \textit{0.279} & \textit{0.289} & \textit{0.124} & \textit{0.180} & \textit{0.225} & \textit{0.220} \\[1pt]
GPT-4o mini [p1] & Part.\,OT ($\rho$=0.8)+min & 0.241\textsubscript{(+28.7\%)} & 0.213\textsubscript{(+15.6\%)} & 0.304\textsubscript{(+16.5\%)} & 0.313\textsubscript{(+28.0\%)} & 0.141\textsubscript{(+34.9\%)} & 0.163\textsubscript{(+8.2\%)} & 0.210\textsubscript{(+8.9\%)} & 0.225\textsubscript{(+24.5\%)} & 0.069\textsubscript{(+44.1\%)} & 0.128\textsubscript{(+7.8\%)} & 0.154\textsubscript{(+9.2\%)} & 0.160\textsubscript{(+17.6\%)} \\
\textit{random} & -- & \textit{0.188} & \textit{0.184} & \textit{0.261} & \textit{0.245} & \textit{0.105} & \textit{0.150} & \textit{0.193} & \textit{0.181} & \textit{0.048} & \textit{0.119} & \textit{0.141} & \textit{0.136} \\[1pt]
GPT-4o mini [p2] & Hung.\,+Pen($\alpha$=0.1)+knn & 0.343\textsubscript{(+17.4\%)} & 0.352\textsubscript{(+21.1\%)} & 0.427\textsubscript{(+16.3\%)} & 0.440\textsubscript{(+21.9\%)} & 0.206\textsubscript{(+8.9\%)} & 0.236\textsubscript{(+9.5\%)} & 0.266\textsubscript{(+4.9\%)} & 0.299\textsubscript{(+14.6\%)} & 0.107\textsubscript{(+5.2\%)} & 0.167\textsubscript{(+3.5\%)} & 0.171\textsubscript{(+0.2\%)} & 0.173\textsubscript{(+0.4\%)} \\
\textit{random} & -- & \textit{0.292} & \textit{0.291} & \textit{0.367} & \textit{0.361} & \textit{0.189} & \textit{0.216} & \textit{0.253} & \textit{0.261} & \textit{0.102} & \textit{0.161} & \textit{0.170} & \textit{0.173} \\[1pt]
GPT-5 mini [p2] & Hung.\,+Pen($\alpha$=0.1)+avg & 0.279\textsubscript{(+8.2\%)} & 0.290\textsubscript{(+7.4\%)} & 0.349\textsubscript{(+7.0\%)} & 0.372\textsubscript{(+7.0\%)} & 0.168\textsubscript{(+6.4\%)} & 0.203\textsubscript{(+2.9\%)} & 0.234\textsubscript{(+5.8\%)} & 0.275\textsubscript{(+5.5\%)} & 0.082\textsubscript{(+3.8\%)} & 0.137\textsubscript{(+0.1\%)} & 0.150\textsubscript{(-0.3\%)} & 0.170\textsubscript{(-0.9\%)} \\
\textit{random} & -- & \textit{0.258} & \textit{0.270} & \textit{0.326} & \textit{0.348} & \textit{0.158} & \textit{0.197} & \textit{0.221} & \textit{0.261} & \textit{0.079} & \textit{0.137} & \textit{0.150} & \textit{0.171} \\[1pt]
\bottomrule
\end{tabular}%
}
\end{table}
\begin{table}[H]
\caption{\textbf{All patients} --- mean $\pm$ std percentage improvement over random selection per (distance metric, aggregation) policy. Bold: column-best mean.}
\label{tab:clinical-pct-all-4m}
\centering
\begin{tabular}{llcccc}
\toprule
Distance metric & Agg. & BS-F1 (\%) & RL-F (\%) & METEOR (\%) & RG-F1 (\%) \\
\midrule
Chamfer & Min & +10.7\,{\scriptsize $\pm$10.2} & +7.9\,{\scriptsize $\pm$7.3} & +4.7\,{\scriptsize $\pm$5.7} & +9.6\,{\scriptsize $\pm$8.1} \\
 & Avg & +11.7\,{\scriptsize $\pm$6.2} & +8.4\,{\scriptsize $\pm$4.5} & +8.1\,{\scriptsize $\pm$4.3} & +9.3\,{\scriptsize $\pm$5.6} \\
 & Knn & +12.8\,{\scriptsize $\pm$8.1} & +9.5\,{\scriptsize $\pm$5.2} & +6.2\,{\scriptsize $\pm$4.0} & +10.5\,{\scriptsize $\pm$7.2} \\
\midrule
Hausdorff & Min & +8.3\,{\scriptsize $\pm$10.0} & +6.1\,{\scriptsize $\pm$7.3} & +2.6\,{\scriptsize $\pm$5.7} & +7.6\,{\scriptsize $\pm$7.3} \\
 & Avg & +8.4\,{\scriptsize $\pm$5.6} & +5.2\,{\scriptsize $\pm$3.5} & +4.1\,{\scriptsize $\pm$3.3} & +5.2\,{\scriptsize $\pm$4.3} \\
 & Knn & +10.3\,{\scriptsize $\pm$7.9} & +7.7\,{\scriptsize $\pm$5.8} & +3.9\,{\scriptsize $\pm$4.9} & +8.2\,{\scriptsize $\pm$6.5} \\
\midrule
OT & Min & +9.0\,{\scriptsize $\pm$6.4} & +7.2\,{\scriptsize $\pm$5.4} & +6.5\,{\scriptsize $\pm$4.4} & +7.9\,{\scriptsize $\pm$5.9} \\
 & Avg & +7.9\,{\scriptsize $\pm$5.8} & +6.1\,{\scriptsize $\pm$4.9} & +8.5\,{\scriptsize $\pm$5.0} & +7.0\,{\scriptsize $\pm$5.5} \\
 & Knn & +9.8\,{\scriptsize $\pm$5.6} & +8.0\,{\scriptsize $\pm$4.9} & +7.0\,{\scriptsize $\pm$4.0} & +8.3\,{\scriptsize $\pm$5.5} \\
\midrule
Hung.\,+NN & Min & +12.2\,{\scriptsize $\pm$7.4} & +9.1\,{\scriptsize $\pm$5.0} & +5.5\,{\scriptsize $\pm$4.0} & +9.9\,{\scriptsize $\pm$6.9} \\
 & Avg & +11.7\,{\scriptsize $\pm$6.1} & +8.7\,{\scriptsize $\pm$4.5} & +7.9\,{\scriptsize $\pm$4.3} & +9.5\,{\scriptsize $\pm$5.5} \\
 & Knn & +12.7\,{\scriptsize $\pm$7.9} & +9.5\,{\scriptsize $\pm$5.3} & +6.1\,{\scriptsize $\pm$4.3} & +10.5\,{\scriptsize $\pm$7.2} \\
\midrule
Hung.\,+Pen($\alpha$=0.1) & Min & +12.1\,{\scriptsize $\pm$7.5} & +9.0\,{\scriptsize $\pm$4.9} & +5.6\,{\scriptsize $\pm$4.0} & +10.1\,{\scriptsize $\pm$7.0} \\
 & Avg & +9.8\,{\scriptsize $\pm$9.3} & +6.4\,{\scriptsize $\pm$7.0} & +6.7\,{\scriptsize $\pm$6.4} & +8.2\,{\scriptsize $\pm$7.0} \\
 & Knn & +12.6\,{\scriptsize $\pm$7.9} & \textbf{+9.6}\,{\scriptsize $\pm$5.2} & +6.0\,{\scriptsize $\pm$4.1} & \textbf{+10.6}\,{\scriptsize $\pm$7.2} \\
\midrule
Hung.\,+Pen($\alpha$=0.5) & Min & +10.5\,{\scriptsize $\pm$10.4} & +7.7\,{\scriptsize $\pm$7.5} & +4.2\,{\scriptsize $\pm$5.8} & +9.3\,{\scriptsize $\pm$8.4} \\
 & Avg & +9.4\,{\scriptsize $\pm$9.4} & +5.7\,{\scriptsize $\pm$6.5} & +6.5\,{\scriptsize $\pm$6.6} & +7.8\,{\scriptsize $\pm$7.5} \\
 & Knn & \textbf{+12.8}\,{\scriptsize $\pm$8.0} & +9.5\,{\scriptsize $\pm$5.5} & +5.9\,{\scriptsize $\pm$4.3} & +10.4\,{\scriptsize $\pm$7.4} \\
\midrule
Part.\,OT ($\rho$=0.5) & Min & +9.1\,{\scriptsize $\pm$9.5} & +6.8\,{\scriptsize $\pm$6.8} & +5.6\,{\scriptsize $\pm$5.8} & +8.6\,{\scriptsize $\pm$7.8} \\
 & Avg & +8.4\,{\scriptsize $\pm$5.4} & +6.7\,{\scriptsize $\pm$4.6} & \textbf{+8.7}\,{\scriptsize $\pm$5.0} & +7.5\,{\scriptsize $\pm$5.4} \\
 & Knn & +11.1\,{\scriptsize $\pm$7.1} & +8.5\,{\scriptsize $\pm$4.8} & +7.1\,{\scriptsize $\pm$3.8} & +9.3\,{\scriptsize $\pm$6.3} \\
\midrule
Part.\,OT ($\rho$=0.8) & Min & +9.1\,{\scriptsize $\pm$9.5} & +6.8\,{\scriptsize $\pm$6.8} & +5.6\,{\scriptsize $\pm$5.8} & +8.6\,{\scriptsize $\pm$7.8} \\
 & Avg & +8.4\,{\scriptsize $\pm$5.4} & +6.7\,{\scriptsize $\pm$4.6} & \textbf{+8.7}\,{\scriptsize $\pm$5.0} & +7.5\,{\scriptsize $\pm$5.4} \\
 & Knn & +11.1\,{\scriptsize $\pm$7.1} & +8.5\,{\scriptsize $\pm$4.8} & +7.1\,{\scriptsize $\pm$3.8} & +9.3\,{\scriptsize $\pm$6.3} \\
\bottomrule
\end{tabular}
\end{table}
\begin{table}[H]
\caption{\textbf{No-Finding stratum} --- mean $\pm$ std percentage improvement over random selection per (distance metric, aggregation) policy. Bold: column-best mean.}
\label{tab:clinical-pct-nofinding-4m}
\centering
\begin{tabular}{llcccc}
\toprule
Distance metric & Agg. & BS-F1 (\%) & RL-F (\%) & METEOR (\%) & RG-F1 (\%) \\
\midrule
Chamfer & Min & +11.5\,{\scriptsize $\pm$10.1} & +10.7\,{\scriptsize $\pm$9.9} & +7.1\,{\scriptsize $\pm$7.6} & +12.0\,{\scriptsize $\pm$9.3} \\
 & Avg & +12.0\,{\scriptsize $\pm$6.0} & +10.4\,{\scriptsize $\pm$5.2} & +9.8\,{\scriptsize $\pm$4.7} & +10.7\,{\scriptsize $\pm$5.9} \\
 & Knn & \textbf{+13.9}\,{\scriptsize $\pm$7.1} & +13.1\,{\scriptsize $\pm$7.0} & +9.2\,{\scriptsize $\pm$5.2} & +13.3\,{\scriptsize $\pm$7.8} \\
\midrule
Hausdorff & Min & +9.0\,{\scriptsize $\pm$10.2} & +8.4\,{\scriptsize $\pm$9.9} & +4.6\,{\scriptsize $\pm$7.7} & +9.5\,{\scriptsize $\pm$8.8} \\
 & Avg & +8.9\,{\scriptsize $\pm$5.7} & +6.8\,{\scriptsize $\pm$5.0} & +5.4\,{\scriptsize $\pm$4.0} & +6.3\,{\scriptsize $\pm$5.1} \\
 & Knn & +11.3\,{\scriptsize $\pm$7.6} & +10.6\,{\scriptsize $\pm$7.7} & +6.4\,{\scriptsize $\pm$6.3} & +10.3\,{\scriptsize $\pm$7.6} \\
\midrule
OT & Min & +9.4\,{\scriptsize $\pm$7.0} & +9.5\,{\scriptsize $\pm$7.8} & +8.3\,{\scriptsize $\pm$6.0} & +9.6\,{\scriptsize $\pm$7.3} \\
 & Avg & +7.8\,{\scriptsize $\pm$5.9} & +7.6\,{\scriptsize $\pm$6.4} & +9.2\,{\scriptsize $\pm$6.1} & +7.8\,{\scriptsize $\pm$6.1} \\
 & Knn & +10.3\,{\scriptsize $\pm$5.9} & +10.6\,{\scriptsize $\pm$7.1} & +9.1\,{\scriptsize $\pm$5.4} & +10.4\,{\scriptsize $\pm$6.9} \\
\midrule
Hung.\,+NN & Min & +13.2\,{\scriptsize $\pm$7.0} & +12.6\,{\scriptsize $\pm$6.7} & +8.4\,{\scriptsize $\pm$5.3} & +12.7\,{\scriptsize $\pm$7.6} \\
 & Avg & +12.6\,{\scriptsize $\pm$6.0} & +11.3\,{\scriptsize $\pm$5.4} & \textbf{+9.9}\,{\scriptsize $\pm$4.8} & +11.3\,{\scriptsize $\pm$5.9} \\
 & Knn & +13.8\,{\scriptsize $\pm$7.1} & +13.1\,{\scriptsize $\pm$7.1} & +9.1\,{\scriptsize $\pm$5.5} & +13.2\,{\scriptsize $\pm$7.8} \\
\midrule
Hung.\,+Pen($\alpha$=0.1) & Min & +13.3\,{\scriptsize $\pm$7.1} & +12.6\,{\scriptsize $\pm$6.6} & +8.5\,{\scriptsize $\pm$5.1} & +13.0\,{\scriptsize $\pm$7.8} \\
 & Avg & +9.5\,{\scriptsize $\pm$9.8} & +7.3\,{\scriptsize $\pm$9.1} & +7.5\,{\scriptsize $\pm$7.8} & +9.0\,{\scriptsize $\pm$8.1} \\
 & Knn & +13.8\,{\scriptsize $\pm$7.2} & \textbf{+13.2}\,{\scriptsize $\pm$7.0} & +9.1\,{\scriptsize $\pm$5.3} & \textbf{+13.4}\,{\scriptsize $\pm$7.9} \\
\midrule
Hung.\,+Pen($\alpha$=0.5) & Min & +11.4\,{\scriptsize $\pm$10.5} & +10.4\,{\scriptsize $\pm$10.4} & +6.5\,{\scriptsize $\pm$7.9} & +11.8\,{\scriptsize $\pm$9.7} \\
 & Avg & +8.9\,{\scriptsize $\pm$9.8} & +6.5\,{\scriptsize $\pm$8.6} & +7.2\,{\scriptsize $\pm$7.9} & +8.6\,{\scriptsize $\pm$8.3} \\
 & Knn & +13.8\,{\scriptsize $\pm$7.3} & +12.9\,{\scriptsize $\pm$7.4} & +8.9\,{\scriptsize $\pm$5.5} & +13.1\,{\scriptsize $\pm$8.0} \\
\midrule
Part.\,OT ($\rho$=0.5) & Min & +9.4\,{\scriptsize $\pm$9.7} & +9.0\,{\scriptsize $\pm$9.5} & +7.3\,{\scriptsize $\pm$7.6} & +10.5\,{\scriptsize $\pm$9.3} \\
 & Avg & +8.5\,{\scriptsize $\pm$5.2} & +8.4\,{\scriptsize $\pm$5.9} & +9.7\,{\scriptsize $\pm$5.7} & +8.5\,{\scriptsize $\pm$5.8} \\
 & Knn & +11.8\,{\scriptsize $\pm$6.7} & +11.5\,{\scriptsize $\pm$6.8} & +9.4\,{\scriptsize $\pm$5.0} & +11.6\,{\scriptsize $\pm$7.4} \\
\midrule
Part.\,OT ($\rho$=0.8) & Min & +9.4\,{\scriptsize $\pm$9.7} & +9.0\,{\scriptsize $\pm$9.5} & +7.3\,{\scriptsize $\pm$7.6} & +10.5\,{\scriptsize $\pm$9.3} \\
 & Avg & +8.5\,{\scriptsize $\pm$5.2} & +8.4\,{\scriptsize $\pm$5.9} & +9.7\,{\scriptsize $\pm$5.7} & +8.5\,{\scriptsize $\pm$5.8} \\
 & Knn & +11.8\,{\scriptsize $\pm$6.7} & +11.5\,{\scriptsize $\pm$6.8} & +9.4\,{\scriptsize $\pm$5.0} & +11.6\,{\scriptsize $\pm$7.3} \\
\bottomrule
\end{tabular}
\end{table}
\begin{table}[H]
\caption{\textbf{Single-finding stratum} --- mean $\pm$ std percentage improvement over random selection per (distance metric, aggregation) policy. Bold: column-best mean.}
\label{tab:clinical-pct-single-4m}
\centering
\begin{tabular}{llcccc}
\toprule
Distance metric & Agg. & BS-F1 (\%) & RL-F (\%) & METEOR (\%) & RG-F1 (\%) \\
\midrule
Chamfer & Min & +10.4\,{\scriptsize $\pm$11.5} & +5.6\,{\scriptsize $\pm$5.7} & +2.4\,{\scriptsize $\pm$4.5} & +9.3\,{\scriptsize $\pm$9.7} \\
 & Avg & +12.6\,{\scriptsize $\pm$9.9} & +6.8\,{\scriptsize $\pm$6.3} & +7.2\,{\scriptsize $\pm$6.1} & +10.8\,{\scriptsize $\pm$9.5} \\
 & Knn & +11.3\,{\scriptsize $\pm$12.3} & +5.4\,{\scriptsize $\pm$6.3} & +2.7\,{\scriptsize $\pm$5.0} & +9.5\,{\scriptsize $\pm$10.0} \\
\midrule
Hausdorff & Min & +7.9\,{\scriptsize $\pm$11.6} & +4.5\,{\scriptsize $\pm$5.1} & +0.8\,{\scriptsize $\pm$4.8} & +7.7\,{\scriptsize $\pm$8.0} \\
 & Avg & +8.8\,{\scriptsize $\pm$8.4} & +4.5\,{\scriptsize $\pm$4.4} & +3.6\,{\scriptsize $\pm$4.3} & +6.8\,{\scriptsize $\pm$6.7} \\
 & Knn & +9.1\,{\scriptsize $\pm$11.8} & +4.7\,{\scriptsize $\pm$5.8} & +1.2\,{\scriptsize $\pm$5.3} & +8.1\,{\scriptsize $\pm$8.8} \\
\midrule
OT & Min & +8.4\,{\scriptsize $\pm$8.7} & +4.6\,{\scriptsize $\pm$5.0} & +4.6\,{\scriptsize $\pm$4.5} & +7.0\,{\scriptsize $\pm$7.5} \\
 & Avg & +8.4\,{\scriptsize $\pm$9.4} & +4.6\,{\scriptsize $\pm$5.8} & +8.2\,{\scriptsize $\pm$6.0} & +7.8\,{\scriptsize $\pm$8.4} \\
 & Knn & +8.7\,{\scriptsize $\pm$8.7} & +4.8\,{\scriptsize $\pm$4.9} & +4.6\,{\scriptsize $\pm$4.8} & +7.0\,{\scriptsize $\pm$7.2} \\
\midrule
Hung.\,+NN & Min & +10.7\,{\scriptsize $\pm$10.6} & +4.9\,{\scriptsize $\pm$5.7} & +2.0\,{\scriptsize $\pm$4.2} & +8.7\,{\scriptsize $\pm$9.1} \\
 & Avg & +11.1\,{\scriptsize $\pm$9.6} & +6.1\,{\scriptsize $\pm$6.1} & +5.9\,{\scriptsize $\pm$6.3} & +9.9\,{\scriptsize $\pm$8.9} \\
 & Knn & +11.1\,{\scriptsize $\pm$11.9} & +5.5\,{\scriptsize $\pm$6.2} & +2.3\,{\scriptsize $\pm$5.0} & +9.3\,{\scriptsize $\pm$9.9} \\
\midrule
Hung.\,+Pen($\alpha$=0.1) & Min & +10.4\,{\scriptsize $\pm$10.3} & +4.7\,{\scriptsize $\pm$5.5} & +1.8\,{\scriptsize $\pm$4.0} & +8.6\,{\scriptsize $\pm$8.9} \\
 & Avg & \textbf{+12.7}\,{\scriptsize $\pm$10.6} & \textbf{+7.7}\,{\scriptsize $\pm$5.7} & +8.2\,{\scriptsize $\pm$5.9} & \textbf{+10.9}\,{\scriptsize $\pm$9.7} \\
 & Knn & +10.8\,{\scriptsize $\pm$11.3} & +5.4\,{\scriptsize $\pm$6.0} & +2.1\,{\scriptsize $\pm$4.7} & +9.2\,{\scriptsize $\pm$9.6} \\
\midrule
Hung.\,+Pen($\alpha$=0.5) & Min & +10.5\,{\scriptsize $\pm$11.7} & +5.5\,{\scriptsize $\pm$5.3} & +2.3\,{\scriptsize $\pm$4.3} & +9.1\,{\scriptsize $\pm$9.7} \\
 & Avg & +12.3\,{\scriptsize $\pm$11.0} & +7.4\,{\scriptsize $\pm$5.7} & \textbf{+8.5}\,{\scriptsize $\pm$6.2} & +10.7\,{\scriptsize $\pm$10.3} \\
 & Knn & +11.6\,{\scriptsize $\pm$12.1} & +5.8\,{\scriptsize $\pm$6.2} & +2.6\,{\scriptsize $\pm$4.9} & +9.5\,{\scriptsize $\pm$10.4} \\
\midrule
Part.\,OT ($\rho$=0.5) & Min & +9.0\,{\scriptsize $\pm$10.1} & +4.8\,{\scriptsize $\pm$4.5} & +4.0\,{\scriptsize $\pm$4.6} & +8.1\,{\scriptsize $\pm$8.3} \\
 & Avg & +8.3\,{\scriptsize $\pm$9.6} & +4.7\,{\scriptsize $\pm$5.9} & +7.6\,{\scriptsize $\pm$6.4} & +8.0\,{\scriptsize $\pm$8.4} \\
 & Knn & +9.7\,{\scriptsize $\pm$10.1} & +5.0\,{\scriptsize $\pm$5.2} & +4.4\,{\scriptsize $\pm$4.9} & +8.0\,{\scriptsize $\pm$7.8} \\
\midrule
Part.\,OT ($\rho$=0.8) & Min & +9.0\,{\scriptsize $\pm$10.1} & +4.8\,{\scriptsize $\pm$4.5} & +4.0\,{\scriptsize $\pm$4.6} & +8.1\,{\scriptsize $\pm$8.3} \\
 & Avg & +8.3\,{\scriptsize $\pm$9.6} & +4.7\,{\scriptsize $\pm$5.9} & +7.6\,{\scriptsize $\pm$6.4} & +8.0\,{\scriptsize $\pm$8.4} \\
 & Knn & +9.7\,{\scriptsize $\pm$10.1} & +5.0\,{\scriptsize $\pm$5.2} & +4.4\,{\scriptsize $\pm$4.9} & +8.0\,{\scriptsize $\pm$7.8} \\
\bottomrule
\end{tabular}
\end{table}
\begin{table}[H]
\caption{\textbf{Multi-finding stratum} --- mean $\pm$ std percentage improvement over random selection per (distance metric, aggregation) policy. Bold: column-best mean.}
\label{tab:clinical-pct-multi-4m}
\centering
\begin{tabular}{llcccc}
\toprule
Distance metric & Agg. & BS-F1 (\%) & RL-F (\%) & METEOR (\%) & RG-F1 (\%) \\
\midrule
Chamfer & Min & +6.7\,{\scriptsize $\pm$14.6} & +1.5\,{\scriptsize $\pm$3.7} & -2.0\,{\scriptsize $\pm$5.0} & +1.8\,{\scriptsize $\pm$6.4} \\
 & Avg & +10.1\,{\scriptsize $\pm$7.3} & +3.4\,{\scriptsize $\pm$3.4} & +3.0\,{\scriptsize $\pm$4.2} & +3.2\,{\scriptsize $\pm$4.4} \\
 & Knn & +7.8\,{\scriptsize $\pm$15.8} & +1.2\,{\scriptsize $\pm$4.1} & -1.9\,{\scriptsize $\pm$5.6} & +1.6\,{\scriptsize $\pm$6.8} \\
\midrule
Hausdorff & Min & +5.5\,{\scriptsize $\pm$12.3} & +1.2\,{\scriptsize $\pm$3.3} & -2.9\,{\scriptsize $\pm$4.5} & +1.6\,{\scriptsize $\pm$5.8} \\
 & Avg & +6.2\,{\scriptsize $\pm$6.5} & +1.0\,{\scriptsize $\pm$2.7} & -0.3\,{\scriptsize $\pm$4.5} & +0.2\,{\scriptsize $\pm$3.8} \\
 & Knn & +6.3\,{\scriptsize $\pm$11.9} & +0.8\,{\scriptsize $\pm$3.6} & -3.1\,{\scriptsize $\pm$5.2} & +0.9\,{\scriptsize $\pm$5.4} \\
\midrule
OT & Min & +8.7\,{\scriptsize $\pm$9.1} & +1.9\,{\scriptsize $\pm$3.1} & +1.9\,{\scriptsize $\pm$5.7} & +2.5\,{\scriptsize $\pm$5.0} \\
 & Avg & +9.9\,{\scriptsize $\pm$8.4} & +3.0\,{\scriptsize $\pm$2.9} & \textbf{+6.6}\,{\scriptsize $\pm$6.2} & +3.7\,{\scriptsize $\pm$4.7} \\
 & Knn & +8.6\,{\scriptsize $\pm$8.5} & +1.8\,{\scriptsize $\pm$2.9} & +1.4\,{\scriptsize $\pm$4.8} & +2.2\,{\scriptsize $\pm$4.6} \\
\midrule
Hung.\,+NN & Min & +7.5\,{\scriptsize $\pm$12.9} & +1.2\,{\scriptsize $\pm$3.6} & -2.3\,{\scriptsize $\pm$4.9} & +1.4\,{\scriptsize $\pm$6.1} \\
 & Avg & +8.2\,{\scriptsize $\pm$8.3} & +2.8\,{\scriptsize $\pm$3.1} & +1.9\,{\scriptsize $\pm$3.6} & +2.5\,{\scriptsize $\pm$3.6} \\
 & Knn & +7.6\,{\scriptsize $\pm$14.8} & +1.1\,{\scriptsize $\pm$4.0} & -2.2\,{\scriptsize $\pm$5.4} & +1.6\,{\scriptsize $\pm$6.6} \\
\midrule
Hung.\,+Pen($\alpha$=0.1) & Min & +7.1\,{\scriptsize $\pm$12.9} & +1.0\,{\scriptsize $\pm$3.4} & -2.2\,{\scriptsize $\pm$4.8} & +1.5\,{\scriptsize $\pm$6.2} \\
 & Avg & +11.0\,{\scriptsize $\pm$10.0} & \textbf{+3.9}\,{\scriptsize $\pm$3.3} & +3.3\,{\scriptsize $\pm$4.7} & \textbf{+4.1}\,{\scriptsize $\pm$5.9} \\
 & Knn & +7.5\,{\scriptsize $\pm$15.2} & +1.2\,{\scriptsize $\pm$4.1} & -2.2\,{\scriptsize $\pm$5.4} & +1.6\,{\scriptsize $\pm$6.7} \\
\midrule
Hung.\,+Pen($\alpha$=0.5) & Min & +6.9\,{\scriptsize $\pm$13.6} & +1.3\,{\scriptsize $\pm$3.4} & -2.3\,{\scriptsize $\pm$4.9} & +1.2\,{\scriptsize $\pm$6.3} \\
 & Avg & \textbf{+11.3}\,{\scriptsize $\pm$10.3} & +3.0\,{\scriptsize $\pm$2.7} & +2.9\,{\scriptsize $\pm$3.8} & +3.1\,{\scriptsize $\pm$4.2} \\
 & Knn & +9.3\,{\scriptsize $\pm$13.7} & +1.5\,{\scriptsize $\pm$3.6} & -2.0\,{\scriptsize $\pm$5.3} & +1.6\,{\scriptsize $\pm$6.8} \\
\midrule
Part.\,OT ($\rho$=0.5) & Min & +7.6\,{\scriptsize $\pm$13.3} & +1.5\,{\scriptsize $\pm$2.5} & +0.7\,{\scriptsize $\pm$4.3} & +2.3\,{\scriptsize $\pm$5.5} \\
 & Avg & +8.5\,{\scriptsize $\pm$6.8} & +2.8\,{\scriptsize $\pm$2.6} & +5.9\,{\scriptsize $\pm$4.4} & +3.5\,{\scriptsize $\pm$4.3} \\
 & Knn & +8.8\,{\scriptsize $\pm$12.6} & +1.4\,{\scriptsize $\pm$2.8} & +0.5\,{\scriptsize $\pm$4.2} & +1.9\,{\scriptsize $\pm$5.2} \\
\midrule
Part.\,OT ($\rho$=0.8) & Min & +7.6\,{\scriptsize $\pm$13.3} & +1.5\,{\scriptsize $\pm$2.5} & +0.7\,{\scriptsize $\pm$4.3} & +2.3\,{\scriptsize $\pm$5.5} \\
 & Avg & +8.5\,{\scriptsize $\pm$6.8} & +2.8\,{\scriptsize $\pm$2.6} & +5.9\,{\scriptsize $\pm$4.4} & +3.5\,{\scriptsize $\pm$4.3} \\
 & Knn & +8.8\,{\scriptsize $\pm$12.6} & +1.4\,{\scriptsize $\pm$2.8} & +0.5\,{\scriptsize $\pm$4.2} & +1.9\,{\scriptsize $\pm$5.2} \\
\bottomrule
\end{tabular}
\end{table}

\subsection{Heatmaps of percent improvement over random (all patients)}
\label{sec:clinical-pct-heatmaps}

\begin{figure}[H]
    \centering
    \includegraphics[width=\linewidth]{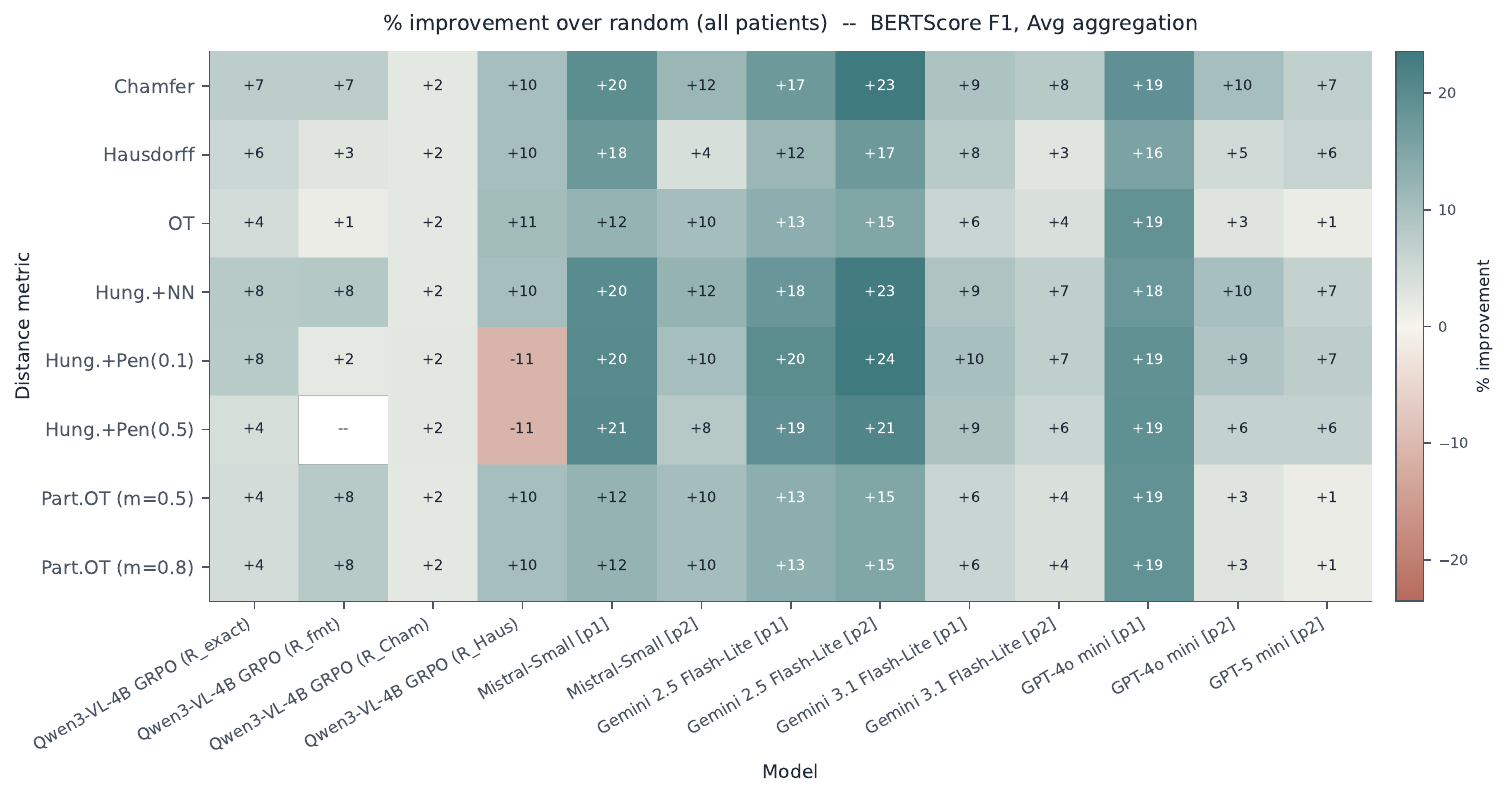}
    \caption{\textbf{Percent improvement over random (all patients) -- BERTScore F1 with Avg aggregation.} Cell value = the per-(model, distance) percentage improvement over random selection on BERTScore F1, computed across the entire test set with the Avg aggregation. Rows are distance metrics; columns are models. Teal cells beat random, coral cells lose to it. Companion to Tab.~\ref{tab:clinical-pct-all-4m} (which averages each column across the same per-model values).}
    \label{fig:clinical-pct-heatmap-bs-f1}
\end{figure}

\begin{figure}[H]
    \centering
    \includegraphics[width=\linewidth]{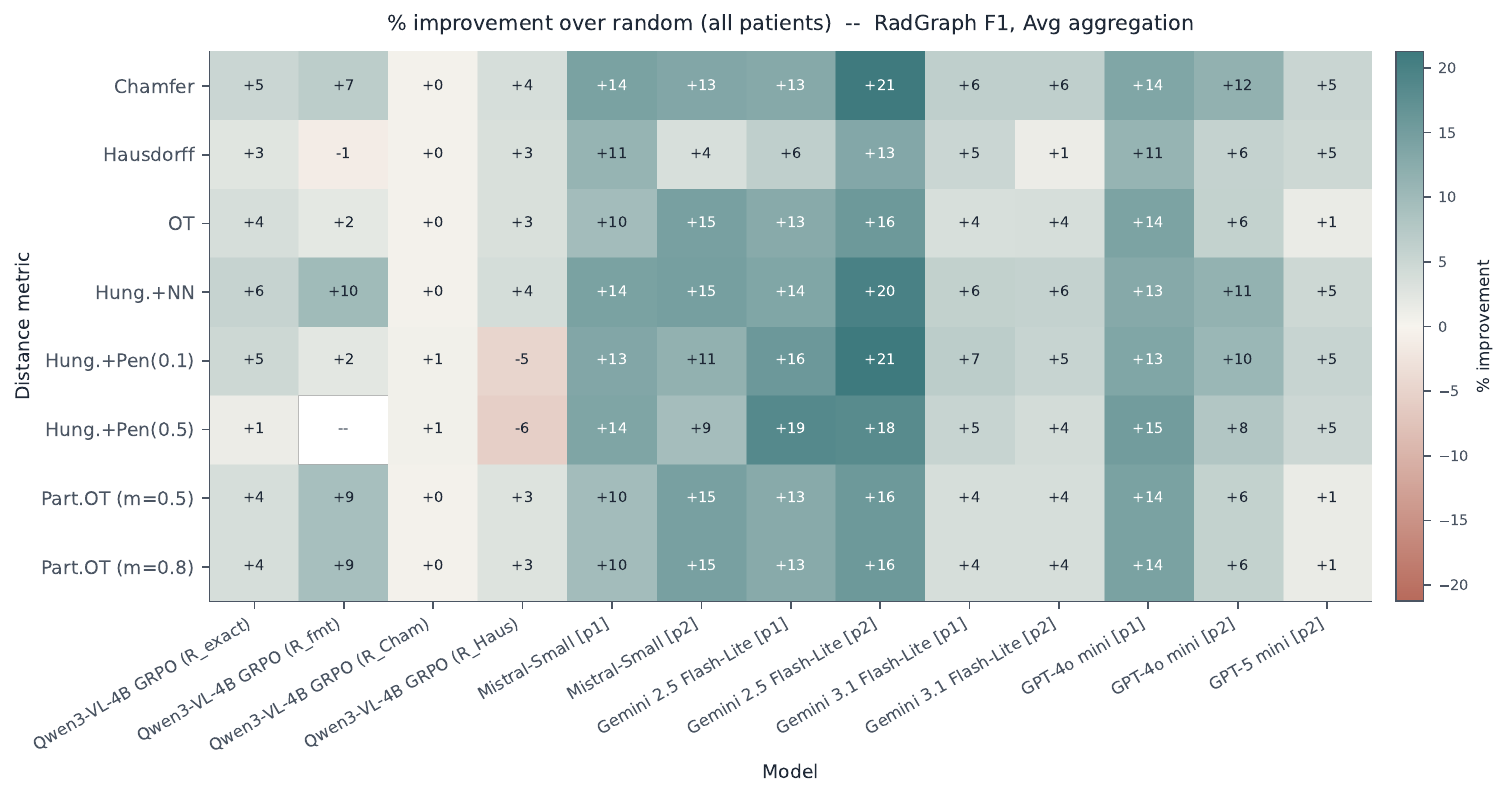}
    \caption{\textbf{Percent improvement over random (all patients) -- RadGraph F1 with Avg aggregation.} Cell value = the per-(model, distance) percentage improvement over random selection on RadGraph F1, computed across the entire test set with the Avg aggregation. Rows are distance metrics; columns are models. Teal cells beat random, coral cells lose to it. Companion to Tab.~\ref{tab:clinical-pct-all-4m} (which averages each column across the same per-model values).}
    \label{fig:clinical-pct-heatmap-rg-f1}
\end{figure}

For the two main-paper clinical metrics (BERTScore F1 and RadGraph F1) we visualise, in Figs.~\ref{fig:clinical-pct-heatmap-bs-f1} and \ref{fig:clinical-pct-heatmap-rg-f1}, the same per-(model, distance) values that feed Tab.~\ref{tab:clinical-pct-all-4m} -- without the cross-model averaging. The aggregation is fixed to Avg throughout. Teal cells indicate that the selection policy beats random selection for that (model, distance) combination; coral cells indicate the opposite.

\section{Full pruning results}
\label{sec:pruning-appendix}

This appendix complements Tab.~\ref{tab:pruning-headline} of the main
paper with every NLP and clinical metric we evaluated, for both the
\textsc{Findings} and \textsc{Impression} sections, under the same three
selection policies (\textit{Random}, \textit{Standard}, \textit{Pruning})
and across every (model, distance metric) combination.

\subsection{Pruning procedure}

Alg.~\ref{alg:pruning} summarises the distance-guided pruning policy
of Sec.~\ref{sec:pruning} in pseudocode. The procedure samples $K$
candidates' opening sentence in lock-step, then alternates a
``decode-one-sentence / score against training distribution / drop
the worst $p\%$ of survivors'' loop until a single candidate remains;
that survivor is decoded to its end-of-sequence token and returned.

\begin{algorithm}[H]
\caption{Distance-guided pruning of stochastic generations.}
\label{alg:pruning}
\begin{algorithmic}[1]
\Require Input X-ray $x$; generative policy $\pi$; training corpus
        $\mathcal{T}=\{r^{(t)}\}_{t=1}^{N}$; candidate budget $K$;
        pruning fraction $p\in(0,1)$
\Ensure Selected response $\hat{y}^{\star}$
\State $\mathcal{C}\!\gets\!$ sample $K$ candidates' first sentence
        from $\pi(\cdot\!\mid\!x)$
\State $t\!\gets\!1$
        \Comment{number of fully decoded sentences per candidate}
\While{$|\mathcal{C}|>1$}
    \State $t\!\gets\!t+1$
    \For{each active candidate $\hat{y}^{(k)}\!\in\!\mathcal{C}$}
        \State decode the $t$-th sentence of $\hat{y}^{(k)}$ from $\pi$
        \State form partial sentence-embedding sets
                $\mathcal{E}^{S}\!\bigl(\hat{y}^{(k)}_{:t}\bigr)$ for
                $S\!\in\!\{F,I\}$
        \State compute the partial-output score
                $\mathfrak{D}\!\bigl(\hat{y}^{(k)}_{:t}\bigr)$
                \Comment{Eq.~\eqref{eq:total-train-dist}}
    \EndFor
    \State drop the highest-scoring
            $\lceil p\,|\mathcal{C}|\rceil$ candidates from $\mathcal{C}$
\EndWhile
\State $\hat{y}^{\star}\!\gets\!$ decode the surviving candidate to its
        end-of-sequence token
\State \Return $\hat{y}^{\star}$
\end{algorithmic}
\end{algorithm}

\begin{table}[H]
\caption{\textbf{Distance-guided pruning of generations (\textsc{Findings}).} For every (model, distance) pair we report the percentage of generation tokens saved by the pruning policy and three headline metrics scored under three selection policies: \textit{Random} (uniform random pick among the $K$ stochastic candidates), \textit{Standard} (full-generation pipeline of Sec.~\ref{sec:response-selection}, distance-based selection on complete candidates) and \textit{Pruning} (distance-guided early-pruning during decoding, this work). Bold marks the column-best of \{random, standard, pruning\} within each metric block.}
\label{tab:pruning-headline-full}
\centering
\small
\setlength{\tabcolsep}{4pt}
\resizebox{\linewidth}{!}{%
\begin{tabular}{lccccccccccccc}
\toprule
 &  & wu & Tok-Save & \multicolumn{3}{c}{BERTScore F1} & \multicolumn{3}{c}{RadGraph F1} & \multicolumn{3}{c}{CheXbert F1} \\
\cmidrule(lr){5-7} \cmidrule(lr){8-10} \cmidrule(lr){11-13}
Model & $\mathcal{D}$ &  & (\%) & Random & Standard & Pruning & Random & Standard & Pruning & Random & Standard & Pruning \\
\midrule
Mistral-Small [p1] & $\mathcal{D}_{\mathrm{Cham}}$ & 1 & 53.4 & 0.201 & \textbf{0.230} & 0.228 & 0.053 & \textbf{0.061} & 0.060 & 0.515 & \textbf{0.616} & 0.595 \\
 & $\mathcal{D}_{\mathrm{Haus}}$ & 1 & 44.2 & 0.201 & \textbf{0.225} & 0.224 & 0.053 & 0.053 & \textbf{0.057} & 0.515 & \textbf{0.619} & 0.605 \\
 & $\mathcal{D}_{\mathrm{Hung}}$ & 1 & 53.4 & 0.201 & \textbf{0.230} & 0.226 & 0.053 & \textbf{0.059} & 0.058 & 0.515 & \textbf{0.616} & 0.599 \\
\midrule
Mistral-Small [p2] & $\mathcal{D}_{\mathrm{Cham}}$ & 1 & 57.4 & 0.282 & 0.325 & \textbf{0.328} & 0.157 & 0.226 & \textbf{0.227} & 0.710 & \textbf{0.726} & 0.725 \\
 & $\mathcal{D}_{\mathrm{Haus}}$ & 1 & 57.6 & 0.282 & 0.325 & \textbf{0.328} & 0.157 & 0.220 & \textbf{0.226} & 0.710 & \textbf{0.726} & 0.725 \\
 & $\mathcal{D}_{\mathrm{Hung}}$ & 1 & 57.2 & 0.282 & 0.326 & \textbf{0.330} & 0.157 & 0.226 & \textbf{0.230} & 0.710 & \textbf{0.726} & 0.725 \\
\midrule
Gemini 2.5 Flash-Lite [p1] & $\mathcal{D}_{\mathrm{Cham}}$ & 2 & 42.1 & 0.196 & 0.234 & \textbf{0.236} & 0.104 & 0.141 & \textbf{0.142} & 0.639 & 0.667 & \textbf{0.672} \\
 & $\mathcal{D}_{\mathrm{Haus}}$ & 1 & 55.9 & 0.196 & \textbf{0.224} & 0.218 & 0.104 & \textbf{0.124} & 0.117 & 0.639 & \textbf{0.665} & 0.648 \\
 & $\mathcal{D}_{\mathrm{Hung}}$ & 2 & 42.2 & 0.196 & 0.235 & \textbf{0.235} & 0.104 & \textbf{0.140} & 0.138 & 0.639 & \textbf{0.670} & 0.667 \\
\midrule
Gemini 2.5 Flash-Lite [p2] & $\mathcal{D}_{\mathrm{Cham}}$ & 1 & 49.4 & 0.240 & 0.280 & \textbf{0.281} & 0.130 & \textbf{0.169} & 0.165 & 0.651 & \textbf{0.697} & 0.691 \\
 & $\mathcal{D}_{\mathrm{Haus}}$ & 1 & 58.8 & 0.240 & \textbf{0.278} & 0.271 & 0.130 & \textbf{0.161} & 0.148 & 0.651 & \textbf{0.701} & 0.669 \\
 & $\mathcal{D}_{\mathrm{Hung}}$ & 1 & 49.3 & 0.240 & \textbf{0.283} & 0.281 & 0.130 & \textbf{0.172} & 0.166 & 0.651 & \textbf{0.695} & 0.688 \\
\midrule
Gemini 3.1 Flash-Lite [p1] & $\mathcal{D}_{\mathrm{Cham}}$ & 1 & 48.9 & 0.244 & \textbf{0.255} & 0.254 & 0.112 & 0.118 & \textbf{0.123} & 0.705 & \textbf{0.706} & 0.699 \\
 & $\mathcal{D}_{\mathrm{Haus}}$ & 1 & 60.1 & 0.244 & \textbf{0.256} & 0.249 & 0.112 & 0.118 & \textbf{0.118} & 0.705 & \textbf{0.708} & \textbf{0.708} \\
 & $\mathcal{D}_{\mathrm{Hung}}$ & 1 & 49.0 & 0.244 & \textbf{0.257} & 0.252 & 0.112 & \textbf{0.119} & 0.118 & 0.705 & \textbf{0.712} & 0.710 \\
\midrule
Gemini 3.1 Flash-Lite [p2] & $\mathcal{D}_{\mathrm{Cham}}$ & 1 & 59.1 & 0.290 & \textbf{0.311} & 0.307 & 0.171 & 0.196 & \textbf{0.197} & 0.701 & \textbf{0.712} & 0.710 \\
 & $\mathcal{D}_{\mathrm{Haus}}$ & 1 & 60.5 & 0.290 & \textbf{0.309} & 0.303 & 0.171 & 0.188 & \textbf{0.189} & 0.701 & 0.707 & \textbf{0.709} \\
 & $\mathcal{D}_{\mathrm{Hung}}$ & 1 & 49.4 & 0.290 & \textbf{0.312} & 0.309 & 0.171 & \textbf{0.192} & 0.192 & 0.701 & \textbf{0.709} & 0.702 \\
\midrule
GPT-4o mini [p1] & $\mathcal{D}_{\mathrm{Cham}}$ & 1 & 55.3 & 0.206 & \textbf{0.244} & 0.238 & 0.080 & \textbf{0.105} & 0.097 & 0.688 & \textbf{0.714} & 0.702 \\
 & $\mathcal{D}_{\mathrm{Haus}}$ & 1 & 55.0 & 0.206 & \textbf{0.240} & 0.238 & 0.080 & \textbf{0.099} & 0.098 & 0.688 & \textbf{0.713} & 0.707 \\
 & $\mathcal{D}_{\mathrm{Hung}}$ & 1 & 55.2 & 0.206 & \textbf{0.244} & 0.238 & 0.080 & \textbf{0.103} & 0.097 & 0.688 & \textbf{0.715} & 0.706 \\
\midrule
GPT-4o mini [p2] & $\mathcal{D}_{\mathrm{Cham}}$ & 1 & 59.1 & 0.291 & 0.323 & \textbf{0.327} & 0.167 & 0.213 & \textbf{0.217} & 0.696 & \textbf{0.721} & 0.719 \\
 & $\mathcal{D}_{\mathrm{Haus}}$ & 1 & 59.3 & 0.291 & \textbf{0.321} & 0.321 & 0.167 & 0.206 & \textbf{0.209} & 0.696 & \textbf{0.721} & 0.716 \\
 & $\mathcal{D}_{\mathrm{Hung}}$ & 1 & 59.0 & 0.291 & \textbf{0.326} & 0.323 & 0.167 & 0.214 & \textbf{0.215} & 0.696 & \textbf{0.721} & 0.720 \\
\midrule
GPT-5 mini [p2] & $\mathcal{D}_{\mathrm{Cham}}$ & 1 & 60.7 & 0.225 & 0.272 & \textbf{0.279} & 0.123 & 0.160 & \textbf{0.164} & 0.697 & 0.725 & \textbf{0.726} \\
 & $\mathcal{D}_{\mathrm{Haus}}$ & 1 & 59.8 & 0.225 & 0.275 & \textbf{0.275} & 0.123 & 0.161 & \textbf{0.165} & 0.697 & \textbf{0.723} & 0.714 \\
 & $\mathcal{D}_{\mathrm{Hung}}$ & 1 & 60.8 & 0.225 & 0.275 & \textbf{0.279} & 0.123 & 0.160 & \textbf{0.164} & 0.697 & 0.725 & \textbf{0.726} \\
\midrule
\multirow{3}{*}{\textbf{Mean}} & $\mathcal{D}_{\mathrm{Cham}}$ & -- & 53.9\,{\scriptsize $\pm$6.1} & 0.242\,{\scriptsize $\pm$0.038} & 0.275\,{\scriptsize $\pm$0.037} & \textbf{0.275}\,{\scriptsize $\pm$0.039} & 0.122\,{\scriptsize $\pm$0.040} & 0.154\,{\scriptsize $\pm$0.054} & \textbf{0.155}\,{\scriptsize $\pm$0.055} & 0.667\,{\scriptsize $\pm$0.062} & \textbf{0.698}\,{\scriptsize $\pm$0.036} & 0.693\,{\scriptsize $\pm$0.041} \\
 & $\mathcal{D}_{\mathrm{Haus}}$ & -- & 56.8\,{\scriptsize $\pm$5.1} & 0.242\,{\scriptsize $\pm$0.038} & \textbf{0.273}\,{\scriptsize $\pm$0.039} & 0.270\,{\scriptsize $\pm$0.041} & 0.122\,{\scriptsize $\pm$0.040} & \textbf{0.148}\,{\scriptsize $\pm$0.054} & 0.148\,{\scriptsize $\pm$0.055} & 0.667\,{\scriptsize $\pm$0.062} & \textbf{0.698}\,{\scriptsize $\pm$0.035} & 0.689\,{\scriptsize $\pm$0.040} \\
 & $\mathcal{D}_{\mathrm{Hung}}$ & -- & 52.8\,{\scriptsize $\pm$5.9} & 0.242\,{\scriptsize $\pm$0.038} & \textbf{0.276}\,{\scriptsize $\pm$0.038} & 0.275\,{\scriptsize $\pm$0.039} & 0.122\,{\scriptsize $\pm$0.040} & \textbf{0.154}\,{\scriptsize $\pm$0.054} & 0.153\,{\scriptsize $\pm$0.056} & 0.667\,{\scriptsize $\pm$0.062} & \textbf{0.699}\,{\scriptsize $\pm$0.036} & 0.694\,{\scriptsize $\pm$0.040} \\
\bottomrule
\end{tabular}%
}
\end{table}

\newpage
\section{Qualitative examples}
\label{sec:examples-appendix}

We provide a set of qualitative examples that illustrate when the
distance-to-training-distribution selection rule (used at inference
time for closed-source LLMs and exploratory ablations) picks a
candidate that is also the closest to the ground truth report
in BERTScore-F1. The cases below are drawn from the chest-X-ray
report-generation task on ReXGradient. For each example we show
the input image, the ground-truth report, the candidate selected
by the distance rule, and one of the rejected candidates that
scored lower in BERTScore-F1 against the ground truth.

        \begin{figure}[H]
          \centering
          \noindent
          \begin{minipage}[c]{0.28\linewidth}
            \centering
            \includegraphics[width=\linewidth, height=4.5cm, keepaspectratio]{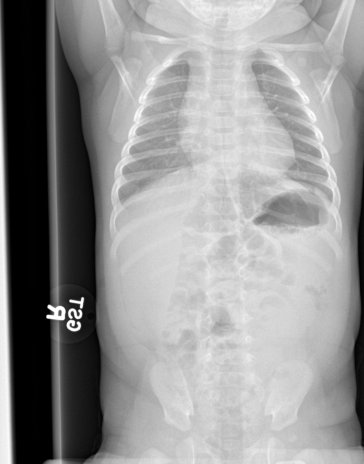}
          \end{minipage}\hfill
          \begin{minipage}[c]{0.69\linewidth}
            \footnotesize\raggedright
            \textbf{Model:} Qwen3-VL-4B GRPO $R_{\mathrm{exact}}$\\
            \textbf{Distance metric:} Partial OT ($\rho{=}0.5$)\\
            \textbf{Aggregation:} min\\
            \textbf{Number of candidates:} $K{=}10$\\[2pt]
            \textit{Each panel below shows a candidate report with its
            BERTScore-F1 against the joined ground-truth report and its
            distance to the training distribution under the chosen
            distance metric and aggregation.}
          \end{minipage}

          \vspace{6pt}

          \begin{tcolorbox}[
    enhanced, breakable, boxsep=2pt, left=4pt, right=4pt, top=2pt, bottom=2pt,
    colback=gray!6, colframe=black!60, coltitle=white, arc=2pt,
    fonttitle=\bfseries\footnotesize, title={Ground truth report}]
  \footnotesize \textbf{Findings:} The lungs are well-aerated and clear. There is no evidence of focal opacification, pleural effusion or pneumothorax. The heart is normal in size; the mediastinal contour is within normal limits. No acute osseous abnormalities are seen. \\[2pt] \textbf{Impression:} No acute cardiopulmonary process seen.
\end{tcolorbox}
\vspace{2pt}
\begin{tcolorbox}[
    enhanced, breakable, boxsep=2pt, left=4pt, right=4pt, top=2pt, bottom=2pt,
    colback=green!6, colframe=green!50!black, coltitle=white, arc=2pt,
    fonttitle=\bfseries\footnotesize, title={Selected candidate\hfill\textnormal{\footnotesize BERTScore-F1: \textbf{0.903}\quad Dist: \textbf{2.94e-12}}}]
  \footnotesize \textbf{Findings:} The lungs are well-aerated and clear. There is no evidence of focal opacification, pleural effusion or pneumothorax. The heart is normal in size; the mediastinal contour is within normal limits. No acute osseous abnormalities are seen. \\[2pt] \textbf{Impression:} 1. No acute cardiopulmonary process seen. 2. No displaced rib fractures identified.
\end{tcolorbox}
\vspace{2pt}
\begin{tcolorbox}[
    enhanced, breakable, boxsep=2pt, left=4pt, right=4pt, top=2pt, bottom=2pt,
    colback=red!4, colframe=red!55!black, coltitle=white, arc=2pt,
    fonttitle=\bfseries\footnotesize, title={Rejected candidate \#1\hfill\textnormal{\footnotesize BERTScore-F1: \textbf{0.211}\quad Dist: \textbf{0.872}}}]
  \footnotesize \textbf{Findings:} The lungs are well expanded and demonstrate no evidence of focal airspace disease or pneumothorax. The cardiopericardial silhouette is enlarged. The Port-A-Cath tip is in the superior vena cava. The patient is rotated on the frontal view. The visualized portions of the bowel gas pattern are grossly unremarkable. \\[2pt] \textbf{Impression:} 1. Cardiomegaly without evidence of acute cardiopulmonary disease. 2. Port-A-Cath tip in the superior vena cava.
\end{tcolorbox}
\vspace{2pt}
\begin{tcolorbox}[
    enhanced, breakable, boxsep=2pt, left=4pt, right=4pt, top=2pt, bottom=2pt,
    colback=red!4, colframe=red!55!black, coltitle=white, arc=2pt,
    fonttitle=\bfseries\footnotesize, title={Rejected candidate \#3\hfill\textnormal{\footnotesize BERTScore-F1: \textbf{0.113}\quad Dist: \textbf{0.689}}}]
  \footnotesize \textbf{Findings:} Umbilical venous catheter tip projects at the inferior T9 level, within the inferior right atrium. Orogastric tube tip projects over the gastric bubble. Lung volumes are low. No focal airspace consolidation or pulmonary edema. No pleural effusion or pneumothorax. Normal cardiomediastinal silhouette. The bowel gas pattern is nonobstructive. No radiopaque calculi. \\[2pt] \textbf{Impression:} 1. Umbilical venous catheter tip projects at the inferior T9 level, within the inferior right atrium. Consider retraction of 1 cm. 2. Low lung volumes.
\end{tcolorbox}
\vspace{2pt}
\begin{tcolorbox}[
    enhanced, breakable, boxsep=2pt, left=4pt, right=4pt, top=2pt, bottom=2pt,
    colback=red!4, colframe=red!55!black, coltitle=white, arc=2pt,
    fonttitle=\bfseries\footnotesize, title={Rejected candidate \#2\hfill\textnormal{\footnotesize BERTScore-F1: \textbf{0.044}\quad Dist: \textbf{0.707}}}]
  \footnotesize \textbf{Findings:} Umbilical artery catheter tip projects at the T6 level. Umbilical vein catheter tip projects at the T7 level. Both lungs are clear. The visualized skeletal structures are unremarkable. \\[2pt] \textbf{Impression:} 1. Umbilical artery catheter tip projects at the T6 level. Umbilical vein catheter tip projects at the T7 level. 2. No acute cardiopulmonary disease.
\end{tcolorbox}
          \caption{The selection rule picks the candidate with the lowest distance to the training distribution; that candidate also has the highest BERTScore-F1 against the ground truth. The three rejected alternatives shown were among the candidates whose distance to the training distribution was the larger under the chosen $(\textsc{metric},\textsc{agg})$ pair, and they correspondingly score lower in BERTScore-F1.}
          \label{fig:example-436}
        \end{figure}

        \begin{figure}[H]
          \centering
          \noindent
          \begin{minipage}[c]{0.28\linewidth}
            \centering
            \includegraphics[width=\linewidth, height=4.5cm, keepaspectratio]{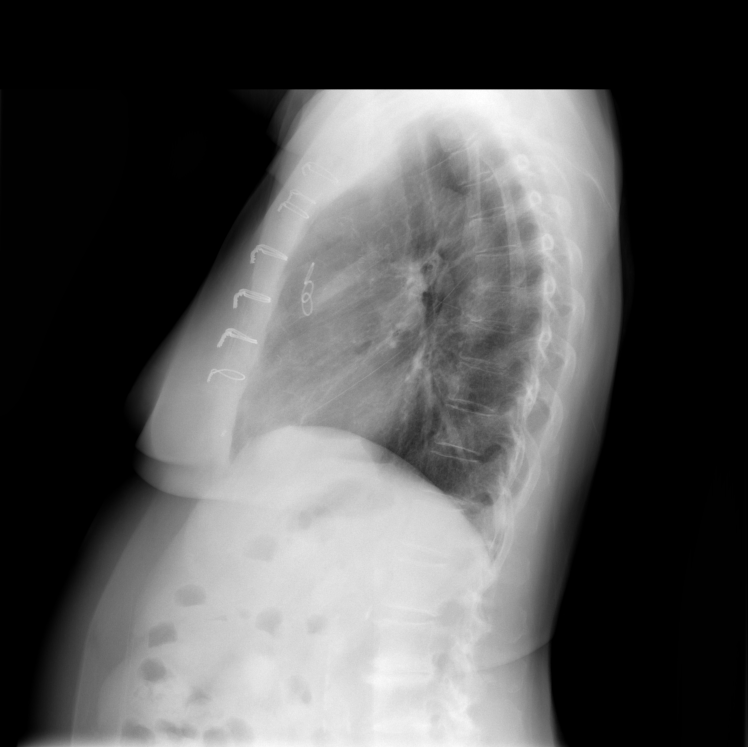}
          \end{minipage}\hfill
          \begin{minipage}[c]{0.69\linewidth}
            \footnotesize\raggedright
            \textbf{Model:} Gemini 2.5 Flash-Lite\\
            \textbf{Distance metric:} Hungarian + Pen ($\alpha{=}0.5$)\\
            \textbf{Aggregation:} avg\\
            \textbf{Number of candidates:} $K{=}5$\\[2pt]
            \textit{Each panel below shows a candidate report with its
            BERTScore-F1 against the joined ground-truth report and its
            distance to the training distribution under the chosen
            distance metric and aggregation.}
          \end{minipage}

          \vspace{6pt}

          \begin{tcolorbox}[
    enhanced, breakable, boxsep=2pt, left=4pt, right=4pt, top=2pt, bottom=2pt,
    colback=gray!6, colframe=black!60, coltitle=white, arc=2pt,
    fonttitle=\bfseries\footnotesize, title={Ground truth report}]
  \footnotesize \textbf{Findings:} Cardiac silhouette is normal in size and configuration. Changes from the recent CABG surgery are noted. No mediastinal widening. No mediastinal or hilar masses or evidence adenopathy. Lungs are clear. No pleural effusion or pneumothorax. Skeletal structures are intact. \\[2pt] \textbf{Impression:} 1. No active cardiopulmonary disease. 2. Postsurgical atelectasis and effusions have resolved since the prior exam.
\end{tcolorbox}
\vspace{2pt}
\begin{tcolorbox}[
    enhanced, breakable, boxsep=2pt, left=4pt, right=4pt, top=2pt, bottom=2pt,
    colback=green!6, colframe=green!50!black, coltitle=white, arc=2pt,
    fonttitle=\bfseries\footnotesize, title={Selected candidate\hfill\textnormal{\footnotesize BERTScore-F1: \textbf{0.364}\quad Dist: \textbf{1.91}}}]
  \footnotesize \textbf{Findings:} Surgical clips and wires are noted anteriorly, likely related to sternal closure. The lungs are clear. The cardiac silhouette and mediastinal contours are within normal limits. No pleural effusion or pneumothorax is identified. Visualized bones are unremarkable. \\[2pt] \textbf{Impression:} Post-sternotomy changes. No acute cardiopulmonary abnormality.
\end{tcolorbox}
\vspace{2pt}
\begin{tcolorbox}[
    enhanced, breakable, boxsep=2pt, left=4pt, right=4pt, top=2pt, bottom=2pt,
    colback=red!4, colframe=red!55!black, coltitle=white, arc=2pt,
    fonttitle=\bfseries\footnotesize, title={Rejected candidate \#1\hfill\textnormal{\footnotesize BERTScore-F1: \textbf{0.216}\quad Dist: \textbf{2.23}}}]
  \footnotesize \textbf{Findings:} Multiple sternal wires are identified anteriorly. The lungs are clear. The cardiomediastinal silhouette is normal. \\[2pt] \textbf{Impression:} Post-sternotomy wires. No acute cardiopulmonary abnormality.
\end{tcolorbox}
\vspace{2pt}
\begin{tcolorbox}[
    enhanced, breakable, boxsep=2pt, left=4pt, right=4pt, top=2pt, bottom=2pt,
    colback=red!4, colframe=red!55!black, coltitle=white, arc=2pt,
    fonttitle=\bfseries\footnotesize, title={Rejected candidate \#2\hfill\textnormal{\footnotesize BERTScore-F1: \textbf{0.142}\quad Dist: \textbf{2.06}}}]
  \footnotesize \textbf{Findings:} Lateral view of the chest shows a metallic surgical wire wrapping around the sternum, consistent with sternal fixation. The lungs are clear. The heart and mediastinum appear within normal limits for this projection. The visualized spine shows degenerative changes. \\[2pt] \textbf{Impression:} Post-sternotomy wires. No acute cardiopulmonary process is identified.
\end{tcolorbox}
\vspace{2pt}
\begin{tcolorbox}[
    enhanced, breakable, boxsep=2pt, left=4pt, right=4pt, top=2pt, bottom=2pt,
    colback=red!4, colframe=red!55!black, coltitle=white, arc=2pt,
    fonttitle=\bfseries\footnotesize, title={Rejected candidate \#3\hfill\textnormal{\footnotesize BERTScore-F1: \textbf{0.222}\quad Dist: \textbf{1.98}}}]
  \footnotesize \textbf{Findings:} Multiple surgical wires are noted along the anterior sternum. The lungs are clear. The heart and mediastinal contours are within normal limits. No acute bony abnormality is identified. \\[2pt] \textbf{Impression:} Post-sternotomy wires. No acute cardiopulmonary process.
\end{tcolorbox}
          \caption{The selection rule picks the candidate with the lowest distance to the training distribution; that candidate also has the highest BERTScore-F1 against the ground truth. The three rejected alternatives shown were among the candidates whose distance to the training distribution was the larger under the chosen $(\textsc{metric},\textsc{agg})$ pair, and they correspondingly score lower in BERTScore-F1.}
          \label{fig:example-434}
        \end{figure}

        \begin{figure}[H]
          \centering
          \noindent
          \begin{minipage}[c]{0.28\linewidth}
            \centering
            \includegraphics[width=\linewidth, height=4.5cm, keepaspectratio]{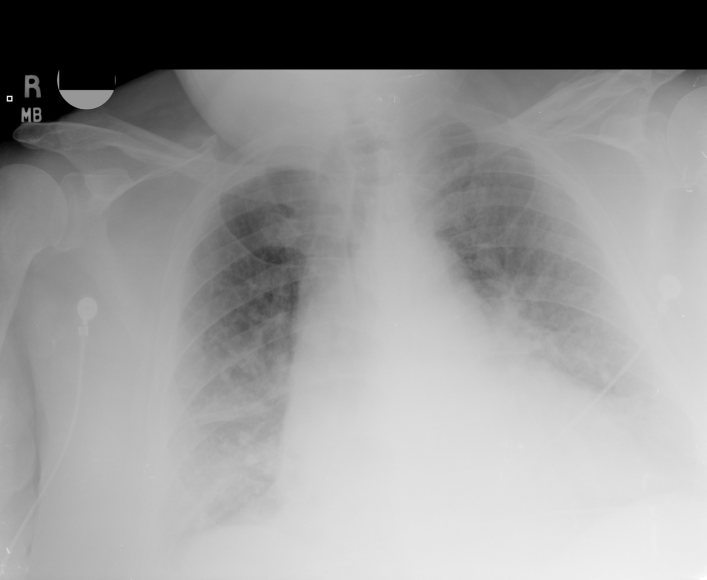}
          \end{minipage}\hfill
          \begin{minipage}[c]{0.69\linewidth}
            \footnotesize\raggedright
            \textbf{Model:} Gemini 2.5 Flash-Lite\\
            \textbf{Distance metric:} Hausdorff\\
            \textbf{Aggregation:} min\\
            \textbf{Number of candidates:} $K{=}5$\\[2pt]
            \textit{Each panel below shows a candidate report with its
            BERTScore-F1 against the joined ground-truth report and its
            distance to the training distribution under the chosen
            distance metric and aggregation.}
          \end{minipage}

          \vspace{6pt}

          \begin{tcolorbox}[
    enhanced, breakable, boxsep=2pt, left=4pt, right=4pt, top=2pt, bottom=2pt,
    colback=gray!6, colframe=black!60, coltitle=white, arc=2pt,
    fonttitle=\bfseries\footnotesize, title={Ground truth report}]
  \footnotesize \textbf{Findings:} Examination limited by positioning and technique. The costophrenic sulci are excluded from the field of view. The cardiac silhouette is enlarged. There are congestive changes bilaterally. It is likely pleural fluid bilaterally as well. Bony thorax is intact. \\[2pt] \textbf{Impression:} Limited examination. CHF pattern. Enlarged cardiac silhouette, congestive changes, and probable effusions.
\end{tcolorbox}
\vspace{2pt}
\begin{tcolorbox}[
    enhanced, breakable, boxsep=2pt, left=4pt, right=4pt, top=2pt, bottom=2pt,
    colback=green!6, colframe=green!50!black, coltitle=white, arc=2pt,
    fonttitle=\bfseries\footnotesize, title={Selected candidate\hfill\textnormal{\footnotesize BERTScore-F1: \textbf{0.203}\quad Dist: \textbf{0.599}}}]
  \footnotesize \textbf{Findings:} Diffuse bilateral airspace opacities are noted, more prominent at the bases, consistent with pulmonary edema. The cardiac silhouette is enlarged. There is a right pleural effusion. The visualized bony structures are unremarkable. \\[2pt] \textbf{Impression:} Pulmonary edema, likely cardiogenic. Right pleural effusion.
\end{tcolorbox}
\vspace{2pt}
\begin{tcolorbox}[
    enhanced, breakable, boxsep=2pt, left=4pt, right=4pt, top=2pt, bottom=2pt,
    colback=red!4, colframe=red!55!black, coltitle=white, arc=2pt,
    fonttitle=\bfseries\footnotesize, title={Rejected candidate \#3\hfill\textnormal{\footnotesize BERTScore-F1: \textbf{0.078}\quad Dist: \textbf{0.77}}}]
  \footnotesize \textbf{Findings:} There are diffuse bilateral opacities, most prominent in the lower lung zones, suggestive of airspace disease. The cardiomediastinal silhouette is enlarged and indistinct, potentially related to underlying disease or effusion. A nasogastric tube is present, terminating in the stomach. \\[2pt] \textbf{Impression:} Diffuse bilateral airspace disease, etiology uncertain, but concerning for pneumonia or ARDS. Cardiomegaly and possible pericardial effusion.
\end{tcolorbox}

\vspace{2pt}
\begin{tcolorbox}[
    enhanced, breakable, boxsep=2pt, left=4pt, right=4pt, top=2pt, bottom=2pt,
    colback=red!4, colframe=red!55!black, coltitle=white, arc=2pt,
    fonttitle=\bfseries\footnotesize, title={Rejected candidate \#1\hfill\textnormal{\footnotesize BERTScore-F1: \textbf{0.029}\quad Dist: \textbf{0.827}}}]
  \footnotesize \textbf{Findings:} Diffuse bilateral opacities are present throughout the lungs, more pronounced at the bases. The cardiac silhouette is enlarged. Subcutaneous air is noted in the right axilla and upper chest wall. \\[2pt] \textbf{Impression:} Diffuse pneumonia or ARDS. Cardiomegaly. Subcutaneous emphysema.
\end{tcolorbox}
\vspace{2pt}
\begin{tcolorbox}[
    enhanced, breakable, boxsep=2pt, left=4pt, right=4pt, top=2pt, bottom=2pt,
    colback=red!4, colframe=red!55!black, coltitle=white, arc=2pt,
    fonttitle=\bfseries\footnotesize, title={Rejected candidate \#2\hfill\textnormal{\footnotesize BERTScore-F1: \textbf{-0.033}\quad Dist: \textbf{0.778}}}]
  \footnotesize \textbf{Findings:} Diffuse bilateral airspace opacities, more prominent in the lower lobes. cardiomegaly. The visualized left upper extremity demonstrates a malpositioned central venous catheter. \\[2pt] \textbf{Impression:} Diffuse bilateral airspace disease, concerning for pneumonia or acute respiratory distress syndrome. Cardiomegaly.
\end{tcolorbox}

          \caption{The selection rule picks the candidate with the lowest distance to the training distribution; that candidate also has the highest BERTScore-F1 against the ground truth. The three rejected alternatives shown were among the candidates whose distance to the training distribution was the larger under the chosen $(\textsc{metric},\textsc{agg})$ pair, and they correspondingly score lower in BERTScore-F1.}
          \label{fig:example-384}
        \end{figure}

\end{document}